\documentclass{article}

\PassOptionsToPackage{numbers,sort&compress}{natbib}
\usepackage[main, final]{neurips_2026}

\usepackage{amsthm}
\newtheorem{lemma}{Lemma}[section]

\usepackage[utf8]{inputenc} 
\usepackage[T1]{fontenc}    
\usepackage{hyperref}       
\usepackage{url}            
\usepackage{booktabs}       
\usepackage{amsfonts}       
\usepackage{nicefrac}       
\usepackage{microtype}      
\usepackage{xcolor}         
\usepackage{amssymb}
\usepackage{multirow}
\usepackage{graphicx}
\usepackage{amsmath}
\usepackage{makecell}
\usepackage{xspace}
\usepackage{color}
\usepackage{wrapfig}
\usepackage{bbding}
\usepackage{tcolorbox}
\usepackage{enumitem}
\usepackage{amsthm}

\usepackage{array}
\usepackage{tabularx}
\usepackage{tcolorbox}
\usepackage{graphicx}
\usepackage{caption}
\usepackage{subcaption}
\usepackage{booktabs}
\usepackage{tabularx}
\newcolumntype{C}{>{\centering\arraybackslash}X}
\usepackage{bm}
\usepackage{pifont}
\usepackage{amssymb}
\usepackage{xcolor}           
\usepackage{tcolorbox}
\definecolor{myorange}{RGB}{230,97,0}
\definecolor{myblue}{RGB}{30,144,255}
\definecolor{mygreen}{RGB}{44,160,44}

\newcolumntype{C}{>{\centering\arraybackslash}X}              
\newcolumntype{Y}{>{\centering\arraybackslash}p{2.1cm}}       
\newcolumntype{Z}{>{\centering\arraybackslash}p{2.7cm}}       

\title{ShiftRAG: Bypassing the Textual Bottleneck via Decoupled Learning and Continuous Soft Tokens}

\author{
Youmin Ko$^{1}$, Jihong Jeong$^{1}$, Hyunjoon Kim$^{1,2}$\thanks{Corresponding author.}\\
$^{1}$Department of Artificial Intelligence, Hanyang University\\
$^{2}$Department of Data Science, Hanyang University\\
\texttt{\{youminkk0213, jjh103, hyunjoonkim\}@hanyang.ac.kr}
}

\begin{document}

\maketitle

\begin{abstract}

Despite advances in retrieval-augmented generation (RAG), suppressing hallucinations during complex reasoning remains a persistent challenge. We identify that existing RAG pipelines suffer from a textual bottleneck: by passing only raw text to the generator, they discard discriminative signals of the retriever. Furthermore, recent attempts to unify retrieval and generation within a single LLM induce severe objective conflict and exacerbate the rank collapse of decoder-only architectures, degrading performance on both tasks. To address these limitations, we introduce ShiftRAG, a decoupled RAG framework that employs a single LLM backbone but isolates discriminative retrieval learning from autoregressive generation. To bridge these decoupled modes, we project retrieval alignments into continuous soft tokens, establishing a high-bandwidth interface that  infuses the generator with retrieval-side signals. Concurrently, our soft orthogonality objective mitigates rank collapse, while layer-wise relevance dynamics preserve the fine-grained latent space required for highly discriminative evidence separation. Extensive  evaluations demonstrate that ShiftRAG significantly improves retrieval and question answering performance over state-of-the-art baselines. ShiftRAG preserves foundational capabilities of the backbone while maintaining a favorable efficiency-performance trade-off on general text encoding and generative tasks. Our code is available\footnote{https://github.com/meaningful96/ShiftRAG}.
\end{abstract}

\section{Introduction}
Retrieval-augmented generation (RAG) \cite{RAG} has emerged as the definitive framework for grounding LLMs with external knowledge across diverse knowledge-intensive domains, including multi-hop question answering (MHQA) \cite{mhqa1,mhqa2,mhqa3} and knowledge graph reasoning \cite{kgqa1,kgqa2,kgqa3,kgqa4}. Current research has advanced RAG through three avenues: (i) learning dense representations specifically for retrieval \cite{DPR, Contriever,ColBERT,ColBERTv2}, (ii) adapting decoder-only LLMs into text encoders \cite{NV-Embed, LLM2Vec, GTE}, and (iii) supporting both retrieval and generation within a single LLM backbone \cite{GritLM, OneGen}. Meanwhile, researchers have proposed to augment reasoning over retrieved evidence through explicit Chain-of-Thought \cite{CoT, CoQ} and latent reasoning \cite{Sim-CoT, CODI}. Despite these sophisticated architectural enhancements, suppressing hallucinations \cite{hallu1,hallu2,hallu3,hallu4,hallu5} during retrieval-based reasoning remains a persistent challenge. We claim that this limitation stems from three factors.

\begin{figure}[t]
    \centering
    \includegraphics[width=1\textwidth]{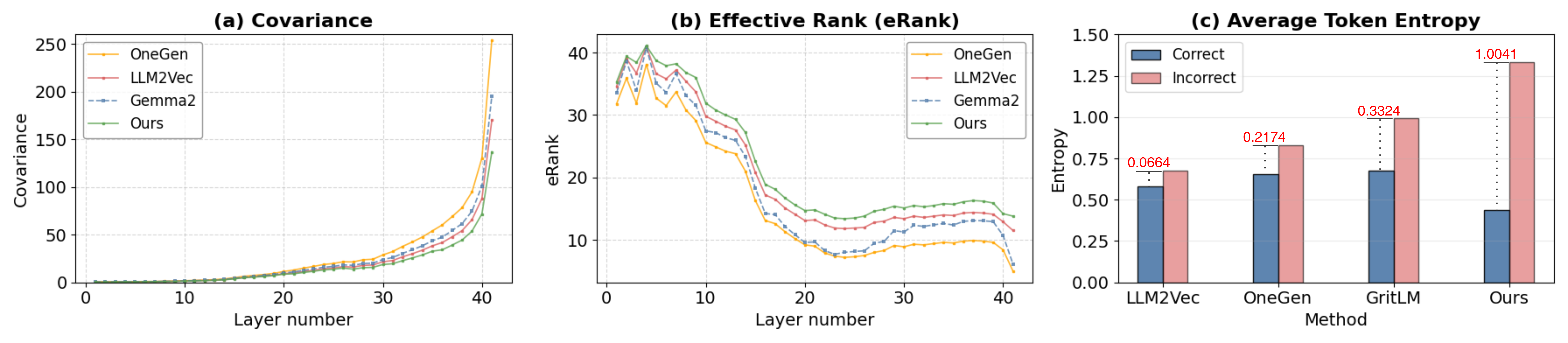}
    \caption{Motivating observations on layer-wise rank collapse and evidence-sensitive generation. (a) Layer-wise token covariance. (b) Effective rank (eRank) of token representations. (c) Average token entropy on 2WikiMultihopQA by answer correctness.}
    \label{fig:intro}
    \vspace{-15pt}
\end{figure}

First, overcoming the rank collapse inherent to decoder-only LLMs \cite{rc5, rc6} while preserving generative capacity remains a fundamental bottleneck. In deep layers, token representations naturally concentrate into a low-dimensional subspace, blurring the fine-grained semantic distinctions necessary to separate true evidence from distractors. Figures \ref{fig:intro}(a) and \ref{fig:intro}(b) diagnose this collapse, revealing high token covariance\footnote{Covariance between hidden representations of tokens, where larger values indicate stronger pairwise directional alignment.} and low effective rank (eRank)\footnote{Smaller values indicate fewer effective independent directions.} \cite{eRank}. Current paradigms fail to resolve this trade-off: while contrastive adaptation in deep encoders \cite{NV-Embed, GTE, LLM2Vec} mitigates rank collapse, it severely degrades autoregressive generation. Unified LLMs \cite{GritLM, OneGen} preserve generation via backbone sharing, but inadvertently exacerbate rank collapse, exhibiting higher covariance and lower eRank than the base model (Gemma 2)\footnote{More detailed analyses based on other LLMs are provided in Appendix \ref{appendix_analysis_eRank_backbones}.}. Sharing a backbone or relying on standard attention mechanisms is insufficient to maintain both generative fluency and the discriminative representations for retrieval.

Second, standard RAG frameworks~\cite{Contriever, ColBERTv2, Proposition} suffer from a textual bottleneck, i.e., they transfer only raw document text to the LLM, discarding the implicit signals (the semantic alignment between question and document embeddings) from the retriever. Consequently, the generator is deprived of the retriever's underlying discriminative confidence. Figure \ref{fig:intro}(c) highlights this limitation: while an evidence-sensitive generator \cite{evi-sense} should exhibit distinct predictive entropy gaps between correct and incorrect outputs, existing state-of-the-art baselines show small entropy differences\footnote{Appendix \ref{appendix_analysis_overconfidence} shows the same trend for other datasets.}. This indicates that raw text alone fails to convey the strength of the supporting evidence. While Chain-of-Thought \cite{CoT, CoQ} and continuous latent reasoning \cite{CODI, Sim-CoT} attempt to extract deeper rationales, they still derive these representations exclusively from explicit text. Lacking a mechanism to directly inject the embedding-level alignment of the retriever, these methods remain constrained by the textual interface.

Third, unifying retrieval and generation under shared adaptation parameters induces severe objective conflict. Discriminative retrieval requires a latent topology optimized for contrastive separation and fine-grained diversity by aligning the representation of a question with its supporting documents and separating it from distractors, whereas autoregressive generation organizes the representation space for sequential next-token prediction. In unified architectures \cite{GritLM, OneGen}, forcing these fundamentally divergent tasks to share parameters creates competing optimization pressures, leading to representation instability and negative transfer. Even if adaptation parameters are decoupled, simultaneous joint training remains suboptimal; the generator is forced to optimize against a continuously shifting retrieval-conditioned interface, destabilizing retrieval convergence and downstream reasoning.

To address these issues, we introduce \textbf{ShiftRAG}, a novel RAG framework that employs a single LLM backbone across decoupled encoder and decoder modes. ShiftRAG transcends the standard textual interface by translating high-dimensional retrieval knowledge directly into latent reasoning signals. Our contributions are threefold. First, we cultivate a highly discriminative embedding space by exploiting layer-wise relevance dynamics, effectively mitigating the deep-layer rank collapse typical of decoder-only LLMs. Second, we project question–document embeddings and relevance scores into generation-friendly soft knowledge tokens. This latent bridge allows implicit retrieval alignments and explicit document text to jointly guide the reasoning of the generator. Third, by training the encoder and decoder modes independently and bridging them only at inference, we circumvent the severe objective conflict that plagues joint parameter optimization.

Extensive evaluations demonstrate that ShiftRAG achieves state-of-the-art retrieval and end-to-end QA performance across multi-hop (HotpotQA, 2WikiMultihopQA, MuSiQue) and single-hop (NaturalQuestions) benchmarks. Furthermore, auxiliary evaluations on text encoding and general generative tasks confirm that our decoupled approach preserves foundational capabilities of the backbone while maintaining a  favorable efficiency-performance trade-off, establishing a robust new paradigm for retrieval-augmented reasoning.

\section{Related Work}

\textbf{RAG.} Traditional dense retrieval maps queries and documents into unified vector spaces~\cite{Contriever, DPR, GTR}, while subsequent methods capture finer-grained interactions~\cite{ColBERT, ColBERTv2} or exploit layer-wise representations~\cite{CoopRAG}. Recently, LLMs have been adapted as powerful text encoders~\cite{GTE, NV-Embed, LLM2Vec} and unified models capable of both retrieval and generation~\cite{GritLM, OneGen, MAGNET, GRACE}. However, despite these architectural advances, the RAG pipeline remains fundamentally disjointed. Because both LLM-based encoders and unified models interface with the generator exclusively via raw retrieved text, they discard the rich and latent discriminative signals formed during the retrieval process, limiting the ability of the generator to weight evidence effectively.


\textbf{Latent Chain-of-Thought.} While Chain-of-Thought (CoT) methods~\cite{CoT, CoQ, CoVe, CoN} enhance reasoning transparency, they suffer from high generation latency and sensitivity to retrieved noise. Latent-CoT approaches~\cite{Coconut, CODI, Sim-CoT} address this by compressing intermediate reasoning into continuous latent vectors. However, because these continuous representations are generated only from autoregressive prediction, they are structurally isolated from external knowledge modules, heavily restricting their integration into RAG systems.

\textbf{Rank Collapse of LLMs.} As LLMs increase in depth, hidden token representations tend to converge directionally, concentrating within a narrow subspace. This phenomenon, known as rank collapse, reduces the effective rank and representational diversity of deep-layer tokens, impairing fine-grained semantic discrimination. Theoretical studies~\cite{rc1, rc2} demonstrate that this contraction accumulates across repeated self-attention layers. While architectural components like attention masks and LayerNorm can influence the geometry of this collapse~\cite{rc3}, the degradation is particularly detrimental in decoder-only Transformers, where it severely weakens token-level expressivity~\cite{rc4, rc5, rc6, rc7}. Despite this established limitation, recent LLM-based text encoders~\cite{NV-Embed, LLM2Vec} optimize for retrieval without explicitly mitigating the rank collapse of their pretrained backbones, hindering the formation of a semantically well-separated embedding space. This bottleneck becomes even more acute in unified LLMs~\cite{GritLM, OneGen}, where forcing generation and representation learning to share the same collapsed hidden space creates severe objective conflict.

\section{Method}

\begin{figure*}[t]
    \centering
    \includegraphics[width=1\textwidth]{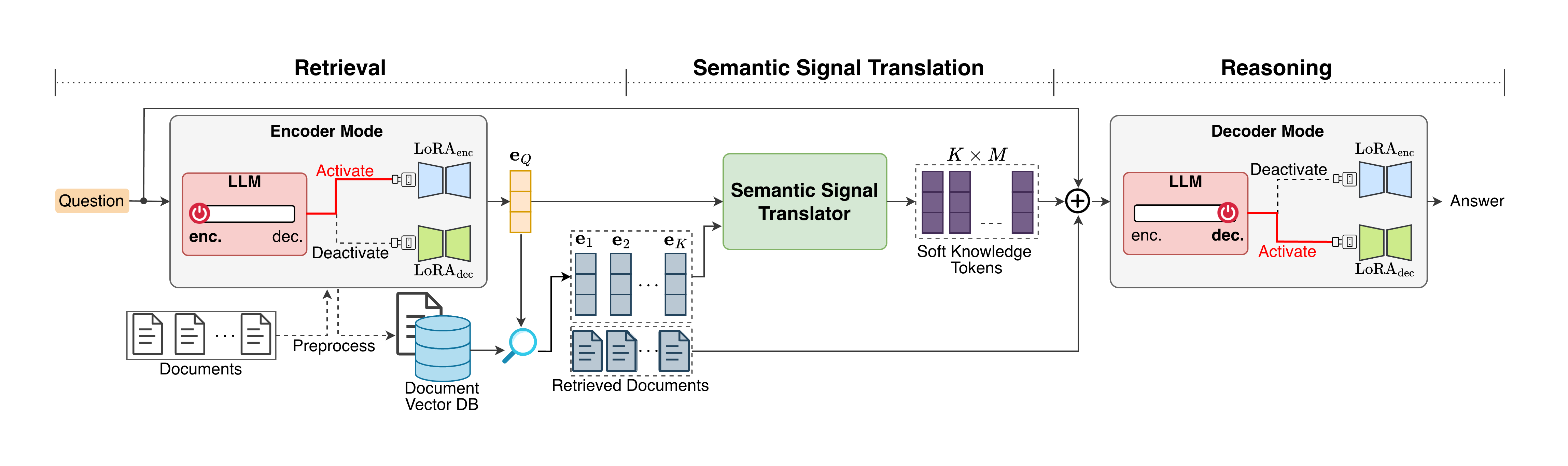}
    \caption{Illustration of the inference pipeline of ShiftRAG.}
    \label{fig:main_overview}
\vspace{-10pt}
\end{figure*}
To enable a single LLM to master both retrieval and reasoning, we fine-tune a pre-trained causal LLM with two task-specific LoRA adapters. In \textbf{encoder mode}, we relax the causal mask to allow bidirectional self-attention, adapting the model to capture deep question-document semantic relevance. In \textbf{decoder mode}, the LLM reverts to its native autoregressive decoding to generate answers, leveraging the discriminative insights gained from the encoder mode.

Figure \ref{fig:main_overview} illustrates the overall inference framework of ShiftRAG, which consists of three stages: retrieval, semantic signal translation, and reasoning. For offline indexing, the encoder-mode LLM maps every document into an embedding space. In the retrieval stage at inference, the encoder mode embeds the input question into this shared space to retrieve top-$K$ relevant documents. In the semantic signal translation stage, we then translate the question-document embedding pairs and their corresponding relevance scores into soft knowledge tokens. These latent guidance vectors bridge the gap between the discriminative encoder and the generative decoder by mapping retrieval signals directly into the token embedding space of the decoder. Finally, in the reasoning stage, the LLM is switched to decoder mode, generating a grounded answer based on the question, the retrieved document text, and the injected soft knowledge tokens.

\subsection{Gap-aware and Collapse-resilient Deep Encoder}\label{sec:train_retrieval}
\begin{figure*}[t]
    \centering
    \includegraphics[width=1\textwidth]{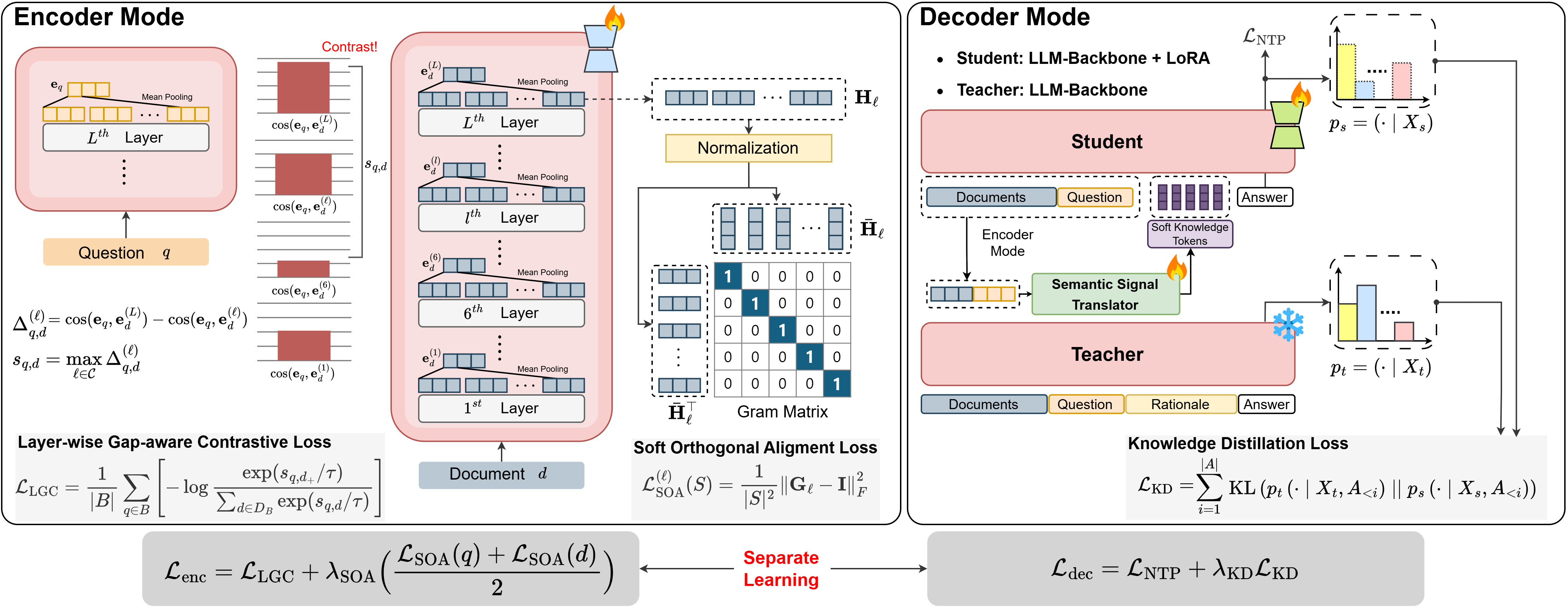}
    \caption{Overall framework of ShiftRAG which separately trains the encoder and decoder modes.}
    \label{fig:main_train_overview}
\vspace{-10pt}
\end{figure*}

To repurpose the decoder-only LLM as an encoder, we enable bidirectional self-attention (as illustrated in Figure \ref{fig:main_train_overview}), allowing the model to aggregate global context. To capture deep question-document relevance, we exploit the layer-wise representation gap (Section \ref{sec:3.2.1}), which captures the semantic maturation of the input from shallow to deep layers. To mitigate rank collapse, we introduce a soft orthogonality constraint that regularizes the hidden representations of all tokens within a sequence (Section \ref{sec:3.2.2}). This ensures that the model preserves discriminative feature diversity, preventing the `narrow cone' effect often seen in deep transformer layers.

\subsubsection{Layer-wise Gap-aware Contrastive Learning}\label{sec:3.2.1}
Prior research on intermediate-layer representations in LLM has revealed that layers encode representations hierarchically: lower layers extract lexical and syntactic surface features, while higher layers capture abstract context-driven semantics. DoLa~\cite{DoLa}, DSAS~\cite{DSAS}, and RaLa~\cite{CoopRAG} show that such layer-wise representational differences are closely related to reasoning and retrieval quality. From these observations, we propose layer-wise gap-aware contrastive (LGC) learning to fine-tune an LLM-based retriever. 


Formally, let $q$ and $d$ denote a question and a document, respectively. Following RaLa~\cite{CoopRAG} and DSAS~\cite{DSAS}, we partition the first half of the $L$ layers of the LLM into five  buckets, and randomly sample one layer from each. Let $\mathcal{C}$ denote the set of these selected candidate layers. For a given question $q$ and document $d$, we define the relevance gap between the final layer $L$ and any candidate layer $\ell \in \mathcal{C}$:
\begin{equation}\label{eq:gap}
\scalebox{1}{$
\Delta_{q,d}^{(\ell)}
=
\cos\left(\mathbf{e}_q,\mathbf{e}_d^{(L)}\right)
-
\cos\left(\mathbf{e}_q,\mathbf{e}_d^{(\ell)}\right)
$}
\end{equation}
where $\mathbf{e}_q$ is the final-layer mean-pooled representation of $q$, and $\mathbf{e}_d^{\left(\ell\right)}$ is the $\ell$-th layer mean-pooled representation of $d$. We define the relevance score between $q$ and $d$ as the maximum relevance gap across all candidate layers:
\begin{equation}\label{eq:relevance_score}
\scalebox{1}{$
s_{q,d}
=
\displaystyle\max_{\ell\in\mathcal{C}}
\Delta_{q,d}^{(\ell)}
$}
\end{equation}
The choice of the layer with the highest gap ensures the encoder focuses on the maximum representational shift. By doing so, we capture the critical point where the encoder transitions from processing lexical surface features to encoding the abstract relevance necessary to resolve the question. This allows the model to adaptively select the lower layer that best reveals the relevance signal for each pair, rather than using a fixed contrasting layer for all pairs.

For each mini-batch $B$ of questions during fine-tuning, let $D_B$ be a set of positive and hard negative documents for every question $q\in B$. The layer-wise gap-aware contrastive loss is defined below:
\begin{equation}\label{eq:loss_lgc}
\scalebox{1}{
$
\mathcal{L}_{\mathrm{LGC}}=\frac{1}{|B|}\sum_{q \in B}\left[-\log\frac{\exp\left(s_{q,d_+}/\tau\right)}{\sum_{d \in D_B}\exp\left(s_{q,d}/\tau\right)}\right]
$
}
\end{equation}
where $d_+$ is the positive document for question $q$, and $\tau$ is a temperature hyperparameter. This objective encourages the positive document to have a larger layer-wise gap score than negative documents, thereby forming a retrieval-oriented embedding space that reflects depth-wise semantic refinement rather than absolute similarity alone. 
\footnote{While RaLa~\cite{CoopRAG} applies layer contrast as a multiplicative weight over token-level average maximum similarity (MaxSim) for reranking, we directly use the gap-aware relevance score in encoder-mode training. This allows layer contrast to enter the retrieval objective as the score itself, rather than as an auxiliary reweighting term, while avoiding the time/memory overhead of MaxSim, as analyzed in Appendix \ref{sec:appendix_Complexity}.} 

For inference, we adopt the final-layer cosine similarity score, i.e., $s_{q, d} = \cos(\mathbf{e}_q,\mathbf{e}_d^{(L)})$, to minimize computational overhead. This allows the model to benefit from the discriminative representations formed during gap-aware training without the latency of multi-layer computation. Further analysis of this trade-off is available in Appendix \ref{sec:appendix_scoring_tradeoff}.

\subsubsection{Soft Orthogonal Alignment Loss}\label{sec:3.2.2}
LLM-based encoders may suffer from rank collapse~\cite{rc6,rc7} in deeper layers, where token representations shrink into a low-dimensional subspace. When this tendency becomes severe, embeddings of different tokens in the same sequence become aligned in similar directions, reducing representation diversity. As a result, the encoder may fail to preserve the fine-grained semantic differences required for retrieval, hindering the latent separation of relevant and irrelevant documents.

To alleviate rank collapse~\cite{rc8, rc9}, we introduce soft orthogonal alignment (SOA) loss, which penalizes excessive correlation among token representations. For an input token sequence $S$, let $\mathbf{H}_{\ell} \in \mathbb{R}^{\vert S \vert \times h}$ denote the hidden representations of all tokens in $S$ at layer $\ell$. We first apply $L_2$ normalization to the matrix row-wise, denoted as  $\bar{\mathbf{H}}_{\ell}$. We then compute the token-level Gram matrix $\mathbf{G}_{\ell}=\bar{\mathbf{H}}_{\ell}\bar{\mathbf{H}}_{\ell}^\top$, which serves as a pairwise cosine similarity matrix. In an ideal representation space in which tokens are mutually orthogonal, $\mathbf{G}_{\ell}$ approaches the identity matrix $\mathbf{I}$, with off-diagonal entries vanishing to 0 and diagonal entries remaining at 1. The SOA loss at layer $\ell$ is defined as the normalized squared Frobenius distance between $\mathbf{G}_{\ell}$ and $\mathbf{I}$: 
\begin{equation}\label{eq:loss_so1}
\scalebox{1}{
$
\mathcal{L}_{\text{SOA}}^{(\ell)} (S) = \frac{1}{\vert S \vert^2} \Vert \mathbf{G}_{\ell} - \mathbf{I} \Vert^2_F
$
}
\end{equation}
This loss softly prevents different tokens from collapsing into overly similar directions while preserving useful semantic correlations. We apply this regularization to the last $n$ layers, in which rank collapse is more likely to affect the sequence representations. The question-side and document-side multi-layer SOA losses are defined as:
\begin{equation}\label{eq:loss_so2}
\scalebox{1}{
$
\mathcal{L}_{\text{SOA}} (q) = \frac{1}{\vert B \vert} \displaystyle\sum_{q \in B} \left(\frac{1}{n} \displaystyle\sum_{\ell=L-n+1}^L \mathcal L_{\text{SOA}}^{(\ell)} (q) \right),\quad \mathcal{L}_{\text{SOA}} (d) = \frac{1}{\vert B \vert} \displaystyle\sum_{d \in D_B} \left(\frac{1}{n} \displaystyle\sum_{\ell=L-n+1}^L \mathcal L_{\text{SOA}}^{(\ell)} (d) \right)
$
}
\end{equation}
While existing regularizers~\cite{rc8, rc9} act on encoder weights or representation dimensionality, SOA regularizes the token representations to preserve representation diversity and retrieval-critical semantic discrimination.

\subsubsection{Training LLM-based Encoder}
We employ a set of LoRA parameters $\text{LoRA}_{\text{enc}}$ and optimize them using the joint objective:
\begin{equation}\label{eq:loss_final_enc}
\scalebox{1}{
$
\mathcal{L}_{\text{enc}} = \mathcal L_{\text{LGC}} + \lambda_{\text{SOA}} \left( \frac{\mathcal{L}_{\text{SOA}} (q) + \mathcal{L}_{\text{SOA}} (d)}{2} \right)
$
}
\end{equation}
where $\lambda_{\text{SOA}}$ is a hyperparameter. Following LLM2Vec, we initially perform a short MNTP warm-up to adapt $\text{LoRA}_\text{enc}$ to bidirectional attention. More details are provided in Appendix \ref{sec:appendix_ret_warm_up}.

\subsection{Semantic Signal Translation}\label{sec:train_generation}
\begin{figure*}[t]
    \centering
    \includegraphics[width=1\textwidth]{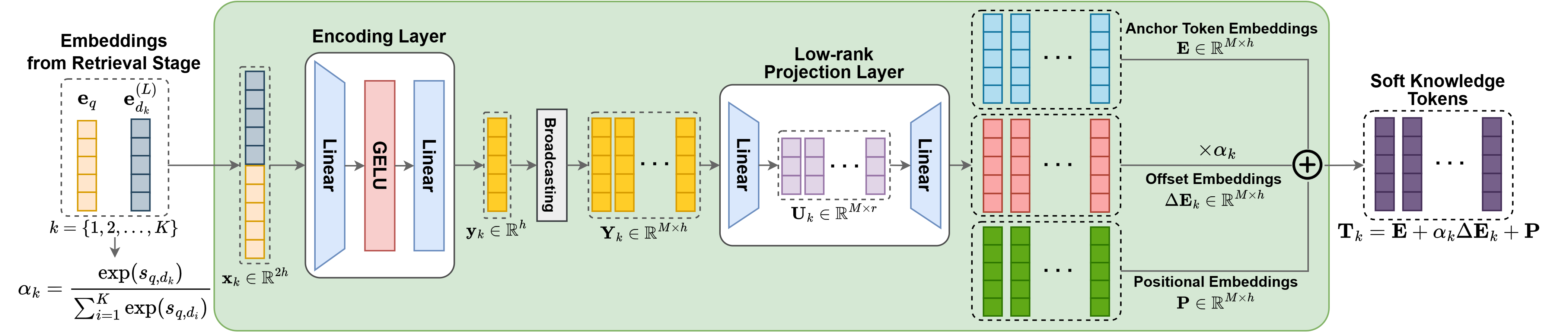}
    \caption{Illustration of Semantic Signal Translator (SST).}
    \label{fig:main_sst}
\vspace{-10pt}
\end{figure*}
Conventional RAG frameworks provide the generator with only raw document text while discarding semantic signals generated by the encoder. Consequently, the burden of determining question-document relevance falls entirely on autoregressive decoding of the generator. By exposing only explicit textual evidence, these methods prevent the LLM from utilizing implicit retrieval priors, e.g., the degree of semantic alignment or relative document importance. To bridge this gap, the semantic signal translator (SST) projects retrieval-stage embeddings into the token embedding space of the generator, effectively injecting discriminative knowledge into reasoning.

As shown in Figure \ref{fig:main_sst}, SST feeds the concatenation of $\mathbf{e}_q$ and $\mathbf{e}_{d_k}$ into the learnable encoding layer to obtain its hidden representation $\mathbf{y}_k\in \mathbb{R}^h$, which is then broadcast into $\mathbf{Y}_k\in \mathbb{R}^{M \times h}$. A learnable low-rank projection layer maps $\mathbf{Y}_k$ to a low-rank matrix $\mathbf{U}_k\in \mathbb{R}^{M \times r}$, where $r$ is the bottleneck dimension for low-rank signal translation, enabling efficient transfer of retrieval-side discriminative knowledge into the token embedding space. Next, we compute offset embeddings $\Delta \mathbf{E}_k \in \mathbb{R}^{M \times h}$ and scale them by the normalized relevance weight $\alpha_k$, computed from similarities over the top-$K$ documents. For each document $d_k$, we compute soft knowledge tokens $\mathbf{T}_k=\mathbf{E}+\alpha_k \Delta \mathbf{E}_k+\mathbf{P}$, where $\mathbf{P}\in\mathbb{R}^{M \times h}$ denotes learnable positional embeddings, and $\mathbf{E}\in \mathbb{R}^{M \times h}$ represents learnable anchor embeddings initialized as pretrained latent vectors \cite{SPRING}. This initialization is detailed in Appendix \ref{sec:appendix_intialization} and analyzed in Appendix \ref{sec:appendix_anchor_initalization}. The anchor embeddings provide vocabulary-aligned references to transform the question-document representations from the LLM encoder into generation-friendly embeddings. Finally, we concatenate the soft knowledge tokens $\mathbf{T}_k$ for top-$K$ documents to form $\mathbf{T}_{1:K}$.

\subsection{Soft Knowledge-guided Reasoning}\label{sec:reasoning}
In the reasoning stage, the LLM generates a grounded answer by integrating retrieved document texts with their corresponding soft knowledge tokens. To bridge the gap between retrieval-side discriminative signals and generative reasoning, we optimize the decode-mode adapters $\text{LoRA}_{\text{dec}}$ via a dual objective: (1) rationale-conditioned self-distillation, where a student LLM with $\text{LoRA}_{\text{dec}}$ learns to mimic the output distribution of the same frozen LLM backbone conditioned on explicit rationales; and (2) next-token prediction (NTP) on the target answer to preserve linguistic coherence and factual grounding. As illustrated in Figure \ref{fig:main_train_overview}, this configuration ensures that the generator treats the soft knowledge tokens as a latent blueprint for the final response.




For objective (1), we employ knowledge distillation (KD) to encode explicit reasoning signals into the soft knowledge tokens $\mathbf{T}_{1:K}$. We formulate the teacher and student contexts prior to answer generation:
\begin{equation}\label{eq:note_kd1}
\scalebox{1}{
$
\begin{aligned}
X_t &= [d_{1:K}, q, \textit{rationale}], \quad
X_s &= [d_{1:K}, q, \mathbf{T}_{1:K}]
\end{aligned}
$
}
\end{equation}

where \textit{rationale} denotes a single-sentence textual rationale (see Appendix \ref{sec:appendix_prompt1}) pre-constructed  for teacher-side supervision. While the teacher utilizes explicit document-level reasoning signals, the student learns to approximate these in $\mathbf{T}_{1:K}$. For a ground-truth answer $A$, the KD loss aligns the student and teacher distributions at each token position $i$:
\begin{equation}\label{eq:loss_kd}
\scalebox{1}{
$
\mathcal{L}_{\text{KD}}=\displaystyle\sum^{\vert A\vert}_{i=1}\text{KL}\left(p_t \left(\cdot\mid X_t,A_{<i}\right) \mid\mid p_s \left(\cdot\mid X_s,A_{<i}\right)\right)
$
}
\end{equation}
where $A_{<i}$ is the answer prefix. Consequently, the student internalizes rationale-guided reasoning without taking text rationales as input. We show the impact of soft token placement in Appendix \ref{appendix_analysis_placement}.

The final objective for the decoder is a weighted sum of the standard NTP loss and the KD loss:
\begin{equation}\label{eq:loss_final_gen}
\scalebox{1}{
$
\mathcal{L}_{\text{dec}} = \mathcal{L}_{\text{NTP}} + \lambda_{\text{KD}}\mathcal{L}_{\text{KD}}
$
}
\end{equation}
where $\mathcal{L}_{\text{KD}}$ transfers rationale-induced reasoning signals into $\mathbf{T}_{1:K}$, $\mathcal{L}_{\text{NTP}}$ supervises generation of the ground-truth answer, and $\lambda_{\text{KD}}$ is a hyperparameter. The prompt used for decoder-mode training is provided in Appendix \ref{sec:appendix_prompt3}.

\section{Experiments}
\subsection{Experimental Setup}\label{sec:experiment_setup}
We evaluate our framework mainly on five datasets: NaturalQuestions~\cite{NaturalQuestions} for single-hop QA, HotpotQA~\cite{HotpotQA}, 2WikiMultihopQA~\cite{2WikiMultihopQA}, and MuSiQue~\cite{MuSiQue} for multi-hop QA, and SimpleQA~\cite{SimpleQA} for factual QA. Each dataset contains 1,000 test questions. For the multi-hop QA benchmarks, we follow the test split configuration of RaLa~\cite{CoopRAG} and compare all retrieval-based baselines under identical conditions. We optimize the encoder-mode LoRA adapter ($r=16$) on the E5~\cite{E5} dataset by following LLM2Vec~\cite{LLM2Vec}. The decoder-mode LoRA adapter is trained with a higher rank ($r=64$) on a hop-aware mixture constructed from both single- and multi-hop QA benchmarks. We use Gemma2-9B~\cite{Gemma2}, LlaMA3.1-8B~\cite{LlaMA3}, Mistral-7B~\cite{Mistral}, and Qwen2.5-7B~\cite{Qwen2} as backbones, apply the SOA loss to the last four layers, and use five soft knowledge tokens for each question–document pair. 
Further details regarding the experimental setup and baselines are described in Appendix \ref{sec:appendix_exsetup}.

\subsection{Main Results}\label{sec:main_result_retrieval}
\begingroup
  \renewcommand{\arraystretch}{0.80}
  \setlength{\tabcolsep}{3pt}

  \begin{table*}[h]
    \centering
    \footnotesize
    \caption{Retrieval performance. The best and second-best performances are presented in bold and underlined, respectively. LLM2Vec, OneGen, and ShiftRAG use Gemma2-9B as the backbone.}
    \label{tab:retrieval_main}
    \begin{tabular*}{\textwidth}{@{\extracolsep{\fill}}lccccccccc}
      \toprule
      \textbf{Methods}
      & \multicolumn{3}{c}{\textbf{HotpotQA}}
      & \multicolumn{3}{c}{\textbf{2WikiMultihopQA}}
      & \multicolumn{3}{c}{\textbf{MuSiQue}} \\
      \cmidrule(lr){2-4} \cmidrule(lr){5-7} \cmidrule(l){8-10}
      &
      R@2 & R@5 & R@10
      & R@2 & R@5 & R@10
      & R@2 & R@5 & R@10 \\
      \midrule
      LLM2Vec
      & 0.6130 & 0.7854 & 0.8612
      & 0.5845 & 0.6451 & 0.6829
      & 0.3598 & 0.5126 & 0.5977 \\
      NV-Embed-v2
      & \underline{0.6318} & \underline{0.7982} & \underline{0.8736}
      & \underline{0.5956} & \underline{0.6632} & \underline{0.7202}
      & \underline{0.3781} & \underline{0.5290} & \underline{0.6236} \\
      OneGen
      & 0.5792 & 0.7450 & 0.8402
      & 0.5584 & 0.6077 & 0.6392
      & 0.3320 & 0.4777 & 0.5658 \\
      GritLM
      & 0.5890 & 0.7487 & 0.8085
      & 0.5558 & 0.6311 & 0.6977
      & 0.3760 & 0.4990 & 0.5983 \\
      \midrule
      ShiftRAG
      & \textbf{0.6598} & \textbf{0.8214} & \textbf{0.8806}
      & \textbf{0.6080} & \textbf{0.7109} & \textbf{0.7530}
      & \textbf{0.4186} & \textbf{0.5634} & \textbf{0.6516} \\
      \bottomrule
    \end{tabular*}
  \end{table*}
  \vspace{-10pt}
\endgroup
\begingroup
  \renewcommand{\arraystretch}{0.80}
  \setlength{\tabcolsep}{3pt}
  \begin{table*}[h]
    \centering
    \footnotesize
    \caption{Question answering performance of ShiftRAG and baselines. The best and second-best performances are presented in bold and underlined, respectively.}
    \label{tab:qa_main}
    \begin{tabular*}{\textwidth}{@{\extracolsep{\fill}}lcccccc}
      \toprule
      \textbf{Methods}
      & \multicolumn{2}{c}{\textbf{HotpotQA}}
      & \multicolumn{2}{c}{\textbf{2WikiMultihopQA}}
      & \multicolumn{2}{c}{\textbf{MuSiQue}} \\
      \cmidrule(lr){2-3} \cmidrule(lr){4-5} \cmidrule(l){6-7}
      &
      EM & F1
      & EM & F1
      & EM & F1 \\
      \midrule
      LLM2Vec
      & 0.5030 & 0.6106
      & 0.4400 & 0.4914
      & 0.2430 & 0.3670 \\
      NV-Embed-v2
      & \underline{0.5140} & \underline{0.6220}
      & \underline{0.4580} & \underline{0.5064}
      & \underline{0.2480} & \underline{0.3783} \\
      GritLM
      & 0.4430 & 0.5258
      & 0.3020 & 0.3665
      & 0.1870 & 0.2872 \\
      OneGen
      & 0.4670 & 0.5869
      & 0.4220 & 0.4785
      & 0.2200 & 0.3342 \\
      \midrule
      ShiftRAG
      & \textbf{0.5610} & \textbf{0.6790}
      & \textbf{0.5080} & \textbf{0.5528}
      & \textbf{0.3050} & \textbf{0.4129} \\
      \bottomrule
    \end{tabular*}
  \end{table*}
\endgroup
\textbf{Retrieval Performance.} Table \ref{tab:retrieval_main} compares the retrieval performance of ShiftRAG with state-of-the-art LLM-based encoder and unified LLM baselines. ShiftRAG consistently achieves the best performance across all datasets and all metrics (full results for all baselines are provided in Appendix \ref{sec:appendix_result_retrieval}). Compared to NV-Embed-v2, i.e., the strongest baseline, ShiftRAG improves Recall@2 by 4.43\% on HotpotQA (0.6318 → 0.6598) and 10.71\% on MuSiQue (0.3781 → 0.4186). Compared against GritLM, i.e., a strong unified LLM baseline, ShiftRAG improves Recall@2 by 12.02\% on HotpotQA (0.5890 → 0.6598) and 11.33\% on MuSiQue (0.3760 → 0.4186). These results show that encoder-mode training forms a more discriminative embedding space than existing LLM-based encoders and unified models. We provide generalization results on NaturalQuestions and MTEB in Appendices \ref{sec:appendix_single-hopQA} and \ref{sec:appendix_general_tasks}, respectively, and analyze scalability in Appendix \ref{sec:appendix_analysis_Scalability}.

\textbf{Question Answering Performance.} Table \ref{tab:qa_main} summarizes the QA performance of ShiftRAG against baselines. To ensure a controlled evaluation of generation capacity, all retrieval-only baselines are paired with Gemma2-9B, the strongest generator among the backbone LLMs. ShiftRAG achieves the highest EM and F1 scores across all datasets (the complete QA results are provided in Appendix \ref{sec:appendix_result_generation}). Compared to NV-Embed-v2, ShiftRAG improves EM by 9.14\% on HotpotQA (0.5140 → 0.5610) and 22.98\% on MuSiQue (0.2480 → 0.3050). Compared with OneGen under the same backbone setting, ShiftRAG improves EM by 20.13\% on HotpotQA (0.4670 → 0.5610) and 38.64\% on MuSiQue (0.2200 → 0.3050). These results demonstrate that translating retrieval-side discriminative knowledge into soft knowledge tokens provides effective reasoning signals that raw text alone cannot provide. Additional generalization experiments on  MMLU~\cite{MMLU}, GSM8K~\cite{GSM8K}, and TyDiQA~\cite{TyDiQA} follow the protocol of GritLM~\cite{GritLM}. The complete results on NaturalQuestions, SimpleQA, and general tasks  
are provided in Appendices \ref{sec:appendix_single-hopQA}, \ref{sec:appendix_simple_qa}, and \ref{sec:appendix_general_tasks}, respectively.



\subsection{Ablation Study}\label{sec:ablation1}
\begingroup
  \renewcommand{\arraystretch}{0.825}
  \setlength{\tabcolsep}{3pt}

  \begin{table*}[h]
    \centering
    \footnotesize
    \caption{Ablation study on encoder-mode training losses. The best and second-best performances are presented in bold and underlined, respectively.}
    \label{tab:ablation_retrieval_loss}
    \begin{tabular*}{\textwidth}{@{\extracolsep{\fill}}lccccccccc}
      \toprule
      \textbf{Category}
      & \multicolumn{3}{c}{\textbf{HotpotQA}}
      & \multicolumn{3}{c}{\textbf{2WikiMultihopQA}}
      & \multicolumn{3}{c}{\textbf{MuSiQue}} \\
      \cmidrule(lr){2-4} \cmidrule(lr){5-7} \cmidrule(l){8-10}
      &
      R@2 & R@5 & R@10
      & R@2 & R@5 & R@10
      & R@2 & R@5 & R@10 \\
      \midrule
      ShiftRAG
      & \textbf{0.6598} & \textbf{0.8214} & \textbf{0.8806}
      & \textbf{0.6080} & \textbf{0.7109} & \textbf{0.7530}
      & \textbf{0.4186} & \textbf{0.5634} & \textbf{0.6516} \\
      \textit{w/o} SOA
      & \underline{0.6535} & \underline{0.8021} & \underline{0.8678}
      & \underline{0.5998} & \underline{0.6984} & \underline{0.7425}
      & \underline{0.4045} & \underline{0.5511} & \underline{0.6493} \\
      \textit{w/o} LGC
      & 0.6360 & 0.7892 & 0.8445
      & 0.5816 & 0.6917 & 0.7084
      & 0.4008 & 0.5457 & 0.6140 \\
      \textit{w/o} Both
      & 0.6120 & 0.7800 & 0.8352
      & 0.5711 & 0.6824 & 0.7005
      & 0.3916 & 0.5382 & 0.6051 \\
      \bottomrule
    \end{tabular*}
  \end{table*}
\endgroup
We conduct an ablation study to evaluate the impact of the encoder-mode training objectives. Table \ref{tab:ablation_retrieval_loss} reports the retrieval performance for three training setups: (a) \textit{w/o} SOA removes SOA loss while retaining LGC; (b) \textit{w/o} LGC replaces the LGC loss with standard InfoNCE loss~\cite{InfoNCE} while retaining SOA, and (c) \textit{w/o} Both optimizes the encoder solely using  standard InfoNCE. \textit{w/o} Both yields the largest degradation.
Between the two components, the LGC loss drives the majority of the performance gains; \textit{w/o} LGC reduces Recall@10 on MuSiQue by 5.77\%. Removing SOA results in a marginal but consistent reduction.
As a result, while exploiting layer-wise semantic progression (LGC) provides a powerful relevance signal, suppressing excessive token correlations (SOA) acts as an effective complementary regularizer to preserve fine-grained semantic discrimination.
\begingroup
  \renewcommand{\arraystretch}{0.825}
  \setlength{\tabcolsep}{3pt}

  \begin{table*}[h]
    \centering
    \footnotesize
    \caption{Ablation study on decoder-mode training components. The best and second-best performances are presented in bold and underlined, respectively.}
    \label{tab:ablation_generation_components}
    \begin{tabular*}{\textwidth}{@{\extracolsep{\fill}}lcccccc}
      \toprule
      \textbf{Category}
      & \multicolumn{2}{c}{\textbf{HotpotQA}}
      & \multicolumn{2}{c}{\textbf{2WikiMultihopQA}}
      & \multicolumn{2}{c}{\textbf{MuSiQue}} \\
      \cmidrule(lr){2-3} \cmidrule(lr){4-5} \cmidrule(l){6-7}
      &
      EM & F1
      & EM & F1
      & EM & F1 \\
      \midrule
      ShiftRAG
      & \textbf{0.5610} & \textbf{0.6790}
      & \textbf{0.5080} & \textbf{0.5528}
      & \textbf{0.3050} & \textbf{0.4129} \\
      \textit{w/o} SST
      & 0.5230 & 0.6468
      & 0.4630 & 0.5194
      & 0.2450 & 0.3639 \\
      \textit{w/o} $\text{LoRA}_{\text{dec}}$
      & \underline{0.5300} & \underline{0.6570}
      & \underline{0.4720} & \underline{0.5212}
      & \underline{0.2650} & \underline{0.3883} \\
      \textit{w/o} Both
      & 0.5020 & 0.6253
      & 0.4400 & 0.4980
      & 0.1970 & 0.2979 \\
      \bottomrule
    \end{tabular*}
  \end{table*}
\endgroup
We investigate the impact of individual components for decoder mode. Table \ref{tab:ablation_generation_components} shows QA performance under four settings: \textit{w/o} SST removes SST and trains only $\text{LoRA}_{\text{dec}}$; \textit{w/o} $\text{LoRA}_{\text{dec}}$ retains SST but removes generator-side LoRA tuning; \textit{w/o} Both uses the backbone without decoder-mode training. 
Between the two components, SST contributes more to QA performance: \textit{w/o} SST reduces MuSiQue EM by 19.67\% ($0.3050 \to 0.2450$), whereas \textit{w/o} $\text{LoRA}_{\text{dec}}$ reduces it by 13.11\% ($0.3050 \to 0.2650$). These results indicate that translating retrieval-side signals into soft knowledge tokens is more effective than fine-tuning the generator only with retrieved documents. We further conduct a detailed SST component ablation in Appendix \ref{sec:appendix_ablation_SST}. We analyze how SST affects confidence calibration and generator-side LoRA adaptation in Appendix \ref{sec:appendix_SST_confidence}.

\subsection{Impact of Learning Strategies}\label{sec:latency}
\begingroup
  \renewcommand{\arraystretch}{0.80}
  \setlength{\tabcolsep}{3pt}

  \begin{table*}[h]
    \centering
    \footnotesize
    \caption{Performance and training time comparisons across different learning strategies and parameter sharing configurations. The best and second-best scores are bolded and underlined, respectively.}
    \label{tab:training_methods_analysis}
    \begin{tabular*}{\textwidth}{@{\extracolsep{\fill}}llccccccccc}
      \toprule
      &
      &
      \multicolumn{4}{c}{\textbf{HotpotQA}}
      & \multicolumn{4}{c}{\textbf{2WikiMultihopQA}}
      & \\
      \cmidrule(lr){3-6} \cmidrule(lr){7-10}
      \multicolumn{1}{c}{\textbf{Training method}}
      & \multicolumn{1}{c}{\textbf{Parameter}}
      & \multicolumn{2}{c}{\textbf{Retrieval}}
      & \multicolumn{2}{c}{\textbf{QA}}
      & \multicolumn{2}{c}{\textbf{Retrieval}}
      & \multicolumn{2}{c}{\textbf{QA}}
      & \textbf{Time (h)} \\
      \cmidrule(lr){3-4} \cmidrule(lr){5-6}
      \cmidrule(lr){7-8} \cmidrule(lr){9-10}
      &
      &
      R@2 & R@5
      & EM & F1
      & R@2 & R@5
      & EM & F1
      & \\
      \midrule
      \multirow{2}{*}{Joint learning}
      & shared
      & 0.4896 & 0.6298
      & 0.1300 & 0.3430
      & 0.3230 & 0.4666
      & 0.1270 & 0.3432
      & \underline{21} \\
      & separate
      & \underline{0.5264} & \underline{0.6865}
      & \underline{0.4730} & \underline{0.5885}
      & \underline{0.5159} & \underline{0.6184}
      & \underline{0.4360} & \underline{0.5115}
      & 23 \\
      \midrule
      Separate learning
      & separate
      & \textbf{0.6598} & \textbf{0.8214}
      & \textbf{0.5610} & \textbf{0.6790}
      & \textbf{0.6080} & \textbf{0.7109}
      & \textbf{0.5080} & \textbf{0.5528}
      & \textbf{10} \\
      \bottomrule
    \end{tabular*}
  \end{table*}
  \vspace{-5pt}
\endgroup

Table \ref{tab:training_methods_analysis} shows the impact of joint learning versus separate learning strategies. Joint learning with shared parameters exhibits severe optimization instability, leading to a large degradation in QA performance, e.g., EM drops precipitously to 0.1300 on HotpotQA and 0.1270 on 2WikiMultihopQA. Our analysis reveals that the encoder and decoder objectives produce negatively aligned gradients on shared LoRA parameters in 84\% of updates. This frequent gradient conflict indicates that the retrieval task and the reasoning task possess fundamentally competing optimization directions. While utilizing separate parameters for each mode mitigates this degradation, simultaneous joint optimization remains suboptimal and computationally expensive. In contrast, separate learning using separate LoRA parameters demonstrates superior efficacy and efficiency. 
These results support our hypothesis that decoupling discriminative retrieval learning from autoregressive generation allows each mode to converge toward its respective optimum without mutual interference.

\subsection{Impact of Semantic Signal Translator}\label{sec:appendix_SST_TextEvidence}
\begingroup
  \renewcommand{\arraystretch}{0.80}
  \setlength{\tabcolsep}{3pt}

  \begin{table*}[h]
    \centering
    \footnotesize
    \caption{Impact of SST-generated soft knowledge tokens on QA performance. The best and second-best performances are presented in bold and underlined, respectively.}
    \label{tab:SST_analysis}
    \begin{tabular*}{\textwidth}{@{\extracolsep{\fill}}lcccccc}
      \toprule
      \textbf{Category}
      & \multicolumn{2}{c}{\textbf{HotpotQA}}
      & \multicolumn{2}{c}{\textbf{2WikiMultihopQA}}
      & \multicolumn{2}{c}{\textbf{MuSiQue}} \\
      \cmidrule(lr){2-3} \cmidrule(lr){4-5} \cmidrule(l){6-7}
      &
      EM & F1
      & EM & F1
      & EM & F1 \\
      \midrule
      ShiftRAG
      & \textbf{0.5610} & \textbf{0.6790}
      & \textbf{0.5080} & \textbf{0.5528}
      & \textbf{0.3050} & \textbf{0.4129} \\
      \textit{Textual Evidence}
      & \underline{0.5330} & \underline{0.6574}
      & \underline{0.4790} & \underline{0.5226}
      & \underline{0.2650} & \underline{0.3814} \\
      \textit{w/o} SST
      & 0.5230 & 0.6468
      & 0.4630 & 0.5194
      & 0.2450 & 0.3639 \\
      \bottomrule
    \end{tabular*}
  \end{table*}
\endgroup

We investigate whether the semantic information encapsulated in our soft knowledge tokens can be reduced to simple textualized relevance signals. Table \ref{tab:SST_analysis} compares three settings: (1) ShiftRAG; (2) \textit{Textual Evidence}, which replaces soft tokens from SST with an explicit hard prompt verbalizing the relevance score of the retrieved documents; and (3) \textit{w/o} SST, which removes the soft tokens. While \textit{Textual Evidence} consistently outperforms the \textit{w/o} SST setting, confirming that the generator benefits from knowing retrieval-side relevance, ShiftRAG achieves higher performance on all datasets, improving MuSiQue EM by 15.09\% ($0.2650 \to 0.3050$) and F1 by 8.26\% ($0.3814 \to 0.4129$) over the textual baseline. These results demonstrate that soft tokens do not merely act as verbalized scores; instead, they serve as an effective interface that transfers complex, high-dimensional discriminative signals that are lost during textualization.

\subsection{Geometric Intuition for Efficacy of Semantic Signal Translator}
\vspace{-10pt}
\begin{figure*}[h]
    \centering
    \includegraphics[width=1\textwidth]{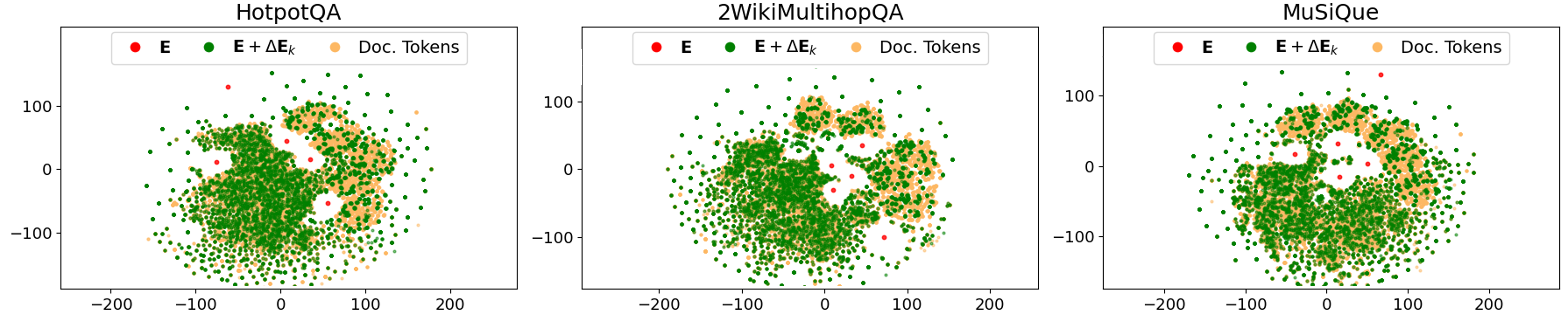}
    \caption{$t$-SNE visualization of anchor token embeddings (\textcolor{red}{red}), anchor token embeddings + offset embeddings (\textcolor{green}{green}), and original document token embeddings (\textcolor{orange}{orange}) for each benchmark.}
    \label{fig:main_tSNE}
\vspace{-5pt}
\end{figure*}
Figure \ref{fig:main_tSNE} visualizes the anchor token embeddings $\mathbf{E}$, the translated embeddings $\mathbf{E}+\Delta\mathbf{E}_k$, and the decoder-mode document token embeddings (\textit{Doc. Tokens}) in a shared $t$-SNE space. The anchor embeddings $\mathbf{E}$ form a localized set of reference points that are poorly aligned with the distribution of document tokens. However, upon incorporating the offset embeddings $\Delta\mathbf{E}_k$, the resultant vectors migrate toward the document token clusters, exhibiting a substantial distributional overlap across all datasets. This shift suggests that $\Delta\mathbf{E}_k$ effectively projects the anchor-token basis into a vocabulary-aligned region that is semantically congruent with the retrieved evidence. Consequently, the soft knowledge tokens enable SST to infuse the generator with retrieval-side knowledge effectively. This visualization confirms that our anchor-plus-translation design serves as a robust bridge between the retriever’s latent space and the generator's reasoning manifold.

\subsection{Inference Latency}\label{sec:appendix_latency_inferece}

\begin{table}[h]
  \centering
  \setlength{\tabcolsep}{6pt}
  \renewcommand{\arraystretch}{0.80}
  \footnotesize
  \caption{End-to-end inference latency, peak VRAM usage, and QA performance on HotpotQA.}
  \label{tab:appendix_latency_inference}
  \begin{tabular}{lccc}
    \toprule
    \textbf{Methods} & \multicolumn{3}{c}{\textbf{Inference Latency}} \\
    \cmidrule(lr){2-4}
    & Time (s) & VRAM (GiB) & EM \\
    \midrule
    NV-Embed-v2  & 8.72 & 38.82 & 0.514 \\
    GritLM       & 2.42 & 20.25 & 0.443 \\
    ShiftRAG         & 2.79 & 24.14 & 0.561 \\
    \bottomrule
  \end{tabular}
\end{table}

We measure inference latency to evaluate the runtime cost of SST-based generation. Table \ref{tab:appendix_latency_inference} shows end-to-end latency, peak VRAM usage, and QA performance on HotpotQA. ShiftRAG requires 2.79s per example and 24.14 GiB of VRAM, while achieving the best EM of 0.561. The latency breakdown is 0.13s for question encoding, 0.02s for Faiss retrieval, 0.0007s for soft knowledge token generation, and 2.64s for answer generation. This breakdown confirms that the primary computational bottleneck resides in the LLM decoding process, while the overhead introduced by retrieval and SST remains negligible. Compared with GritLM~\cite{GritLM}, ShiftRAG is slightly slower (2.42s → 2.79s) and uses more VRAM (20.25 → 24.14 GiB), but improves EM by 26.64\% (0.443 → 0.561). Compared with NV-Embed-v2~\cite{NV-Embed}, ShiftRAG is faster (8.72s → 2.79s), uses less VRAM (38.82 → 24.14 GiB), and improves EM by 9.14\% (0.514 → 0.561). These results indicate that ShiftRAG provides a favorable inference-time trade-off between efficiency and QA accuracy. The same trend is observed on 2WikiMultihopQA and MuSiQue. Training latency analysis is provided in Appendix \ref{sec:appendix_latency_training}.

\subsection{Impact of the Number of Soft Knowledge Tokens}
\vspace{-10pt}
\begin{figure*}[h]
    \centering
    \includegraphics[width=1\textwidth]{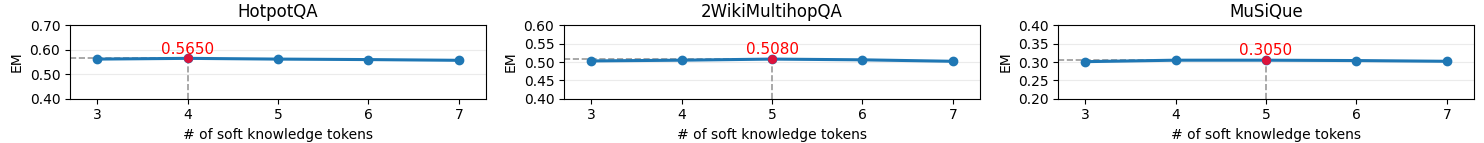}
    \caption{Impact of the number of soft knowledge tokens on QA performance.}
    \label{fig:main_NumSTs}
\vspace{-5pt}
\end{figure*}
We analyze the sensitivity of QA performance to the number $M$ of soft knowledge tokens per document. As shown in Figure \ref{fig:main_NumSTs}, the performance remains stable across the range of 3 to 7 tokens. 
While the optimal configuration varies slightly across datasets, the maximum variation is a negligible 1.43\%. Hence, we fix $M$ to 5 for all experiments. Appendix \ref{sec:appendix_hyperparameter} shows other hyperparameter studies.

\section{Conclusion and Limitation}\label{sec:conclusion}
We introduce ShiftRAG, a decoupled RAG framework that isolates discriminative retrieval learning from autoregressive generation. By proposing the semantic signal translator (SST), we successfully shift implicit high-dimensional retrieval alignments into latent soft knowledge tokens, providing an effective bridge between the retriever and generator. Furthermore, the soft orthogonality objective effectively mitigates rank collapse, while layer-wise gap-aware contrastive learning preserves the fine-grained latent topology required for robust evidence discrimination. 
Empirical evaluations demonstrate that ShiftRAG achieves superior retrieval accuracy and downstream QA quality.

Despite its strong empirical performance, our framework presents two limitations. Unlike explicit CoT pipelines, the soft knowledge tokens are continuous and inherently uninterpretable. Because our architecture relies on extracting and translating internal embedding states, it requires white-box access to the model weights; it cannot be applied to black-box LLMs.

\acksection
This work was partly supported by  Institute of Information \& communications Technology Planning \& Evaluation (IITP) grant funded by the Korea government (MSIT) (No. RS-2020-II201373, Artificial Intelligence Graduate School Program(Hanyang University)), by the Institute of Information \& Communications Technology Planning \& Evaluation (IITP) grant funded by the Korea government (MSIT, Ministry of Science and ICT) (No. RS-2026-25548925), and by the Basic Science Research Program through the National Research Foundation of Korea (NRF) funded by the Ministry of Education (No. RS-2025-25424679).

\newpage
\small
\bibliography{neurips_2026}

@misc{GPT-5-nano,
  title={{OpenAI GPT-5 System Card}},
  author={Aaditya Singh and Adam Fry and Adam Perelman and others},
  year={2025},
  eprint={2601.03267},
  archivePrefix={arXiv},
  primaryClass={cs.CL},
  url={https://arxiv.org/abs/2601.03267}
}

@inproceedings{RAG,
author = {Lewis, Patrick and Perez, Ethan and Piktus, Aleksandra and Petroni, Fabio and Karpukhin, Vladimir and Goyal, Naman and K\"{u}ttler, Heinrich and Lewis, Mike and Yih, Wen-tau and Rockt\"{a}schel, Tim and Riedel, Sebastian and Kiela, Douwe},
title = {Retrieval-augmented generation for knowledge-intensive NLP tasks},
year = {2020},
isbn = {9781713829546},
publisher = {Curran Associates Inc.},
address = {Red Hook, NY, USA},
booktitle = {Proceedings of the 34th International Conference on Neural Information Processing Systems},
articleno = {793},
numpages = {16},
location = {Vancouver, BC, Canada},
series = {NIPS '20}
}

@inproceedings{DPR,
    title = "Dense Passage Retrieval for Open-Domain Question Answering",
    author = "Karpukhin, Vladimir  and
      Oguz, Barlas  and
      Min, Sewon  and
      Lewis, Patrick  and
      Wu, Ledell  and
      Edunov, Sergey  and
      Chen, Danqi  and
      Yih, Wen-tau",
    editor = "Webber, Bonnie  and
      Cohn, Trevor  and
      He, Yulan  and
      Liu, Yang",
    booktitle = "Proceedings of the 2020 Conference on Empirical Methods in Natural Language Processing (EMNLP)",
    month = nov,
    year = "2020",
    address = "Online",
    publisher = "Association for Computational Linguistics",
    url = "https://aclanthology.org/2020.emnlp-main.550/",
    doi = "10.18653/v1/2020.emnlp-main.550",
    pages = "6769--6781"
}

@inproceedings{Contriever,
    title = "Unsupervised Dense Retrieval with Relevance-Aware Contrastive Pre-Training",
    author = "Lei, Yibin  and
      Ding, Liang  and
      Cao, Yu  and
      Zan, Changtong  and
      Yates, Andrew  and
      Tao, Dacheng",
    editor = "Rogers, Anna  and
      Boyd-Graber, Jordan  and
      Okazaki, Naoaki",
    booktitle = "Findings of the Association for Computational Linguistics: ACL 2023",
    month = jul,
    year = "2023",
    address = "Toronto, Canada",
    publisher = "Association for Computational Linguistics",
    url = "https://aclanthology.org/2023.findings-acl.695/",
    doi = "10.18653/v1/2023.findings-acl.695",
    pages = "10932--10940"
}

@inproceedings{GTR,
    title = "Large Dual Encoders Are Generalizable Retrievers",
    author = "Ni, Jianmo  and
      Qu, Chen  and
      Lu, Jing  and
      Dai, Zhuyun  and
      Hernandez Abrego, Gustavo  and
      Ma, Ji  and
      Zhao, Vincent  and
      Luan, Yi  and
      Hall, Keith  and
      Chang, Ming-Wei  and
      Yang, Yinfei",
    editor = "Goldberg, Yoav  and
      Kozareva, Zornitsa  and
      Zhang, Yue",
    booktitle = "Proceedings of the 2022 Conference on Empirical Methods in Natural Language Processing",
    month = dec,
    year = "2022",
    address = "Abu Dhabi, United Arab Emirates",
    publisher = "Association for Computational Linguistics",
    url = "https://aclanthology.org/2022.emnlp-main.669/",
    doi = "10.18653/v1/2022.emnlp-main.669",
    pages = "9844--9855"
}

@inproceedings{ColBERT,
author = {Khattab, Omar and Zaharia, Matei},
title = {ColBERT: Efficient and Effective Passage Search via Contextualized Late Interaction over BERT},
year = {2020},
isbn = {9781450380164},
publisher = {Association for Computing Machinery},
address = {New York, NY, USA},
url = {https://doi.org/10.1145/3397271.3401075},
doi = {10.1145/3397271.3401075},
booktitle = {Proceedings of the 43rd International ACM SIGIR Conference on Research and Development in Information Retrieval},
pages = {39–48},
numpages = {10},
location = {Virtual Event, China},
series = {SIGIR '20}
}

@inproceedings{ColBERTv2,
    title = "{C}ol{BERT}v2: Effective and Efficient Retrieval via Lightweight Late Interaction",
    author = "Santhanam, Keshav  and
      Khattab, Omar  and
      Saad-Falcon, Jon  and
      Potts, Christopher  and
      Zaharia, Matei",
    editor = "Carpuat, Marine  and
      de Marneffe, Marie-Catherine  and
      Meza Ruiz, Ivan Vladimir",
    booktitle = "Proceedings of the 2022 Conference of the North American Chapter of the Association for Computational Linguistics: Human Language Technologies",
    month = jul,
    year = "2022",
    address = "Seattle, United States",
    publisher = "Association for Computational Linguistics",
    url = "https://aclanthology.org/2022.naacl-main.272/",
    doi = "10.18653/v1/2022.naacl-main.272",
    pages = "3715--3734"
}

@inproceedings{
CoopRAG,
title={Cooperative Retrieval-Augmented Generation for Question Answering: Mutual Information Exchange and Ranking by Contrasting Layers},
author={Youmin Ko and Sung Jong Seo and Hyunjoon Kim},
booktitle={The Thirty-ninth Annual Conference on Neural Information Processing Systems},
year={2026},
url={https://openreview.net/forum?id=D2XdJf1tXW}
}

@misc{GTE,
      title={Towards General Text Embeddings with Multi-stage Contrastive Learning}, 
      author={Zehan Li and Xin Zhang and Yanzhao Zhang and Dingkun Long and Pengjun Xie and Meishan Zhang},
      year={2023},
      eprint={2308.03281},
      archivePrefix={arXiv},
      primaryClass={cs.CL},
      url={https://arxiv.org/abs/2308.03281}, 
}

@inproceedings{
NV-Embed,
title={{NV}-Embed: Improved Techniques for Training {LLM}s as Generalist Embedding Models},
author={Chankyu Lee and Rajarshi Roy and Mengyao Xu and Jonathan Raiman and Mohammad Shoeybi and Bryan Catanzaro and Wei Ping},
booktitle={The Thirteenth International Conference on Learning Representations},
year={2025},
url={https://openreview.net/forum?id=lgsyLSsDRe}
}

@inproceedings{
LLM2Vec,
title={{LLM}2Vec: Large Language Models Are Secretly Powerful Text Encoders},
author={Parishad BehnamGhader and Vaibhav Adlakha and Marius Mosbach and Dzmitry Bahdanau and Nicolas Chapados and Siva Reddy},
booktitle={First Conference on Language Modeling},
year={2024},
url={https://openreview.net/forum?id=IW1PR7vEBf}
}

@inproceedings{
GritLM,
title={Generative Representational Instruction Tuning},
author={Niklas Muennighoff and Hongjin SU and Liang Wang and Nan Yang and Furu Wei and Tao Yu and Amanpreet Singh and Douwe Kiela},
booktitle={The Thirteenth International Conference on Learning Representations},
year={2025},
url={https://openreview.net/forum?id=BC4lIvfSzv}
}

@inproceedings{OneGen,
    title = "{O}ne{G}en: Efficient One-Pass Unified Generation and Retrieval for {LLM}s",
    author = "Zhang, Jintian  and
      Peng, Cheng  and
      Sun, Mengshu  and
      Chen, Xiang  and
      Liang, Lei  and
      Zhang, Zhiqiang  and
      Zhou, Jun  and
      Chen, Huajun  and
      Zhang, Ningyu",
    editor = "Al-Onaizan, Yaser  and
      Bansal, Mohit  and
      Chen, Yun-Nung",
    booktitle = "Findings of the Association for Computational Linguistics: EMNLP 2024",
    month = nov,
    year = "2024",
    address = "Miami, Florida, USA",
    publisher = "Association for Computational Linguistics",
    url = "https://aclanthology.org/2024.findings-emnlp.237/",
    doi = "10.18653/v1/2024.findings-emnlp.237",
    pages = "4088--4119"
}

@inproceedings{MAGNET,
    title = "{MAGNET}: Augmenting Generative Decoders with Representation Learning and Infilling Capabilities",
    author = "Khosla, Savya  and
      Tiwari, Aditi  and
      Kafle, Kushal  and
      Jenni, Simon  and
      Zhao, Handong  and
      Collomosse, John  and
      Shi, Jing",
    editor = "Che, Wanxiang  and
      Nabende, Joyce  and
      Shutova, Ekaterina  and
      Pilehvar, Mohammad Taher",
    booktitle = "Proceedings of the 63rd Annual Meeting of the Association for Computational Linguistics (Volume 1: Long Papers)",
    month = jul,
    year = "2025",
    address = "Vienna, Austria",
    publisher = "Association for Computational Linguistics",
    url = "https://aclanthology.org/2025.acl-long.1325/",
    doi = "10.18653/v1/2025.acl-long.1325",
    pages = "27328--27346",
    ISBN = "979-8-89176-251-0"
}

@inproceedings{
GRACE,
title={{GRACE}: Generative Representation Learning via Contrastive Policy Optimization},
author={Jiashuo Sun and Shixuan Liu and Zhaochen Su and Xianrui Zhong and Pengcheng Jiang and Bowen Jin and Peiran Li and Weijia Shi and Jiawei Han},
booktitle={The Fourteenth International Conference on Learning Representations},
year={2026},
url={https://openreview.net/forum?id=hs9lwjH1bJ}
}

@inproceedings{CoT,
author = {Wei, Jason and Wang, Xuezhi and Schuurmans, Dale and Bosma, Maarten and Ichter, Brian and Xia, Fei and Chi, Ed H. and Le, Quoc V. and Zhou, Denny},
title = {Chain-of-Thought Prompting Elicits Reasoning in Large Language Models},
year = {2022},
isbn = {9781713871088},
publisher = {Curran Associates Inc.},
address = {Red Hook, NY, USA},
booktitle = {Proceedings of the 36th International Conference on Neural Information Processing Systems},
articleno = {1800},
numpages = {14},
location = {New Orleans, LA, USA},
series = {NIPS '22}
}

@inproceedings{CoQ,
    title = "Making Long-Context Language Models Better Multi-Hop Reasoners",
    author = "Li, Yanyang  and
      Liang, Shuo  and
      Lyu, Michael  and
      Wang, Liwei",
    editor = "Ku, Lun-Wei  and
      Martins, Andre  and
      Srikumar, Vivek",
    booktitle = "Proceedings of the 62nd Annual Meeting of the Association for Computational Linguistics (Volume 1: Long Papers)",
    month = aug,
    year = "2024",
    address = "Bangkok, Thailand",
    publisher = "Association for Computational Linguistics",
    url = "https://aclanthology.org/2024.acl-long.135/",
    doi = "10.18653/v1/2024.acl-long.135",
    pages = "2462--2475"
}

@inproceedings{CoVe,
    title = "Chain-of-Verification Reduces Hallucination in Large Language Models",
    author = "Dhuliawala, Shehzaad  and
      Komeili, Mojtaba  and
      Xu, Jing  and
      Raileanu, Roberta  and
      Li, Xian  and
      Celikyilmaz, Asli  and
      Weston, Jason",
    editor = "Ku, Lun-Wei  and
      Martins, Andre  and
      Srikumar, Vivek",
    booktitle = "Findings of the Association for Computational Linguistics: ACL 2024",
    month = aug,
    year = "2024",
    address = "Bangkok, Thailand",
    publisher = "Association for Computational Linguistics",
    url = "https://aclanthology.org/2024.findings-acl.212/",
    doi = "10.18653/v1/2024.findings-acl.212",
    pages = "3563--3578"
}

@inproceedings{CoN,
    title = "Chain-of-Note: Enhancing Robustness in Retrieval-Augmented Language Models",
    author = "Yu, Wenhao  and
      Zhang, Hongming  and
      Pan, Xiaoman  and
      Cao, Peixin  and
      Ma, Kaixin  and
      Li, Jian  and
      Wang, Hongwei  and
      Yu, Dong",
    editor = "Al-Onaizan, Yaser  and
      Bansal, Mohit  and
      Chen, Yun-Nung",
    booktitle = "Proceedings of the 2024 Conference on Empirical Methods in Natural Language Processing",
    month = nov,
    year = "2024",
    address = "Miami, Florida, USA",
    publisher = "Association for Computational Linguistics",
    url = "https://aclanthology.org/2024.emnlp-main.813/",
    doi = "10.18653/v1/2024.emnlp-main.813",
    pages = "14672--14685"
}

@misc{
Coconut,
title={Training Large Language Model to Reason in a Continuous Latent Space},
author={Shibo Hao and Sainbayar Sukhbaatar and DiJia Su and Xian Li and Zhiting Hu and Jason E Weston and Yuandong Tian},
year={2025},
url={https://openreview.net/forum?id=tG4SgayTtk}
}

@inproceedings{CODI,
    title = "{CODI}: Compressing Chain-of-Thought into Continuous Space via Self-Distillation",
    author = "Shen, Zhenyi  and
      Yan, Hanqi  and
      Zhang, Linhai  and
      Hu, Zhanghao  and
      Du, Yali  and
      He, Yulan",
    editor = "Christodoulopoulos, Christos  and
      Chakraborty, Tanmoy  and
      Rose, Carolyn  and
      Peng, Violet",
    booktitle = "Proceedings of the 2025 Conference on Empirical Methods in Natural Language Processing",
    month = nov,
    year = "2025",
    address = "Suzhou, China",
    publisher = "Association for Computational Linguistics",
    url = "https://aclanthology.org/2025.emnlp-main.36/",
    doi = "10.18653/v1/2025.emnlp-main.36",
    pages = "677--693",
    ISBN = "979-8-89176-332-6"
}

@inproceedings{
Sim-CoT,
title={{SIM}-CoT: Supervised Implicit Chain-of-Thought},
author={Xilin Wei and Xiaoran Liu and Yuhang Zang and Xiaoyi Dong and Yuhang Cao and Jiaqi Wang and Xipeng Qiu and Dahua Lin},
booktitle={The Fourteenth International Conference on Learning Representations},
year={2026},
url={https://openreview.net/forum?id=6YRJ4jmVQl}
}

@inproceedings{
eRank,
title={Matrix Information Theory for Self-Supervised Learning},
author={Yifan Zhang and Zhiquan Tan and Jingqin Yang and Weiran Huang and Yang Yuan},
booktitle={Forty-first International Conference on Machine Learning},
year={2024},
url={https://openreview.net/forum?id=wleAlsklEh}
}

@inproceedings{rc1,
  title={Attention is not all you need: Pure attention loses rank doubly exponentially with depth},
  author={Dong, Yihe and Cordonnier, Jean-Baptiste and Loukas, Andreas},
  booktitle={International conference on machine learning},
  pages={2793--2803},
  year={2021},
  organization={PMLR}
}

@inproceedings{rc2,
author = {Noci, Lorenzo and Anagnostidis, Sotiris and Biggio, Luca and Orvieto, Antonio and Singh, Sidak Pal and Lucchi, Aurelien},
title = {Signal propagation in transformers: theoretical perspectives and the role of rank collapse},
year = {2022},
isbn = {9781713871088},
publisher = {Curran Associates Inc.},
address = {Red Hook, NY, USA},
booktitle = {Proceedings of the 36th International Conference on Neural Information Processing Systems},
articleno = {1972},
numpages = {14},
location = {New Orleans, LA, USA},
series = {NIPS '22}
}

@inproceedings{rc3,
author = {Wu, Xinyi and Ajorlou, Amir and Wang, Yifei and Jegelka, Stefanie and Jadbabaie, Ali},
title = {On the role of attention masks and LayerNorm in transformers},
year = {2024},
isbn = {9798331314385},
publisher = {Curran Associates Inc.},
address = {Red Hook, NY, USA},
booktitle = {Proceedings of the 38th International Conference on Neural Information Processing Systems},
articleno = {472},
numpages = {36},
location = {Vancouver, BC, Canada},
series = {NIPS '24}
}

@inproceedings{
rc4,
title={Transformers need glasses! Information over-squashing in language tasks},
author={Federico Barbero and Andrea Banino and Steven Kapturowski and Dharshan Kumaran and Jo{\~a}o Guilherme Madeira Ara{\'u}jo and Alex Vitvitskyi and Razvan Pascanu and Petar Veli{\v{c}}kovi{\'c}},
booktitle={The Thirty-eighth Annual Conference on Neural Information Processing Systems},
year={2024},
url={https://openreview.net/forum?id=93HCE8vTye}
}

@article{
rc5,
title={When Attention Collapses: How Degenerate Layers in {LLM}s Enable Smaller, Stronger Models},
author={Sunny Sanyal and Ravid Shwartz-Ziv and Alex Dimakis and sujay sanghavi},
journal={Transactions on Machine Learning Research},
issn={2835-8856},
year={2026},
url={https://openreview.net/forum?id=2zQn0bUoPf},
note={}
}

@misc{rc6,
      title={Is Random Attention Sufficient for Sequence Modeling? Disentangling Trainable Components in the Transformer}, 
      author={Yihe Dong and Lorenzo Noci and Mikhail Khodak and Mufan Li},
      year={2025},
      eprint={2506.01115},
      archivePrefix={arXiv},
      primaryClass={cs.LG},
      url={https://arxiv.org/abs/2506.01115}, 
}

@inproceedings{
rc7,
title={From Condensation to Rank Collapse: A Two-Stage Analysis of Transformer Training Dynamics},
author={Zheng-An Chen and Tao Luo},
booktitle={The Thirty-ninth Annual Conference on Neural Information Processing Systems},
year={2026},
url={https://openreview.net/forum?id=gm5mkiTGOy}
}

@inproceedings{
rc8,
title={{LDR}eg: Local Dimensionality Regularized Self-Supervised Learning},
author={Hanxun Huang and Ricardo J. G. B. Campello and Sarah Monazam Erfani and Xingjun Ma and Michael E. Houle and James Bailey},
booktitle={The Twelfth International Conference on Learning Representations},
year={2024},
url={https://openreview.net/forum?id=oZyAqjAjJW}
}

@inproceedings{
rc9,
title={Preventing Dimensional Collapse in Self-Supervised Learning via Orthogonality Regularization},
author={Junlin He and Jinxiao Du and Wei Ma},
booktitle={The Thirty-eighth Annual Conference on Neural Information Processing Systems},
year={2024},
url={https://openreview.net/forum?id=Y3FjKSsfmy}
}

@inproceedings{HotpotQA,
    title = "{H}otpot{QA}: A Dataset for Diverse, Explainable Multi-hop Question Answering",
    author = "Yang, Zhilin  and
      Qi, Peng  and
      Zhang, Saizheng  and
      Bengio, Yoshua  and
      Cohen, William  and
      Salakhutdinov, Ruslan  and
      Manning, Christopher D.",
    editor = "Riloff, Ellen  and
      Chiang, David  and
      Hockenmaier, Julia  and
      Tsujii, Jun{'}ichi",
    booktitle = "Proceedings of the 2018 Conference on Empirical Methods in Natural Language Processing",
    month = oct # "-" # nov,
    year = "2018",
    address = "Brussels, Belgium",
    publisher = "Association for Computational Linguistics",
    url = "https://aclanthology.org/D18-1259/",
    doi = "10.18653/v1/D18-1259",
    pages = "2369--2380"
}

@inproceedings{2WikiMultihopQA,
    title = "Constructing A Multi-hop {QA} Dataset for Comprehensive Evaluation of Reasoning Steps",
    author = "Ho, Xanh  and
      Duong Nguyen, Anh-Khoa  and
      Sugawara, Saku  and
      Aizawa, Akiko",
    editor = "Scott, Donia  and
      Bel, Nuria  and
      Zong, Chengqing",
    booktitle = "Proceedings of the 28th International Conference on Computational Linguistics",
    month = dec,
    year = "2020",
    address = "Barcelona, Spain (Online)",
    publisher = "International Committee on Computational Linguistics",
    url = "https://aclanthology.org/2020.coling-main.580/",
    doi = "10.18653/v1/2020.coling-main.580",
    pages = "6609--6625"
}

@article{MuSiQue,
    title = "MuSiQue: Multihop Questions via Single-hop Question Composition",
    author = "Trivedi, Harsh  and
      Balasubramanian, Niranjan  and
      Khot, Tushar  and
      Sabharwal, Ashish",
    editor = "Roark, Brian  and
      Nenkova, Ani",
    journal = "Transactions of the Association for Computational Linguistics",
    volume = "10",
    year = "2022",
    address = "Cambridge, MA",
    publisher = "MIT Press",
    url = "https://aclanthology.org/2022.tacl-1.31/",
    doi = "10.1162/tacl_a_00475",
    pages = "539--554"
}

@misc{SimpleQA,
      title={Measuring short-form factuality in large language models}, 
      author={Jason Wei and Nguyen Karina and Hyung Won Chung and Yunxin Joy Jiao and Spencer Papay and Amelia Glaese and John Schulman and William Fedus},
      year={2024},
      eprint={2411.04368},
      archivePrefix={arXiv},
      primaryClass={cs.CL},
      url={https://arxiv.org/abs/2411.04368}, 
}

@article{NaturalQuestions,
    title = "Natural Questions: A Benchmark for Question Answering Research",
    author = "Kwiatkowski, Tom  and
      Palomaki, Jennimaria  and
      Redfield, Olivia  and
      Collins, Michael  and
      Parikh, Ankur  and
      Alberti, Chris  and
      Epstein, Danielle  and
      Polosukhin, Illia  and
      Devlin, Jacob  and
      Lee, Kenton  and
      Toutanova, Kristina  and
      Jones, Llion  and
      Kelcey, Matthew  and
      Chang, Ming-Wei  and
      Dai, Andrew M.  and
      Uszkoreit, Jakob  and
      Le, Quoc  and
      Petrov, Slav",
    editor = "Lee, Lillian  and
      Johnson, Mark  and
      Roark, Brian  and
      Nenkova, Ani",
    journal = "Transactions of the Association for Computational Linguistics",
    volume = "7",
    year = "2019",
    address = "Cambridge, MA",
    publisher = "MIT Press",
    url = "https://aclanthology.org/Q19-1026/",
    doi = "10.1162/tacl_a_00276",
    pages = "452--466"
}

@inproceedings{
LoRA,
title={Lo{RA}: Low-Rank Adaptation of Large Language Models},
author={Edward J Hu and yelong shen and Phillip Wallis and Zeyuan Allen-Zhu and Yuanzhi Li and Shean Wang and Lu Wang and Weizhu Chen},
booktitle={International Conference on Learning Representations},
year={2022},
url={https://openreview.net/forum?id=nZeVKeeFYf9}
}

@inproceedings{E5,
    title = "Improving Text Embeddings with Large Language Models",
    author = "Wang, Liang  and
      Yang, Nan  and
      Huang, Xiaolong  and
      Yang, Linjun  and
      Majumder, Rangan  and
      Wei, Furu",
    editor = "Ku, Lun-Wei  and
      Martins, Andre  and
      Srikumar, Vivek",
    booktitle = "Proceedings of the 62nd Annual Meeting of the Association for Computational Linguistics (Volume 1: Long Papers)",
    month = aug,
    year = "2024",
    address = "Bangkok, Thailand",
    publisher = "Association for Computational Linguistics",
    url = "https://aclanthology.org/2024.acl-long.642/",
    doi = "10.18653/v1/2024.acl-long.642",
    pages = "11897--11916"
}

@inproceedings{MTEB,
    title = "{MTEB}: Massive Text Embedding Benchmark",
    author = "Muennighoff, Niklas  and
      Tazi, Nouamane  and
      Magne, Loic  and
      Reimers, Nils",
    editor = "Vlachos, Andreas  and
      Augenstein, Isabelle",
    booktitle = "Proceedings of the 17th Conference of the European Chapter of the Association for Computational Linguistics",
    month = may,
    year = "2023",
    address = "Dubrovnik, Croatia",
    publisher = "Association for Computational Linguistics",
    url = "https://aclanthology.org/2023.eacl-main.148/",
    doi = "10.18653/v1/2023.eacl-main.148",
    pages = "2014--2037"
}

@inproceedings{
MMLU,
title={Measuring Massive Multitask Language Understanding},
author={Dan Hendrycks and Collin Burns and Steven Basart and Andy Zou and Mantas Mazeika and Dawn Song and Jacob Steinhardt},
booktitle={International Conference on Learning Representations},
year={2021},
url={https://openreview.net/forum?id=d7KBjmI3GmQ}
}

@article{GSM8K,
  title={Training verifiers to solve math word problems},
  author={Cobbe, Karl and Kosaraju, Vineet and Bavarian, Mohammad and Chen, Mark and Jun, Heewoo and Kaiser, Lukasz and Plappert, Matthias and Tworek, Jerry and Hilton, Jacob and Nakano, Reiichiro and others},
  journal={arXiv preprint arXiv:2110.14168},
  year={2021}
}

@article{TyDiQA,
    title = "{T}y{D}i {QA}: A Benchmark for Information-Seeking Question Answering in Typologically Diverse Languages",
    author = "Clark, Jonathan H.  and
      Choi, Eunsol  and
      Collins, Michael  and
      Garrette, Dan  and
      Kwiatkowski, Tom  and
      Nikolaev, Vitaly  and
      Palomaki, Jennimaria",
    editor = "Johnson, Mark  and
      Roark, Brian  and
      Nenkova, Ani",
    journal = "Transactions of the Association for Computational Linguistics",
    volume = "8",
    year = "2020",
    address = "Cambridge, MA",
    publisher = "MIT Press",
    url = "https://aclanthology.org/2020.tacl-1.30/",
    doi = "10.1162/tacl_a_00317",
    pages = "454--470"
}

@misc{Gemma2,
  title={{Gemma 2: Improving Open Language Models at a Practical Size}},
  author={{Gemma Team} and Morgane Riviere and Shreya Pathak and Pier Giuseppe Sessa and others},
  year={2024},
  eprint={2408.00118},
  archivePrefix={arXiv},
  primaryClass={cs.CL},
  url={https://arxiv.org/abs/2408.00118}
}

@misc{Qwen2,
      title={Qwen2.5 Technical Report}, 
      author={Qwen and : and An Yang and Baosong Yang and Beichen Zhang and Binyuan Hui and Bo Zheng and Bowen Yu and Chengyuan Li and Dayiheng Liu and Fei Huang and Haoran Wei and Huan Lin and Jian Yang and Jianhong Tu and Jianwei Zhang and Jianxin Yang and Jiaxi Yang and Jingren Zhou and Junyang Lin and Kai Dang and Keming Lu and Keqin Bao and Kexin Yang and Le Yu and Mei Li and Mingfeng Xue and Pei Zhang and Qin Zhu and Rui Men and Runji Lin and Tianhao Li and Tianyi Tang and Tingyu Xia and Xingzhang Ren and Xuancheng Ren and Yang Fan and Yang Su and Yichang Zhang and Yu Wan and Yuqiong Liu and Zeyu Cui and Zhenru Zhang and Zihan Qiu},
      year={2025},
      eprint={2412.15115},
      archivePrefix={arXiv},
      primaryClass={cs.CL},
      url={https://arxiv.org/abs/2412.15115}, 
}

@misc{LlaMa3,
  title={{The Llama 3 Herd of Models}},
  author={Aaron Grattafiori and Abhimanyu Dubey and Abhinav Jauhri and others},
  year={2024},
  eprint={2407.21783},
  archivePrefix={arXiv},
  primaryClass={cs.AI},
  url={https://arxiv.org/abs/2407.21783}
}

@misc{Mistral,
      title={Mistral 7B}, 
      author={Albert Q. Jiang and Alexandre Sablayrolles and Arthur Mensch and Chris Bamford and Devendra Singh Chaplot and Diego de las Casas and Florian Bressand and Gianna Lengyel and Guillaume Lample and Lucile Saulnier and Lélio Renard Lavaud and Marie-Anne Lachaux and Pierre Stock and Teven Le Scao and Thibaut Lavril and Thomas Wang and Timothée Lacroix and William El Sayed},
      year={2023},
      eprint={2310.06825},
      archivePrefix={arXiv},
      primaryClass={cs.CL},
      url={https://arxiv.org/abs/2310.06825}, 
}

@inproceedings{Proposition,
    title = "Dense {X} Retrieval: What Retrieval Granularity Should We Use?",
    author = "Chen, Tong  and
      Wang, Hongwei  and
      Chen, Sihao  and
      Yu, Wenhao  and
      Ma, Kaixin  and
      Zhao, Xinran  and
      Zhang, Hongming  and
      Yu, Dong",
    editor = "Al-Onaizan, Yaser  and
      Bansal, Mohit  and
      Chen, Yun-Nung",
    booktitle = "Proceedings of the 2024 Conference on Empirical Methods in Natural Language Processing",
    month = nov,
    year = "2024",
    address = "Miami, Florida, USA",
    publisher = "Association for Computational Linguistics",
    url = "https://aclanthology.org/2024.emnlp-main.845/",
    doi = "10.18653/v1/2024.emnlp-main.845",
}

@InProceedings{AlternatingLearning1,
  title = 	 {Modality Competition: What Makes Joint Training of Multi-modal Network Fail in Deep Learning? ({P}rovably)},
  author =       {Huang, Yu and Lin, Junyang and Zhou, Chang and Yang, Hongxia and Huang, Longbo},
  booktitle = 	 {Proceedings of the 39th International Conference on Machine Learning},
  pages = 	 {9226--9259},
  year = 	 {2022},
  editor = 	 {Chaudhuri, Kamalika and Jegelka, Stefanie and Song, Le and Szepesvari, Csaba and Niu, Gang and Sabato, Sivan},
  volume = 	 {162},
  series = 	 {Proceedings of Machine Learning Research},
  month = 	 {17--23 Jul},
  publisher =    {PMLR},
  url = 	 {https://proceedings.mlr.press/v162/huang22e.html}
}

@misc{
AlternatingLearning2,
title={Multimodal Representation Learning by Alternating Unimodal Adaptation},
author={XiaoHui Zhang and Jaehong Yoon and Mohit Bansal and Huaxiu Yao},
year={2024},
url={https://openreview.net/forum?id=JFEFORc7fP}
}

@inproceedings{
AlternatingLearning3,
title={A Closer Look at Multimodal Representation Collapse},
author={Abhra Chaudhuri and Anjan Dutta and Tu Bui and Serban Georgescu},
booktitle={Forty-second International Conference on Machine Learning},
year={2025},
url={https://openreview.net/forum?id=Vf9f7eNX6T}
}

@inproceedings{beta-smoothness1,
author = {Hardt, Moritz and Recht, Benjamin and Singer, Yoram},
title = {Train faster, generalize better: stability of stochastic gradient descent},
year = {2016},
publisher = {JMLR.org},
booktitle = {Proceedings of the 33rd International Conference on International Conference on Machine Learning - Volume 48},
pages = {1225–1234},
numpages = {10},
location = {New York, NY, USA},
series = {ICML'16}
}

@article{beta-smoothness2,
  title={Stochastic first-and zeroth-order methods for nonconvex stochastic programming},
  author={Ghadimi, Saeed and Lan, Guanghui},
  journal={SIAM journal on optimization},
  volume={23},
  number={4},
  pages={2341--2368},
  year={2013},
  publisher={SIAM}
}

@inproceedings{Lipschitz1 ,
author = {Scaman, Kevin and Virmaux, Aladin},
title = {Lipschitz regularity of deep neural networks: analysis and efficient estimation},
year = {2018},
publisher = {Curran Associates Inc.},
address = {Red Hook, NY, USA},
booktitle = {Proceedings of the 32nd International Conference on Neural Information Processing Systems},
pages = {3839–3848},
numpages = {10},
location = {Montr\'{e}al, Canada},
series = {NIPS'18}
}

@inproceedings{Lipschitz2,
author = {Fazlyab, Mahyar and Robey, Alexander and Hassani, Hamed and Morari, Manfred and Pappas, George J.},
title = {Efficient and accurate estimation of lipschitz constants for deep neural networks},
year = {2019},
publisher = {Curran Associates Inc.},
address = {Red Hook, NY, USA},
booktitle = {Proceedings of the 33rd International Conference on Neural Information Processing Systems},
articleno = {1025},
numpages = {12}
}

@inproceedings{
DSAS,
title={{DSAS}: A Universal Plug-and-Play Framework for Attention Optimization in Multi-Document Question Answering},
author={Jiakai Li and Rongzheng Wang and Yizhuo Ma and Shuang Liang and Guangchun Luo and Ke Qin},
booktitle={The Thirty-ninth Annual Conference on Neural Information Processing Systems},
year={2026},
url={https://openreview.net/forum?id=y68Q09Vc4K}
}

@inproceedings{
DoLa,
title={DoLa: Decoding by Contrasting Layers Improves Factuality in Large Language Models},
author={Yung-Sung Chuang and Yujia Xie and Hongyin Luo and Yoon Kim and James R. Glass and Pengcheng He},
booktitle={The Twelfth International Conference on Learning Representations},
year={2024},
url={https://openreview.net/forum?id=Th6NyL07na}
}

@article{hallu1,
author = {Ji, Ziwei and Lee, Nayeon and Frieske, Rita and Yu, Tiezheng and Su, Dan and Xu, Yan and Ishii, Etsuko and Bang, Ye Jin and Madotto, Andrea and Fung, Pascale},
title = {Survey of Hallucination in Natural Language Generation},
year = {2023},
issue_date = {December 2023},
publisher = {Association for Computing Machinery},
address = {New York, NY, USA},
volume = {55},
number = {12},
issn = {0360-0300},
url = {https://doi.org/10.1145/3571730},
doi = {10.1145/3571730},
journal = {ACM Comput. Surv.},
month = mar,
articleno = {248},
numpages = {38}
}

@inproceedings{hallu2,
    title = "Controlling Risk of Retrieval-augmented Generation: A Counterfactual Prompting Framework",
    author = "Chen, Lu  and
      Zhang, Ruqing  and
      Guo, Jiafeng  and
      Fan, Yixing  and
      Cheng, Xueqi",
    editor = "Al-Onaizan, Yaser  and
      Bansal, Mohit  and
      Chen, Yun-Nung",
    booktitle = "Findings of the Association for Computational Linguistics: EMNLP 2024",
    month = nov,
    year = "2024",
    address = "Miami, Florida, USA",
    publisher = "Association for Computational Linguistics",
    url = "https://aclanthology.org/2024.findings-emnlp.133/",
    doi = "10.18653/v1/2024.findings-emnlp.133",
    pages = "2380--2393"
}

@inproceedings{hallu3,
 author = {Jesson, Andrew and Beltran-Velez, Nicolas and Chu, Quentin and Karlekar, Sweta and Kossen, Jannik and Gal, Yarin and Cunningham, John P. and Blei, David},
 booktitle = {Advances in Neural Information Processing Systems},
 editor = {A. Globerson and L. Mackey and D. Belgrave and A. Fan and U. Paquet and J. Tomczak and C. Zhang},
 pages = {31154--31201},
 publisher = {Curran Associates, Inc.},
 title = {Estimating the Hallucination Rate of Generative AI},
 url = {https://proceedings.neurips.cc/paper_files/paper/2024/file/3791f5fc0e8e43730466afd2bcdb7493-Paper-Conference.pdf},
 volume = {37},
 year = {2024}
}

@inproceedings{hallu4,
    title = "A Comprehensive Survey of Hallucination in Large Language, Image, Video and Audio Foundation Models",
    author = "Sahoo, Pranab  and
      Meharia, Prabhash  and
      Ghosh, Akash  and
      Saha, Sriparna  and
      Jain, Vinija  and
      Chadha, Aman",
    editor = "Al-Onaizan, Yaser  and
      Bansal, Mohit  and
      Chen, Yun-Nung",
    booktitle = "Findings of the Association for Computational Linguistics: EMNLP 2024",
    month = nov,
    year = "2024",
    address = "Miami, Florida, USA",
    publisher = "Association for Computational Linguistics",
    url = "https://aclanthology.org/2024.findings-emnlp.685/",
    doi = "10.18653/v1/2024.findings-emnlp.685",
    pages = "11709--11724"
}

@article{hallu5,
   title={A Survey on Hallucination in Large Language Models: Principles, Taxonomy, Challenges, and Open Questions},
   volume={43},
   ISSN={1558-2868},
   url={http://dx.doi.org/10.1145/3703155},
   DOI={10.1145/3703155},
   number={2},
   journal={ACM Transactions on Information Systems},
   publisher={Association for Computing Machinery (ACM)},
   author={Huang, Lei and Yu, Weijiang and Ma, Weitao and Zhong, Weihong and Feng, Zhangyin and Wang, Haotian and Chen, Qianglong and Peng, Weihua and Feng, Xiaocheng and Qin, Bing and Liu, Ting},
   year={2025},
   month=jan, pages={1–55} }

@Article{mhqa1,
AUTHOR = {Han, Kyubeen and Jang, Youngjin and Kim, Harksoo},
TITLE = {Explainable Multi-Hop Question Answering: A Rationale-Based Approach},
JOURNAL = {Big Data and Cognitive Computing},
VOLUME = {9},
YEAR = {2025},
NUMBER = {11},
ARTICLE-NUMBER = {273},
URL = {https://www.mdpi.com/2504-2289/9/11/273},
ISSN = {2504-2289},
DOI = {10.3390/bdcc9110273}
}

@inproceedings{mhqa2,
    title = "A Survey for Efficient Open Domain Question Answering",
    author = "Zhang, Qin  and
      Chen, Shangsi  and
      Xu, Dongkuan  and
      Cao, Qingqing  and
      Chen, Xiaojun  and
      Cohn, Trevor  and
      Fang, Meng",
    editor = "Rogers, Anna  and
      Boyd-Graber, Jordan  and
      Okazaki, Naoaki",
    booktitle = "Proceedings of the 61st Annual Meeting of the Association for Computational Linguistics (Volume 1: Long Papers)",
    month = jul,
    year = "2023",
    address = "Toronto, Canada",
    publisher = "Association for Computational Linguistics",
    url = "https://aclanthology.org/2023.acl-long.808/",
    doi = "10.18653/v1/2023.acl-long.808",
    pages = "14447--14465"
}

@article{mhqa3,
  title={Multi-hop question answering},
  author={Mavi, Vaibhav and Jangra, Anubhav and Jatowt, Adam},
  journal={Foundations and Trends{\textregistered} in Information Retrieval},
  volume={17},
  number={4},
  pages={457--586},
  year={2024},
  publisher={Emerald Publishing Limited}
}

@article{kgqa1, title={ProgRAG: Hallucination-Resistant Progressive Retrieval and Reasoning over Knowledge Graphs}, volume={40}, url={https://ojs.aaai.org/index.php/AAAI/article/view/40545}, DOI={10.1609/aaai.v40i39.40545}, abstractNote={Large Language Models (LLMs) demonstrate strong reasoning capabilities but still struggle with hallucinations and limited transparency. Recently, KG-enhanced LLMs that integrate knowledge graphs (KGs) have been shown to improve reasoning performance, particularly for complex, knowledge-intensive tasks. However, these methods still face significant challenges, including inaccurate retrieval and reasoning failures, often exacerbated by long input contexts that obscure relevant information. Furthermore, many of these approaches rely on LLMs to directly retrieve evidence from KGs, and to self-assess the sufficiency of this evidence, which often results in premature or incorrect reasoning. To address the retrieval and reasoning failures, we propose ProgRAG, a multi-hop knowledge graph question answering (KGQA) framework that decomposes complex questions into sub-questions, and progressively extends partial reasoning paths by answering each sub-question. At each step, external retrievers gather candidate evidence, which is then refined through uncertainty-aware pruning by the LLM. Finally, the context for LLM reasoning is optimized by organizing and rearranging the partial reasoning paths obtained from the sub-question answers. Experiments on two well-known datasets, WebQSP and CWQ, demonstrate that ProgRAG outperforms existing baselines in multi-hop KGQA, offering improved reliability and reasoning quality.}, number={39}, journal={Proceedings of the AAAI Conference on Artificial Intelligence}, author={Park, Minbae and Yang, Hyemin and Kim, Jeonghyun and Park, Kunsoo and Kim, Hyunjoon}, year={2026}, month={Mar.}, pages={32674-32682} }

@inproceedings{
kgqa2,
title={Think-on-Graph: Deep and Responsible Reasoning of Large Language Model on Knowledge Graph},
author={Jiashuo Sun and Chengjin Xu and Lumingyuan Tang and Saizhuo Wang and Chen Lin and Yeyun Gong and Lionel Ni and Heung-Yeung Shum and Jian Guo},
booktitle={The Twelfth International Conference on Learning Representations},
year={2024},
url={https://openreview.net/forum?id=nnVO1PvbTv}
}

@inproceedings{
kgqa3,
title={Think-on-Graph 2.0: Deep and Faithful Large Language Model Reasoning with Knowledge-guided Retrieval Augmented Generation},
author={Shengjie Ma and Chengjin Xu and Xuhui Jiang and Muzhi Li and Huaren Qu and Cehao Yang and Jiaxin Mao and Jian Guo},
booktitle={The Thirteenth International Conference on Learning Representations},
year={2025},
url={https://openreview.net/forum?id=oFBu7qaZpS}
}

@inproceedings{
kgqa4,
title={Simple is Effective: The Roles of Graphs and Large Language Models in Knowledge-Graph-Based Retrieval-Augmented Generation},
author={Mufei Li and Siqi Miao and Pan Li},
booktitle={The Thirteenth International Conference on Learning Representations},
year={2025},
url={https://openreview.net/forum?id=JvkuZZ04O7}
}

@inproceedings{
evi-sense,
title={Retrieval is Not Enough: Enhancing {RAG} through Test-Time Critique and Optimization},
author={Jiaqi Wei and Hao Zhou and Xiang Zhang and Di Zhang and Zijie Qiu and Noah Wei and Jinzhe Li and Wanli Ouyang and Siqi Sun},
booktitle={The Thirty-ninth Annual Conference on Neural Information Processing Systems},
year={2026},
url={https://openreview.net/forum?id=cnUq7GkS6d}
}

@article{SPRING, title={One Token Can Help! Learning Scalable and Pluggable Virtual Tokens for Retrieval-Augmented Large Language Models}, volume={39}, url={https://ojs.aaai.org/index.php/AAAI/article/view/34813}, DOI={10.1609/aaai.v39i24.34813}, abstractNote={Retrieval-augmented generation (RAG) is a promising way to improve large language models (LLMs) for generating more factual, accurate, and up-to-date content. Existing methods either optimize prompts to guide LLMs in leveraging retrieved information or directly fine-tune LLMs to adapt to RAG scenarios. Although fine-tuning can yield better performance, it often compromises the LLMs’ general generation capabilities by modifying their parameters. This limitation poses challenges in practical applications, especially when LLMs are already deployed, as parameter adjustments may affect their original functionality. To address this, we propose a novel method that involves learning scalable and pluggable virtual tokens for RAG. By maintaining the LLMs’ original parameters and fine-tuning only the embeddings of these pluggable tokens, our approach not only enhances LLMs’ performance but also preserves their general generation capabilities. Furthermore, we design several training strategies to improve the scalability, flexibility, and generalizability of our method. Comprehensive experiments across 12 question-answering tasks demonstrate the superiority of our approach.}, number={24}, journal={Proceedings of the AAAI Conference on Artificial Intelligence}, author={Zhu, Yutao and Huang, Zhaoheng and Dou, Zhicheng and Wen, Ji-Rong}, year={2025}, month={Apr.}, pages={26166-26174} }

@article{InfoNCE,
  title={Representation learning with contrastive predictive coding},
  author={Oord, Aaron van den and Li, Yazhe and Vinyals, Oriol},
  journal={arXiv preprint arXiv:1807.03748},
  year={2018}
}
\bibliographystyle{plainnat}

\newpage
\appendix

\section*{Appendices}
In this supplementary material, we provide further details on the following topics:
\setlist[itemize]{%
  label=\textbullet,     
  leftmargin=1.5em,      
  itemsep=0.25em,         
  topsep=0.5em             
}

\begin{itemize}
  \item \textbf{Appendix \ref{sec:appendix_exsetup}: Experimental Setup}
  \item \textbf{Appendix \ref{sec:appendix_warmup}: Warm-up Training}
    \begin{itemize}
      \item \ref{sec:appendix_ret_warm_up}: Encoder mode warm-up training
      \item \ref{sec:appendix_intialization}: Anchor Token Initialization

    \end{itemize}
  \item \textbf{Appendix \ref{sec:appendix_results}: Overall Performance}
    \begin{itemize}
        \item \ref{sec:appendix_result_retrieval}: Retrieval Performance
        \item \ref{sec:appendix_result_generation}: Question Answering Performance
        \item \ref{sec:appendix_single-hopQA}: Retrieval and Question Answering Performance on Single-hop QA Benchmark
        \item \ref{sec:appendix_simple_qa}: Question Answering Performance on SimpleQA
        \item \ref{sec:appendix_general_tasks}: Embedding and Generative Generalization
        \item \ref{sec:appendix_LatentCoT}: Performance Comparison with Latent-CoT Baselines
        \item \ref{sec:appendix_ablation_SST}: Ablation Study: Semantic Signal Translator
        \item \ref{sec:appendix_review3}: Performance Comparison with Simple Projection Baselines
        \item \ref{sec:appendix_review4}: Impact of Causal-Mask Relaxation
    \end{itemize}
  \item \textbf{Appendix \ref{sec:appendix_latency}: Latency}
    \begin{itemize}
        \item \ref{sec:appendix_latency_training}: Training Latency
        \item \ref{sec:appendix_Complexity}: Complexity Comparison with RaLa 
        \item \ref{sec:appendix_scoring_tradeoff}: Effectiveness-Efficiency Trade-off of Different Relevance Scoring Methods
    \end{itemize}
  \item \textbf{Appendix \ref{sec:appendix_analysis}: Analysis}
    \begin{itemize}
        \item \ref{sec:appendix_analysis_Scalability}: Scalability Analysis
        \item \ref{sec:appendix_anchor_initalization}: Impact of Anchor Token Initialization
        \item \ref{sec:appendix_SST_confidence}: Impact of SST on Confidence Calibration and LoRA Adaptation
        \item \ref{sec:appendix_review2}: Impact of SST with Fixed External Retrievers
        \item \ref{appendix_analysis_learning_strategy}: Theoretical Interpretation of Learning Strategy
        \item \ref{appendix_analysis_overconfidence}: Overconfidence Analysis on Correct and Incorrect Cases
        \item \ref{appendix_analysis_placement}: Effect of Soft Knowledge Token Placement
        \item \ref{appendix_analysis_tSNE}: $t$-SNE Visualization Before and After Training with Bidirectional and Unidirectional Attention
        \item \ref{appendix_analysis_eRank_backbones}: Effective Rank Before and After Encoder-mode Training across LLM Backbones
        \item \ref{sec:appendix_review5}: SST Effectiveness on Modern Large LLMs
    \end{itemize}
  \item \textbf{Appendix \ref{sec:appendix_hyperparameter}: Hyperparameter Sensitivity}
    \begin{itemize}
        \item \ref{sec:appendix_hyperparameter_batchsize}: Impact of Batch Size
        \item \ref{sec:appendix_hyperparameter_LoRA_rank}: Impact of LoRA Rank
        \item \ref{sec:appendix_hyperparameter_lambda_SOA}: Impact of $\lambda_{\text{SOA}}$
        \item \ref{sec:appendix_hyperparameter_labda_KD}: Impact of $\lambda_{\text{KD}}$
        \item \ref{sec:appendix_hyperparameter_tau}: Impact of $\tau$ in $\mathcal{L}_{\text{LGC}}$
        \item \ref{sec:appendix_hyperparameter_bucket_size}: Impact of Candidate Layers in $\mathcal{L}_{\text{LGC}}$
        \item \ref{sec:appendix_hyperparameter_candidate_layers}: Impact of Number of Layers in $\lambda_{\text{SOA}}$
        \item \ref{sec:appendix_hyperparameter_llm_backbone}: Impact of LLM Backbones
    \end{itemize}
  \item \textbf{Appendix \ref{sec:appendix_case_study}: Case Study}
    \begin{itemize}
        \item \ref{sec:case1}: Knowledge Comparison across Different Attention Policies
    \end{itemize}
  \item \textbf{Appendix \ref{sec:appendix_prompt}: Prompt}
    \begin{itemize}
        \item \ref{sec:appendix_prompt1}: Prompt for Rationale Generation
        \item \ref{sec:appendix_prompt2}: Prompt for Anchor Token Initialization
        \item \ref{sec:appendix_prompt3}: Prompt for Decoder-Mode Training
    \end{itemize}

\end{itemize}
\clearpage

\section{Experimental Setup}\label{sec:appendix_exsetup}
\begingroup
  \renewcommand{\arraystretch}{0.80}
  \setlength{\tabcolsep}{3pt}

  \begin{table*}[h]
    \centering
    \footnotesize
    \caption{Training-set statistics for decoder mode.}
    \label{tab:appendix_generation_statistics}
    \begin{tabular*}{\textwidth}{@{\extracolsep{\fill}}lccccc}
      \toprule
      &
      \textbf{1-hop}
      & \textbf{2-hop}
      & \textbf{3-hop}
      & \textbf{4-hop}
      & \textbf{Total} \\
      \midrule
      Examples
      & 2,000 & 6,000 & 500 & 1,500 & 10,000 \\
      \bottomrule
    \end{tabular*}
  \end{table*}
\endgroup
\begingroup
  \renewcommand{\arraystretch}{0.80}
  \setlength{\tabcolsep}{3pt}

  \begin{table*}[h]
    \centering
    \footnotesize
    \caption{Dataset statistics (test sets)}
    \label{tab:appendix_dataset_statistics}
    \begin{tabular*}{\textwidth}{@{\extracolsep{\fill}}lccccc}
      \toprule
      \textbf{Category}
      & \multicolumn{1}{c}{\textbf{Single-hop QA}}
      & \multicolumn{3}{c}{\textbf{Multi-hop QA}}
      & \multicolumn{1}{c}{\textbf{Factual QA}} \\
      \cmidrule(lr){2-2} \cmidrule(lr){3-5} \cmidrule(l){6-6}
      &
      \textbf{NQ}
      & \textbf{HotpotQA}
      & \textbf{2WikiMultihopQA}
      & \textbf{MuSiQue}
      & \textbf{SimpleQA} \\
      \midrule
      \textit{Questions}
      & 1,000
      & 1,000
      & 1,000
      & 1,000
      & 1,000 \\
      \textit{Documents}
      & 9,633
      & 9,811
      & 6,119
      & 11,656
      & 20,686 \\
      \bottomrule
    \end{tabular*}
  \end{table*}
\endgroup

\textbf{Datasets.} We train encoder mode on the E5~\cite{E5} dataset following LLM2Vec~\cite{LLM2Vec}. For decoder mode, we construct a hop-aware mixed training set by sampling instances from the training splits of NaturalQuestions~\cite{NaturalQuestions}, HotpotQA~\cite{HotpotQA}, 2WikiMultihopQA~\cite{2WikiMultihopQA}, and MuSiQue~\cite{MuSiQue} according to hop count. Table ~\ref{tab:appendix_generation_statistics}  shows the training-set statistics. The final training set contains 10,000 examples: 2,000 1-hop, 6,000 2-hop, 500 3-hop, and 1,500 4-hop examples. Table ~\ref{tab:appendix_dataset_statistics} shows the test-set statistics for HotpotQA, 2WikiMultihopQA, MuSiQue, and SimpleQA~\cite{SimpleQA}. Each benchmark contains 1,000 questions, and \textit{Documents} denotes the size of the retrieval pool. We follow the test split configuration of RaLa~\cite{CoopRAG} and compare all baselines under identical conditions. 

\textbf{Baselines.} We evaluate ShiftRAG against four categories of baselines. Encoder-based retrievers include Contriever~\cite{Contriever}, GTR \cite{GTR}, ColBERTv2 \cite{ColBERTv2}, and RaLa \cite{CoopRAG}. LLM-based retrievers include GTE-Qwen2-7B-Instruct \cite{GTE}, LLM2Vec \cite{LLM2Vec}, and NV-Embed-v2 \cite{NV-Embed}. Unified LLM baselines include OneGen \cite{OneGen} and GritLM \cite{GritLM}. Latent-CoT baselines include Coconut \cite{Coconut}, CODI \cite{CODI}, and Sim-CoT \cite{Sim-CoT}. For fair comparison, we fine-tune LLM2Vec and OneGen using the same three backbones as ShiftRAG: Gemma2-9B \cite{Gemma2}, LlaMA3.1-8B \cite{LlaMA3}, and Mistral-7B \cite{Mistral}. We use Gemma2-9B as the default backbone because it achieves the strongest retrieval and QA performance for both ShiftRAG and the fine-tuned baselines among the tested LLM backbones. Unless otherwise specified, we report results using Gemma2-9B. Since GritLM requires full fine-tuning, we use the open-weight checkpoint released on Hugging Face without additional training. Latent-CoT methods are not originally designed for retrieval-based QA; therefore, their adapted results are reported separately in Appendix \ref{sec:appendix_LatentCoT}.

\textbf{Evaluation Metrics.} For evaluating ShiftRAG, we select the most commonly used retrieval, QA, embedding and generative benchmarks. For the retrieval task, to measure the ability to identify relevant information within a large-scale corpus, we use \textbf{Recall@$K$} (R@$K$) as the primary metric, where $K \in \{2, 5, 10\}$. It measures the proportion of instances where the ground-truth document is ranked within the top-$K$ retrieved results. This metric focuses on the capacity of the model to narrow down the search space accurately.

For the question answering task, to measure the precision of the model in generating correct answers based on provided contexts or internal knowledge, we employ two standard metrics: \textbf{Exact Match (EM)} measures the percentage of predictions that exactly match the ground-truth answers. \textbf{F1-score} evaluates the harmonic mean of precision and recall based on the word-level overlap between the predicted and ground truth answers, providing a broader assessment of semantic alignment.

Following the setup of GritLM \cite{GritLM} and the MTEB \cite{MTEB} benchmark, we also evaluate the quality of the learned embedding representations through three different tasks:
\begin{itemize}
    \item[(1)] \textbf{Clustering (Clust.)}: We perform K-means clustering on the text embeddings and assess the alignment between the cluster assignments and ground-truth labels using the \textbf{V-measure}.
    \item[(2)] \textbf{Semantic Textual Similarity (STS)}: We compute the cosine similarity between sentence embeddings and report the \textbf{Spearman correlation} with human-annotated similarity scores to verify how well the embeddings capture semantic relatedness.
    \item[(3)] \textbf{Summarization (Summ.)}: We measure the cosine similarity between the embeddings of original texts and their summaries. The performance is scored via \textbf{Spearman correlation} against human quality ratings to evaluate the model's ability to encode core information.
\end{itemize}

To assess reasoning and multilingual proficiency, we follow the generative evaluation framework of GritLM \cite{GritLM}, utilizing the following benchmarks:
\begin{itemize}
    \item[(1)] \textbf{MMLU}: We evaluate the model’s broad world knowledge across various disciplines (humanities, social sciences, etc.) in a Zero-shot (0 FS) setting, using \textbf{Exact Match (EM)} as the evaluation metric.
    \item[(2)] \textbf{GSM8K}: To test multi-step mathematical reasoning, we apply 8-shot (8 FS) prompting with Chain-of-Thought (CoT). The performance is measured by the \textbf{Exact Match (EM)} of the final numerical answer.
    \item[(3)] \textbf{TyDiQA}: We evaluate multilingual QA capabilities using the Gold Passage (GP) and zero-shot (0 FS) setup. Given the linguistic diversity of the tasks, we report the \textbf{F1-score} to assess the accuracy and robustness of the generated responses across different languages.
\end{itemize}

\textbf{Implementation Details.}  We train ShiftRAG on four NVIDIA RTX A6000 GPUs and run inference on a single GPU. Both encoder mode and decoder mode are trained with LoRA~\cite{LoRA} adapters. For the encoder, we employ a global batch size of 16 and a LoRA rank of $r=16$; for the decoder, we use a global batch size of 8 and $r=64$. The LGC loss is calculated with a bucket size of four layers and a temperature of $\tau=0.05$. The SOA loss is applied to the final $n=4$ candidate layers with a coefficient of $\lambda_{\text{SOA}}=1\times{10}^{-3}$. In decoder mode, the knowledge distillation weight is set to $\lambda_{\text{KD}}=0.75$. SST is configured to generate five soft knowledge tokens per question–document pair. Experiments on MTEB~\cite{MTEB}, MMLU~\cite{MMLU}, GSM8K~\cite{GSM8K}, and TyDiQA~\cite{TyDiQA} follow the protocol of GritLM~\cite{GritLM}. We evaluate our approach across several backbone LLMs, including Gemma2-9B~\cite{Gemma2}, LlaMA3.1-8B~\cite{LlaMA3}, Mistral-7B~\cite{Mistral}, and Qwen2.5-7B~\cite{Qwen2}.

\section{Warm-up Training}\label{sec:appendix_warmup}
\subsection{Encoder mode warm-up training}\label{sec:appendix_ret_warm_up}
Before optimizing the encoder-specific LoRA adapters, $\text{LoRA}_\text{enc}$ with the primary objective $\mathcal{L}_\text{enc}$, we perform a brief warm-up stage by using masked next-token prediction (MNTP), following LLM2Vec. Let $S=(x_1,…,x_{\vert S \vert})$ be an input token sequence, $\tilde S$ its masked version, and $M_S$ the set of masked positions. Under bidirectional attention, MNTP predicts each masked token $x_i$ using the final-layer token representation at the previous position $i-1$. This design preserves the autoregressive next-token prediction format inherent to decoder-only LLM pre-training:
\begin{equation}\label{eq:warmup_mntp}
\mathcal{L}_\text{MNTP}(S)= - \frac{1}{\vert M_S \vert} \displaystyle\sum_{i \in M_S} \log \Big[ \text{softmax}\big( W_{\text{LM}}\mathbf{z}_{\tilde S, i-1}^{(L)} \big)\Big]
\end{equation}
where $\mathbf{z}_{\tilde S, i-1}^{(L)} \in \mathbb R^h$ denotes the token representation of the final layer at position $i-1$ computed from $\tilde S$, and $W_\text{LM}$ represents the language modeling head. For $i=1$, the previous position corresponds to the \texttt{[BOS]} token. For a mini-batch $B$ of questions and the corresponding document set $D_B$, we average MNTP over both questions and documents:
\begin{equation}\label{eq:warmup_mntp_batch}
\mathcal{L}_\text{MNTP}(B, D_B)= \frac{1}{2}\Big( \frac{1}{\vert B \vert} \displaystyle\sum_{q \in B}\mathcal{L}_{\text{MNTP}}(q) + \frac{1}{\vert D_B \vert} \displaystyle\sum_{d \in D_B}\mathcal{L}_{\text{MNTP}}(d) \Big)
\end{equation}
The warm-up objective integrates thiw with the batch-averaged SOA penalty:
\begin{equation}\label{eq:warmup}
\mathcal{L}_{\text{warm}} = \mathcal{L}_{\text{MNTP}} (B, D_B) + \lambda_{\text{SOA}}\frac{\mathcal{L}_{\text{SOA}}(q) + \mathcal{L}_{\text{SOA}}(d)}{2}
\end{equation}
In our experimantal setup, out of a total training budget of 3,000 steps,  only the initial 1,000 steps utilize MNTP in place of the contrastive loss $\mathcal{L}_{\text{LGC}}$. In practice, the warm-up stage takes less than one hour.

\subsection{Anchor Token Initialization}\label{sec:appendix_intialization}
In our decoder-mode training, we initialize the anchor token embeddings as the pretrained latent vectors~\cite{SPRING} to align soft knowledge tokens with the token embedding space of the generator. For a question $q$ and a retrieved document $d_k$, SST constructs soft knowledge tokens $\mathbf{T}_k \in \mathbb{R}^{M \times h}$ by combining anchor token embeddings $\mathbf{E} \in \mathbb{R}^{M \times h}$ with document-conditioned translation embeddings and positional embeddings, as described in Section \ref{sec:reasoning}. Since the document-conditioned translation embeddings are derived from encoder-mode representations, the resulting soft knowledge tokens may not be well aligned with the token embedding space used by the generator. Following SPRING~\cite{SPRING}, we freeze the student LLM and optimize only the anchor token embeddings $\mathbf{E}$, which serve as the trainable basis of $\mathbf{T}_k$.

For each training question-document pair $(q,d_k)$, we use an offline-generated target rationale that explains whether $d_k$ helps answer the question and how it contributes to the answer. The rationale construction procedure and the prompt for anchor token initialization are provided in Appendices \ref{sec:appendix_prompt1} and \ref{sec:appendix_prompt2}, respectively. Given $d_k$, $q$, and $\mathbf{E}$, we form the initialization context as
\begin{equation}\label{eq:warmup}
X^\text{ATI} = [\text{Instruction}, d_k, q, \mathbf{E}]
\end{equation}
The frozen student LLM then autoregressively predicts the target rationale from $X^\text{ATI}$ using the standard next-token prediction (NTP) objective. During this stage, gradients are back-propagated only to $\mathbf{E}$, while all other parameters remain fixed. This procedure therefore does not tune the student LLM itself; it only learns a small set of anchor embeddings that can be plugged into the generator's token space.

This initialization is performed once before the decoder-mode training. By optimizing $\mathbf{E}$ with rationale supervision, this stage provides a vocabulary-aligned basis for SST-generated soft knowledge tokens and stabilizes the later translation of retrieval-side signals into the decoder. The initialized anchor embeddings do not encode document-specific knowledge by themselves; rather, they act as reference points that allow the document-conditioned translation embeddings to produce generation-friendly soft knowledge tokens. 
The target rationales are strictly reserved for the training pretraining anchor token embeddings and for teacher-side supervision employed by knowledge distillation.
Empirically, the optimization of anchor token embeddings $\mathbf{E}$ takes less than one hour.

\section{Overall Performance}\label{sec:appendix_results}
\subsection{Retrieval Performance}\label{sec:appendix_result_retrieval}
\begingroup
  \renewcommand{\arraystretch}{0.80}
  \setlength{\tabcolsep}{3pt}
  \begin{table*}[h]
    \centering
    \footnotesize
    \caption{Complete retrieval performance of ShiftRAG and baselines. The best and second-best performances are presented in bold and underlined, respectively. We conduct five runs with different random seeds and report the mean; the maximum standard deviation is ±0.05.}
    \label{tab:appendix_retrieval_performance}
    \begin{tabular*}{\textwidth}{@{\extracolsep{\fill}}lccccccccc}
      \toprule
      \multirow{2}{*}{\textbf{Methods}} 
      & \multicolumn{3}{c}{\textbf{HotpotQA}} 
      & \multicolumn{3}{c}{\textbf{2WikiMultihopQA}} 
      & \multicolumn{3}{c}{\textbf{MuSiQue}} \\
      \cmidrule(lr){2-4} \cmidrule(lr){5-7} \cmidrule(lr){8-10}
      & R@2 & R@5 & R@10 & R@2 & R@5 & R@10 & R@2 & R@5 & R@10 \\
      \midrule
      \multicolumn{10}{c}{\textit{Encoder-based Retrieval}} \\
      \midrule
      Contriever & 0.4789 & 0.6185 & 0.7214 & 0.5060 & 0.6349 & 0.6386 & 0.3319 & 0.4646 & 0.5648 \\
      GTR (T5-base) & 0.4913 & 0.6409 & 0.7379 & 0.5154 & 0.6256 & 0.6487 & 0.3314 & 0.4799 & 0.5754 \\
      ColBERTv2 & 0.5034 & 0.6971 & 0.7561 & 0.5460 & 0.6355 & 0.6565 & 0.3392 & 0.4823 & 0.5859 \\
      RaLa & 0.5485 & 0.6800 & 0.7415 & 0.5510 & 0.6563 & 0.7007 & 0.3530 & 0.5047 & 0.5882 \\
      \midrule
      \multicolumn{10}{c}{\textit{LLM-based Retrieval}} \\
      \midrule
      GTE-Qwen2-7B-Instruct & 0.5892 & 0.7636 & 0.8296 & 0.5474 & 0.6222 & 0.6586 & 0.3408 & 0.4630 & 0.5757 \\
      LLM2Vec [Mistral-7B] & 0.6085 & 0.7674 & 0.8403 & 0.5782 & 0.6393 & 0.6756 & 0.3571 & 0.4827 & 0.5901 \\
      LLM2Vec [LlaMA3.1-8B] & 0.6096 & 0.7642 & 0.8532 & 0.5985 & 0.6762 & 0.7080 & 0.3740 & 0.5005 & 0.5832 \\
      LLM2Vec [Qwen2.5-7B] & 0.6008 & 0.7660 & 0.8388 & 0.5778 & 0.6342 & 0.6780 & 0.3327 & 0.4580 & 0.5860 \\
      LLM2Vec [Gemma2-9B] & 0.6130 & 0.7854 & 0.8612 & 0.5845 & 0.6451 & 0.6829 & 0.3598 & 0.5126 & 0.5977 \\
      NV-Embed-v2 & 0.6318 & 0.7982 & 0.8736 & 0.5956 & 0.6632 & 0.7202 & 0.3781 & 0.5290 & 0.6236 \\
      \midrule
      \multicolumn{10}{c}{\textit{Unified LLM}} \\
      \midrule
      OneGen [Mistral-7B] & 0.5855 & 0.7442 & 0.8037 & 0.5525 & 0.6273 & 0.6935 & 0.3738 & 0.4960 & 0.5947 \\
      OneGen [LlaMA3.1-8B] & 0.5950 & 0.7553 & 0.8352 & 0.5950 & 0.6580 & 0.6964 & 0.3458 & 0.4772 & 0.5786 \\
      OneGen [Qwen2.5-7B] & 0.5898 & 0.7634 & 0.8360 & 0.5760 & 0.6326 & 0.6690 & 0.3221 & 0.4468 & 0.5770 \\
      OneGen [Gemma2-9B] & 0.5792 & 0.7450 & 0.8402 & 0.5584 & 0.6077 & 0.6392 & 0.3320 & 0.4777 & 0.5658 \\
      GritLM [Mistral-7B] & 0.5890 & 0.7487 & 0.8085 & 0.5558 & 0.6311 & 0.6977 & 0.3760 & 0.4990 & 0.5983 \\
      \midrule
      ShiftRAG [Mistral-7B] & 0.6235 & 0.7890 & 0.8536 & 0.5837 & 0.6683 & 0.7304 & 0.4019 & 0.5296 & 0.6320 \\
      ShiftRAG [LlaMA3.1-8B] & \textbf{0.6685} & \underline{0.8180} & \textbf{0.8825} & \underline{0.6025} & \underline{0.6960} & \underline{0.7438} & \textbf{0.4338} & \underline{0.5593} & \underline{0.6485} \\
      ShiftRAG [Qwen2.5-7B] & 0.6505 & 0.8020 & 0.8795 & 0.5837 & 0.6757 & 0.7188 & 0.3753 & 0.5039 & 0.5999 \\
      ShiftRAG [Gemma2-9B] & \underline{0.6598} & \textbf{0.8214} & \underline{0.8806} & \textbf{0.6080} & \textbf{0.7109} & \textbf{0.7530} & \underline{0.4186} & \textbf{0.5634} & \textbf{0.6516} \\
      \bottomrule
    \end{tabular*}
  \end{table*}
\endgroup

Table \ref{tab:appendix_retrieval_performance} shows the complete retrieval results, extending Table \ref{tab:retrieval_main} with encoder-based retrievers and backbone-wise results. ShiftRAG consistently outperforms encoder-based baselines by a large margin. Compared with RaLa~\cite{CoopRAG}, the strongest encoder-based retriever, ShiftRAG with Gemma2-9B improves HotpotQA R@2 by 20.29\% (0.5485→0.6598), 2WikiMultihopQA R@10 by 7.46\% (0.7007→0.7530), and MuSiQue R@2 by 18.58\% (0.3530→0.4186). These results indicate that encoder-mode training with LLM backbones learns a stronger semantic embedding space than conventional encoder-based retrievers. In backbone-matched comparisons, every ShiftRAG variant outperforms the corresponding LLM2Vec and OneGen variants across all datasets and metrics. With LlaMA3.1-8B, ShiftRAG improves HotpotQA R@2 by 9.66\% over LLM2Vec~\cite{LLM2Vec} (0.6096→0.6685) and MuSiQue R@2 by 25.45\% over OneGen (0.3458→0.4338). With Gemma2-9B, ShiftRAG improves 2WikiMultihopQA R@10 by 10.27\% over LLM2Vec (0.6829→0.7530) and 17.80\% over OneGen (0.6392→0.7530). These results show that the retrieval gains are robust across backbone choices. Among ShiftRAG variants, LlaMA3.1-8B achieves the best R@2 on HotpotQA and MuSiQue, while Gemma2-9B achieves the best performance on all 2WikiMultihopQA metrics and the strongest R@5/R@10 on MuSiQue. Overall, these results show that separating retrieval-side discriminative learning from autoregressive answer generation and connecting them through SST-generated soft knowledge tokens improves end-to-end QA more effectively than retrieval-only or unified LLM baselines.

\subsection{Question Answering Performance}\label{sec:appendix_result_generation}

\begingroup
  \renewcommand{\arraystretch}{0.80}
  \setlength{\tabcolsep}{3pt}

  \begin{table*}[h]
    \centering
    \footnotesize
    \caption{Complete QA performance of ShiftRAG and baselines. The best and second-best performances are presented in bold and underlined, respectively. We conduct five runs with different random seeds and report the mean; the maximum standard deviation is ±0.05.}
    \label{tab:appendix_qa_performance}
    \begin{tabular*}{\textwidth}{@{\extracolsep{\fill}}lcccccc}
      \toprule
      \multirow{2}{*}{\textbf{Methods}} 
      & \multicolumn{2}{c}{\textbf{HotpotQA}} 
      & \multicolumn{2}{c}{\textbf{2WikiMultihopQA}} 
      & \multicolumn{2}{c}{\textbf{MuSiQue}} \\
      \cmidrule(lr){2-3} \cmidrule(lr){4-5} \cmidrule(lr){6-7}
      & EM & F1 & EM & F1 & EM & F1 \\
      \midrule
      \multicolumn{7}{c}{\textit{Encoder-based Retrieval}} \\
      \midrule
      Contriever & 0.3660 & 0.4436 & 0.3310 & 0.3611 & 0.1990 & 0.2697 \\
      GTR (T5-base) & 0.3810 & 0.4611 & 0.3450 & 0.3705 & 0.2070 & 0.2804 \\
      ColBERTv2 & 0.4030 & 0.4829 & 0.3650 & 0.3965 & 0.2190 & 0.2968 \\
      RaLa & 0.4460 & 0.5400 & 0.4040 & 0.4397 & 0.2430 & 0.3284 \\
      \midrule
      \multicolumn{7}{c}{\textit{LLM-based Retrieval}} \\
      \midrule
      GTE-Qwen2-7B & 0.4790 & 0.5801 & 0.4340 & 0.4723 & 0.2320 & 0.3528 \\
      LLM2Vec [Mistral-7B] & 0.4400 & 0.5220 & 0.3010 & 0.3598 & 0.1920 & 0.2916 \\
      LLM2Vec [LlaMA3.1-8B] & 0.4240 & 0.5228 & 0.3260 & 0.4318 & 0.1800 & 0.2664 \\
      LLM2Vec [Qwen2.5-7B] & 0.4400 & 0.5365 & 0.3430 & 0.3928 & 0.1770 & 0.2776 \\
      LLM2Vec [Gemma2-9B] & 0.5030 & 0.6106 & 0.4400 & 0.4914 & 0.2430 & 0.3670 \\
      NV-Embed-v2 & 0.5140 & 0.6220 & 0.4580 & 0.5064 & 0.2480 & 0.3783 \\
      \midrule
      \multicolumn{7}{c}{\textit{Unified LLM}} \\
      \midrule
      OneGen [Mistral-7B] & 0.4200 & 0.5036 & 0.2820 & 0.3356 & 0.1770 & 0.2774 \\
      OneGen [LlaMA3.1-8B] & 0.4120 & 0.5096 & 0.2850 & 0.3860 & 0.1660 & 0.2532 \\
      OneGen [Qwen2.5-7B] & 0.4230 & 0.5082 & 0.2960 & 0.3478 & 0.1680 & 0.2596 \\
      OneGen [Gemma2-9B] & 0.4670 & 0.5869 & 0.4220 & 0.4785 & 0.2200 & 0.3342 \\
      GritLM [Mistral-7B] & 0.4430 & 0.5258 & 0.3020 & 0.3665 & 0.1870 & 0.2872 \\
      \midrule
      ShiftRAG [Mistral-7B] & 0.4670 & 0.5882 & 0.3250 & 0.4007 & 0.2180 & 0.3146 \\
      ShiftRAG [LlaMA3.1-8B] & 0.4630 & 0.5947 & 0.4200 & \underline{0.5124} & \underline{0.2360} & \underline{0.3602} \\
      ShiftRAG [Qwen2.5-7B] & \underline{0.4840} & \underline{0.6025} & \underline{0.4210} & 0.4876 & 0.2050 & 0.3324 \\
      ShiftRAG [Gemma2-9B] & \textbf{0.5610} & \textbf{0.6790} & \textbf{0.5080} & \textbf{0.5528} & \textbf{0.3050} & \textbf{0.4129} \\
      \bottomrule
    \end{tabular*}
  \end{table*}
\endgroup

Table \ref{tab:appendix_qa_performance} shows the complete QA results, extending Table \ref{tab:qa_main} with encoder-based retrievers and backbone-wise results. To ensure a controlled evaluation of generation capacity, all retrieval-only baselines are paired with Gemma2-9B, the strongest generator among the backbone LLMs. ShiftRAG with Gemma2-9B achieves the best overall QA performance across all datasets and metrics. Compared to RaLa~\cite{CoopRAG}, i.e., the strongest encoder-based retriever, ShiftRAG improves EM by 25.78\% on HotpotQA (0.4460 → 0.5610), 25.74\% on 2WikiMultihopQA (0.4040 → 0.5080), and 25.51\% on MuSiQue (0.2430 → 0.3050). Compared with NV-Embed-v2~\cite{NV-Embed}, i.e., the strongest LLM-based retrieval baseline, ShiftRAG improves EM by 9.14\% (0.5140 → 0.5610), 10.92\% (0.4580 → 0.5080), and 22.98\% (0.2480 → 0.3050), respectively. The backbone-matched results show the same trend. Every variant of ShiftRAG outperforms the corresponding LLM2Vec~\cite{LLM2Vec} and OneGen~\cite{OneGen} variants across all datasets and all metrics. For example, with LlaMA3.1-8B, ShiftRAG improves 2WikiMultihopQA EM by 28.83\% over LLM2Vec (0.3260 → 0.4200) and 47.37\% over OneGen (0.2850 → 0.4200). With Gemma2-9B, ShiftRAG improves HotpotQA EM by 11.53\% over LLM2Vec (0.5030 → 0.5610) and 20.13\% over OneGen (0.4670 → 0.5610). These results indicate that the QA gains are robust across varying backbone architectures. Among ShiftRAG variants, Gemma2-9B achieves the best QA performance on all datasets and all metrics. 
Collectively, our findings support the conclusion that decoupling retrieval-oriented discriminative learning from generative adaptation, and bridging them via SST-generated soft knowledge tokens, is more effective for end-to-end QA than either retrieval-only or unified LLM approaches.

\subsection{Retrieval and Question Answering Performance on Single-hop QA Benchmark}\label{sec:appendix_single-hopQA}

\begingroup
  \renewcommand{\arraystretch}{0.80}
  \setlength{\tabcolsep}{3pt}

  \begin{table*}[h]
    \centering
    \footnotesize
    \caption{Retrieval and QA performance on NaturalQuestions. The best and second-best performances are presented in bold and underlined, respectively.}
    \label{tab:appendix_nq}
    \begin{tabular*}{\textwidth}{@{\extracolsep{\fill}}lccccc}
      \toprule
      \textbf{Methods}
      & \multicolumn{3}{c}{\textbf{Retrieval}}
      & \multicolumn{2}{c}{\textbf{QA}} \\
      \cmidrule(lr){2-4} \cmidrule(l){5-6}
      &
      R@2 & R@5 & R@10
      & EM & F1 \\
      \midrule
      RaLa
      & 0.5923 & 0.8180 & 0.9145 & 0.3010 & 0.5036 \\
      LLM2Vec
      & 0.6006 & \underline{0.8217} & 0.9260 & 0.3150 & 0.5108 \\
      NV-Embed-v2
      & \underline{0.6078} & 0.8144 & \underline{0.9318} & \underline{0.3220} & \underline{0.5207} \\
      OneGen
      & 0.5648 & 0.7768 & 0.8746 & 0.2830 & 0.4983 \\
      GritLM
      & 0.5776 & 0.7789 & 0.9090 & 0.2900 & 0.5062 \\
      ShiftRAG
      & \textbf{0.6290} & \textbf{0.8660} & \textbf{0.9780}
      & \textbf{0.3350} & \textbf{0.5256} \\
      \bottomrule
    \end{tabular*}
  \end{table*}
\endgroup

Table \ref{tab:appendix_nq} shows retrieval and QA performance on NaturalQuestions~\cite{NaturalQuestions}, which evaluates whether ShiftRAG generalizes beyond multi-hop QA. ShiftRAG achieves the best performance across all retrieval and QA metrics. Compared with the strongest baseline for each metric, ShiftRAG improves Recall@2 by 3.49\% over NV-Embed-v2 (0.6078 → 0.6290), Recall@5 by 5.39\% over LLM2Vec~\cite{LLM2Vec} (0.8217 → 0.8660), and Recall@10 by 4.96\% over NV-Embed-v2~\cite{NV-Embed} (0.9318 → 0.9780). The same trend holds for QA, where ShiftRAG improves EM by 4.04\% (0.3220 → 0.3350) and F1 by 0.94\% (0.5207 → 0.5256) over NV-Embed-v2. These results indicate that the proposed training methods for the encoder and decoder modes are not limited to multi-hop reasoning, but remain effective in the single-hop setting.

\subsection{Question Answering Performance on SimpleQA}\label{sec:appendix_simple_qa}

\begin{table}[h]
  \centering
  \setlength{\tabcolsep}{6pt}
  \renewcommand{\arraystretch}{0.80}
  \footnotesize
  \caption{QA performance on the recent factual QA benchmark SimpleQA. The best and second-best performances are presented in bold and underlined, respectively.}
  \label{tab:appendix_simple_qa}
  \begin{tabular}{lcc}
    \toprule
    \textbf{Methods}
      & \multicolumn{2}{c}{\textbf{SimpleQA}} \\
    \cmidrule(lr){2-3}
      & EM & F1 \\
    \midrule
    RaLa   & 0.5450 & 0.6371 \\
    OneGen & 0.4970 & 0.5826 \\
    GritLM & \underline{0.5860} & \underline{0.7182} \\
    ShiftRAG   & \textbf{0.6740} & \textbf{0.8005} \\
    \bottomrule
  \end{tabular}
\end{table}

Table \ref{tab:appendix_simple_qa} shows QA performance on SimpleQA~\cite{SimpleQA}, which is excluded from training and used only at inference time. ShiftRAG achieves the best performance, reaching 0.6740 EM and 0.8005 F1-score. Compared with GritLM~\cite{GritLM}, i.e.g, the strongest baseline, ShiftRAG improves EM by 15.02\% ($0.5860\to 0.6740$) and F1-score by 11.46\% ($0.7182\to 0.8005$). These results indicate that ShiftRAG generalizes well to recent factual QA without task-specific training.

\subsection{Embedding and Generative Generalization}\label{sec:appendix_general_tasks}

\begingroup
  \setlength{\textfloatsep}{4pt}
  \renewcommand{\arraystretch}{0.80}

  \begin{table*}[h]
    \centering
    \footnotesize
    \caption{Embedding and generative generalization under the common Mistral-7B backbone. Left: performance of Clustering, STS, and Summarization on the MTEB benchmark. Right: generative performance on MMLU, GSM8K, and TyDiQA. The best and second-best performances are presented in bold and underlined, respectively.}
    \label{tab:appendix_general_tasks}

    \begin{minipage}[t]{0.49\textwidth}
      \centering
      \begin{tabular*}{\textwidth}{@{\extracolsep{\fill}}lccc}
        \toprule
        \multicolumn{4}{c}{\textbf{MTEB Performance}} \\
        \midrule
        Task $\rightarrow$
        & Clust.
        & STS
        & Summ. \\
        Metric $\rightarrow$
        & V-Meas.
        & Spear.
        & Spear. \\
        \# Dataset $\rightarrow$
        & 11
        & 10
        & 1 \\
        \midrule
        LLM2Vec
        & 0.4410 & \underline{0.8370} & 0.3105 \\
        GritLM
        & \underline{0.5082} & 0.8303 & \underline{0.3565} \\
        ShiftRAG
        & \textbf{0.5110} & \textbf{0.8414} & \textbf{0.3759} \\
        \bottomrule
      \end{tabular*}
    \end{minipage}
    \hfill
    \begin{minipage}[t]{0.49\textwidth}
      \centering
      \begin{tabular*}{\textwidth}{@{\extracolsep{\fill}}lccc}
        \toprule
        \multicolumn{4}{c}{\textbf{Generative Performance}} \\
        \midrule
        Dataset $\rightarrow$
        & MMLU
        & GSM8K
        & TyDiQA \\
        Metric $\rightarrow$
        & EM
        & EM
        & F1 \\
        Setup $\rightarrow$
        & 0 FS
        & 8FS, CoT
        & 0FS, GP \\
        \midrule
        Mistral-7B
        & 0.6018 & 0.4450 & \underline{0.5590} \\
        GritLM
        & \underline{0.5768} & \textbf{0.5762} & 0.5545 \\
        ShiftRAG
        & \textbf{0.6195} & 0.5536 & \textbf{0.5681} \\
        \bottomrule
      \end{tabular*}
    \end{minipage}
  \end{table*}
\endgroup

Table \ref{tab:appendix_general_tasks} reports the embedding and generative generalization performance of ShiftRAG, utilizing the Mistral-7B backbone to ensure a fair comparison with GritLM~\cite{GritLM}. To assess general embedding quality beyond retrieval, we evaluate ShiftRAG on MTEB~\cite{MTEB} task groups that do not directly measure the retrieval performance: Clustering, STS, and Summarization. ShiftRAG achieves state-of-the-art results across all three categories, improving over the strongest competitor by 0.55\% on Clustering ($0.5082 \to 0.5110$), 0.53\% on STS ($0.8370 \to 0.8414$), and 5.44\% on Summarization ($0.3565 \to 0.3759$). These results indicate that our encoder mode preserves and enhances semantic representation quality across diverse text embedding tasks.

For generative generalization, we evaluate MMLU~\cite{MMLU}, GSM8K~\cite{GSM8K}, and TyDiQA~\cite{TyDiQA} under the same protocol as GritLM. ShiftRAG outperforms the strongest baselines on MMLU and TyDiQA, with gains of 2.94\% ($0.6018\to 0.6195$) and 1.63\% ($0.5590\to 0.5681$), respectively. While ShiftRAG trails GritLM on GSM8K, it improves the base Mistral-7B performance by 24.40\% ($0.4450 \to 0.5536$), despite the absence of math-intensive training data. To adapt SST for these diverse tasks, we use the following inputs: (a) for MMLU, we input candidate-answer embeddings; (b) for GSM8K, we split each question into two segments and use their embeddings; (c) for TyDiQA, we use question–paragraph embeddings. Overall, ShiftRAG successfully retains its generative intelligence while simultaneously excelling as a high-performance retriever under the same backbone.

\subsection{Performance Comparison with Latent-CoT Baselines}\label{sec:appendix_LatentCoT}

\begingroup
  \renewcommand{\arraystretch}{0.80}
  \setlength{\tabcolsep}{3pt}

  \begin{table*}[h]
    \centering
    \footnotesize
    \caption{QA performance comparison with retrieval-adapted Latent-CoT baselines. The best and second-best performances are presented in bold and underlined, respectively.}
    \label{tab:appendix_latent_cot}
    \begin{tabular*}{\textwidth}{@{\extracolsep{\fill}}lcccccc}
      \toprule
      \textbf{Methods}
      & \multicolumn{2}{c}{\textbf{HotpotQA}}
      & \multicolumn{2}{c}{\textbf{2WikiMultihopQA}}
      & \multicolumn{2}{c}{\textbf{MuSiQue}} \\
      \cmidrule(lr){2-3} \cmidrule(lr){4-5} \cmidrule(l){6-7}
      &
      EM & F1
      & EM & F1
      & EM & F1 \\
      \midrule
      Coconut
      & 0.5160 & 0.6185
      & 0.4120 & 0.4404
      & 0.1895 & 0.2290 \\
      CODI
      & 0.5270 & 0.6254
      & 0.4210 & 0.4484
      & 0.1970 & 0.2413 \\
      Sim-CoT
      & \underline{0.5360} & \underline{0.6416}
      & \underline{0.4420} & \underline{0.5146}
      & \underline{0.2070} & \underline{0.2659} \\
      \midrule
      ShiftRAG
      & \textbf{0.5610} & \textbf{0.6790}
      & \textbf{0.5080} & \textbf{0.5528}
      & \textbf{0.3050} & \textbf{0.4129} \\
      \bottomrule
    \end{tabular*}
  \end{table*}
\endgroup

Table \ref{tab:appendix_latent_cot} compares ShiftRAG with Latent-CoT baselines adapted to the retrieval-based QA setting. Since Coconut~\cite{Coconut}, CODI~\cite{CODI}, and Sim-CoT~\cite{Sim-CoT} are originally designed to use only the question as input, we provide them with the same retrieved documents used by ShiftRAG during both training and inference. ShiftRAG achieves the best performance across all datasets and all metrics. Compared with Sim-CoT, i.e., the strongest Latent-CoT baseline, ShiftRAG improves EM by 14.93\% on 2WikiMultihopQA ($0.4420 \to 0.5080$) and 47.34\% on MuSiQue ($0.2070 \to 0.3050$). Since all methods use identical retrieved documents, the gains mainly come from how the retrieved knowledge is consumed during generation. These results indicate that soft knowledge tokens generated by SST provide a more effective interface for transferring retrieval-side discriminative knowledge than directly adapting Latent-CoT methods to retrieved text.

\subsection{Ablation Study: Semantic Signal Translator}\label{sec:appendix_ablation_SST}
\begingroup
  \renewcommand{\arraystretch}{0.80}
  \setlength{\tabcolsep}{3pt}

  \begin{table*}[h]
    \centering
    \footnotesize
    \caption{Ablation study on SST components. The best and second-best performances are presented in bold and underlined, respectively.}
    \label{tab:appendix_ablation_SST}
    \begin{tabular*}{\textwidth}{@{\extracolsep{\fill}}lcccccc}
      \toprule
      \textbf{Category}
      & \multicolumn{2}{c}{\textbf{HotpotQA}}
      & \multicolumn{2}{c}{\textbf{2WikiMultihopQA}}
      & \multicolumn{2}{c}{\textbf{MuSiQue}} \\
      \cmidrule(lr){2-3} \cmidrule(lr){4-5} \cmidrule(l){6-7}
      &
      EM & F1
      & EM & F1
      & EM & F1 \\
      \midrule
      Ours
      & \textbf{0.5610} & \textbf{0.6790}
      & \textbf{0.5080} & \textbf{0.5506}
      & \textbf{0.2960} & \textbf{0.4067} \\
      \textit{w/o} relevance weight
      & \underline{0.5570} & \underline{0.6765}
      & \underline{0.5060} & \underline{0.5480}
      & \underline{0.2910} & \underline{0.4025} \\
      \textit{w/o} low-rank projection layer
      & 0.5520 & 0.6754
      & 0.4970 & 0.5421
      & 0.2880 & 0.3976 \\
      \textit{w/o} anchor tokens
      & 0.5400 & 0.6533
      & 0.4890 & 0.5306
      & 0.2800 & 0.3705 \\
      \textit{w/o} SST
      & 0.5230 & 0.6468
      & 0.4630 & 0.5194
      & 0.2450 & 0.3639 \\
      \bottomrule
    \end{tabular*}
  \end{table*}
\endgroup

Table \ref{tab:appendix_ablation_SST} shows the effect of each SST component: \textit{w/o} relevance weight fixes the retrieval-score-based relevance weight $\alpha_k$ to 1, \textit{w/o} low-rank projection layer removes the low-rank projection layer and directly uses the output of the encoding layer as the offset embeddings, \textit{w/o} anchor tokens removes the anchor token embeddings, and \textit{w/o} SST removes the SST-generated soft knowledge tokens. Removing SST gives the largest degradation, reducing MuSiQue EM by 17.23\% ($0.2960 \to 0.2450$). Within SST, anchor tokens have the largest effect, followed by the low-rank projection layer and relevance weighting: removing anchor tokens reduces MuSiQue EM by 5.41\% ($0.2960 \to 0.2800$), removing the low-rank projection layer reduces it by 2.70\% ($0.2960 \to 0.2880$), whereas removing the relevance weight reduces it by 1.69\% ($0.2960 \to 0.2910$). These results indicate that anchor tokens provide a stable token-space basis for translating retrieval-oriented embeddings into generation-friendly soft knowledge tokens, the low-rank projection layer supports compact document-conditioned signal translation, and relevance weighting helps reflect the relative importance of retrieved documents during reasoning.

\subsection{Performance Comparison with Simple Projection Baselines}\label{sec:appendix_review3}
\begin{table}[h]
  \centering
  \setlength{\tabcolsep}{6pt}
  \renewcommand{\arraystretch}{0.80}
  \footnotesize
  \caption{QA performance comparison between SST and simple embedding-to-soft-token projection baselines. All methods use the same question--document representations and number of soft knowledge tokens, with other training and inference settings fixed.}
  \label{tab:appendix_sst_projection}
  \begin{tabular}{lcccccc}
    \toprule
    \textbf{Methods}
    & \multicolumn{2}{c}{\textbf{HotpotQA}}
    & \multicolumn{2}{c}{\textbf{2WikiMultihopQA}}
    & \multicolumn{2}{c}{\textbf{MuSiQue}} \\
    \cmidrule(lr){2-3}
    \cmidrule(lr){4-5}
    \cmidrule(lr){6-7}
    & EM & F1 & EM & F1 & EM & F1 \\
    \midrule
    1-layer MLP
    & 0.4505 & 0.5704
    & 0.4008 & 0.4572
    & 0.2376 & 0.3339 \\
    2-layer MLP
    & \underline{0.4651} & \underline{0.5894}
    & \underline{0.4125} & \underline{0.4595}
    & \underline{0.2446} & \underline{0.3355} \\
    ShiftRAG
    & \textbf{0.5610} & \textbf{0.6790}
    & \textbf{0.5080} & \textbf{0.5528}
    & \textbf{0.3050} & \textbf{0.4129} \\
    \bottomrule
  \end{tabular}
\end{table}
We investigate whether the benefits of SST can be achieved by a simpler embedding-to-soft-token projection. Table \ref{tab:appendix_sst_projection} compares SST with 1- and 2-layer MLPs, which directly project the same question–document representations into the same number of soft knowledge tokens. We replace only the translation module while keeping the remaining training and inference settings fixed. Unlike the component ablations in Appendix C.7, this comparison evaluates the SST design against alternative projection architectures.

ShiftRAG consistently achieves the best performance across all datasets and metrics. Although the 2-layer MLP outperforms the 1-layer MLP, both remain below SST. Compared with the 2-layer MLP, ShiftRAG improves EM by 20.62\% on HotpotQA (0.4651 → 0.5610) and 24.69\% on MuSiQue (0.2446 → 0.3050). These results indicate that directly projecting retrieval representations into soft tokens is less effective than the proposed SST design, supporting its role as an interface for translating retrieval-side discriminative signals into generation-friendly embeddings

\subsection{Impact of Causal-Mask Relaxation}\label{sec:appendix_review4}
\begin{table}[h]
  \centering
  \setlength{\tabcolsep}{6pt}
  \renewcommand{\arraystretch}{0.90}
  \footnotesize
  \caption{Effect of causal-mask relaxation on retrieval performance. Uni. and Bi. denote unidirectional and bidirectional attention in encoder mode, respectively. The best and second-best performances are presented in bold and underlined, respectively.}
  \label{tab:appendix_causal_mask}
  \resizebox{\linewidth}{!}{
  \begin{tabular}{lccccccccc}
    \toprule
    \textbf{Methods}
    & \multicolumn{3}{c}{\textbf{HotpotQA}}
    & \multicolumn{3}{c}{\textbf{2WikiMultihopQA}}
    & \multicolumn{3}{c}{\textbf{MuSiQue}} \\
    \cmidrule(lr){2-4}
    \cmidrule(lr){5-7}
    \cmidrule(lr){8-10}
    & R@2 & R@5 & R@10
    & R@2 & R@5 & R@10
    & R@2 & R@5 & R@10 \\
    \midrule
    RaLa
    & 0.5485 & 0.6800 & 0.7415
    & \underline{0.5510} & \underline{0.6563} & \underline{0.7007}
    & 0.3530 & 0.5047 & 0.5882 \\
    ShiftRAG (Uni.)
    & \underline{0.5938} & \underline{0.7475} & \underline{0.8190}
    & 0.5506 & 0.6440 & 0.6944
    & \underline{0.3818} & \underline{0.5127} & \underline{0.5909} \\
    ShiftRAG (Bi.)
    & \textbf{0.6598} & \textbf{0.8214} & \textbf{0.8806}
    & \textbf{0.6080} & \textbf{0.7109} & \textbf{0.7530}
    & \textbf{0.4186} & \textbf{0.5634} & \textbf{0.6516} \\
    \bottomrule
  \end{tabular}
  }
\end{table}

We analyze the effect of causal-mask relaxation on encoder-mode retrieval performance. Table \ref{tab:appendix_causal_mask} compares two ShiftRAG variants that differ only in their encoder-side attention policy: unidirectional attention retains the original causal mask, whereas bidirectional attention allows each token to attend to the entire input sequence. We also include RaLa as a dedicated bidirectional encoder baseline. The comparison between the two ShiftRAG variants isolates the contribution of attention policy within our encoder-mode training, complementing the qualitative analysis of their embedding distributions.

Bidirectional attention consistently improves retrieval performance across all datasets and metrics. Compared with unidirectional attention, it improves Recall@2 by 11.11\% on HotpotQA (0.5938 → 0.6598) and 10.43\% on 2WikiMultihopQA (0.5506 → 0.6080). ShiftRAG with bidirectional attention also outperforms RaLa across all reported metrics, improving HotpotQA Recall@5 by 20.79\% (0.6800 → 0.8214) and MuSiQue Recall@2 by 18.58\% (0.3530 → 0.4186). These results indicate that causal-mask relaxation contributes substantially to adapting the decoder-only backbone for evidence retrieval, supporting the use of bidirectional attention in encoder mode.

\section{Latency}\label{sec:appendix_latency}
\subsection{Training Latency}\label{sec:appendix_latency_training}

\begin{table}[h]
  \centering
  \setlength{\tabcolsep}{6pt}
  \renewcommand{\arraystretch}{0.80}
  \footnotesize
  \caption{Training latency and data size comparison between ShiftRAG and baselines. Time denotes end-to-end training time measured.}
  \label{tab:appendix_latency_train}
  \begin{tabular}{lcc}
    \toprule
    \textbf{Methods} & \textbf{Time} & \textbf{Data Size} \\
    \midrule
    LLM2Vec      & 5h 30m  & 96,000    \\
    OneGen       & 14h     & 174,500   \\
    GritLM       & 48h     & 2,886,912 \\
    ShiftRAG         & 10h 10m & 106,000   \\
    \bottomrule
  \end{tabular}
\end{table}

We analyze training cost to examine whether separated dual-mode training introduces substantial overhead. Table \ref{tab:appendix_latency_train} shows the measured end-to-end training time and the size of training data for ShiftRAG and baselines. LLM2Vec~\cite{LLM2Vec}, OneGen~\cite{OneGen}, and ShiftRAG are trained with LoRA, whereas GritLM~\cite{GritLM} requires full fine-tuning. ShiftRAG uses 96,000 retrieval examples and 10,000 generation examples, resulting in 106,000 training examples in total. Although ShiftRAG trains both encoder mode and decoder mode, it requires 10h 10m, which is shorter than OneGen (14h) and much shorter than GritLM (48h). Compared with LLM2Vec, ShiftRAG requires additional training time because it learns decoder mode as well, but yields stronger retrieval and QA performance. These results indicate that separate LoRA-based training provides a favorable trade-off between training efficiency and performance.

\subsection{Complexity Comparison with RaLa}\label{sec:appendix_Complexity}
We analyze the computational complexity and training memory of RaLa-style contrasting compared with our gap-aware relevance scoring. Following the notation in Section \ref{sec:train_retrieval}, let $\vert q \vert$ and $\vert d \vert$ denote the sequence lengths of a question and a document, respectively, $h$ denote the hidden dimension size, and $\vert \mathcal C \vert$ denote the number of candidate layers used for layer contrast. In the unoptimized RaLa formulation, layer contrast is computed over token-level MaxSim across all question–document token pairs and layers from each bucket. Therefore, its scoring complexity is $O(\vert \mathcal C \vert \cdot \vert q \vert \cdot \vert d \vert \cdot h)$. The optimized RaLa formulation reduces the layer-contrast computation to a \texttt{[CLS]}-based gap-aware weight, but its main matching score still relies on token-level average MaxSim. Thus, its asymptotic complexity becomes $O(\vert q \vert \cdot \vert d \vert \cdot h)$, which is asymptotically identical to ColBERT-style late interaction. In contrast, our gap-aware relevance scoring computes layer-wise semantic alignment gaps over mean-pooled question and document representations. Once the pooled representations are obtained, the scoring complexity is $O(\vert\mathcal C \vert \cdot h)$, and even when mean pooling is counted, the cost grows only linearly with the sequence length, rather than bilinearly with $\vert q \vert \vert d \vert$. Thus, our method removes the token-token interaction term required by average MaxSim while using the layer-wise semantic alignment gap itself as the contrastive similarity score in the LGC loss.

This difference is particularly important for LLM-based retrieval encoders. Since LLM hidden states have much larger hidden dimensions than conventional encoder backbones, applying average MaxSim to LLM hidden states substantially increases peak memory usage. In our preliminary training with a Gemma2-2B backbone, average MaxSim either caused out-of-memory errors or required reducing the per-device batch size from 8 to 2. To maintain the same global batch size, the gradient accumulation step had to be increased by $4\times$, resulting in roughly $16\times$ longer training time in our setup. Our mean-pooled gap-aware relevance scoring avoids this token-level MaxSim computation, allows training with the original per-device batch size, and more directly reflects layer-wise semantic alignment in the retrieval objective. We further speculate that token-level average MaxSim may be less suitable for decoder-only LLM encoders whose late-layer token representations already exhibit rank-collapse tendencies, because MaxSim assumes sufficiently discriminative token-level representations.

\subsection{Effectiveness-Efficiency Trade-off of Different Relevance Scoring Methods}\label{sec:appendix_scoring_tradeoff}

\begingroup
  \renewcommand{\arraystretch}{0.80}
  \setlength{\tabcolsep}{2pt}

  \begin{table*}[h]
    \centering
    \footnotesize
    \caption{Effectiveness-efficiency trade-off between cosine similarity and gap-aware relevance scoring. The best retrieval performances are presented in bold.}
    \label{tab:appendix_scoring_tradeoff}
    \begin{tabular*}{\textwidth}{@{\extracolsep{\fill}}lcccccccccccc}
      \toprule
      \multirow{2}{*}{\textbf{Scoring Methods}} 
      & \multicolumn{4}{c}{\textbf{HotpotQA}} 
      & \multicolumn{4}{c}{\textbf{2WikiMultihopQA}} 
      & \multicolumn{4}{c}{\textbf{MuSiQue}} \\
      \cmidrule(lr){2-5} \cmidrule(lr){6-9} \cmidrule(lr){10-13}
      & R@2 & R@5 & R@10 & Time (s) & R@2 & R@5 & R@10 & Time (s) & R@2 & R@5 & R@10 & Time (s) \\
      \midrule
      cosine & 0.6598 & 0.8214 & 0.8806 & 0.93 & 0.6080 & 0.7109 & 0.7530 & 0.96 & 0.4186 & \textbf{0.5634} & 0.6516 & 0.87 \\
      gap-aware & \textbf{0.6621} & \textbf{0.8275} & \textbf{0.8869} & 3.17 & \textbf{0.6099} & \textbf{0.7126} & \textbf{0.7571} & 3.73 & \textbf{0.4188} & 0.5632 & \textbf{0.6527} & 3.12 \\
      \bottomrule
    \end{tabular*}
  \end{table*}
\endgroup

We evaluate the trade-off between retrieval effectiveness and inference efficiency across different relevance scoring methods. Table \ref{tab:appendix_scoring_tradeoff} compares standard final-layer cosine similarity,  i.e.,  $s_{q, d} = \cos(\mathbf{e}_q,\mathbf{e}_d^{(L)})$, against the gap-aware relevance score defined in Equation \ref{eq:relevance_score}. The gap-aware metric yields marginal retrieval gains, improving HotpotQA R@10 by 0.72\% ($0.8806 \to 0.8869$) and 2WikiMultihopQA R@10 by 0.54\% ($0.7530\to 0.7571$). However, it incurs a substantial 3.63× increase ($0.92s\to 3.34s$) in average latency. Given that the performance gains are disproportionate to the added computational cost, we prioritize inference efficiency and adopt final-layer cosine similarity during inference.

\section{Analysis}\label{sec:appendix_analysis}
\subsection{Scalability Analysis}\label{sec:appendix_analysis_Scalability}

\begingroup
  \renewcommand{\arraystretch}{0.80}
  \setlength{\tabcolsep}{3pt}

  \begin{table*}[h]
    \centering
    \footnotesize
    \caption{Scalability analysis on R@10 under increasing candidate pool sizes. Relative gain is computed against each baseline at the same candidate size. The best and second-best performances are presented in bold and underlined, respectively.}
    \label{tab:appendix_analysis_scalability}
    \begin{tabular*}{\textwidth}{@{\extracolsep{\fill}}lcccccc}
      \toprule
      \textbf{Methods}
      & \multicolumn{6}{c}{\textbf{Candidate Pool Size}} \\
      \cmidrule(lr){2-7}
      & 10,000 & 20,000 & 30,000 & 40,000 & 50,000 & 60,000 \\
      \midrule
      RaLa
      & 0.4789 & 0.4574 & 0.4207 & 0.3781 & 0.3351 & 0.3235 \\
      GritLM
      & 0.5890 & 0.5708 & 0.5307 & 0.4824 & 0.4341 & 0.3901 \\
      NV-Embed-v2
      & \underline{0.6318} & \underline{0.6165} & \underline{0.5787} & \underline{0.5339} & \underline{0.4904} & \underline{0.4494} \\
      ShiftRAG
      & \textbf{0.7530} & \textbf{0.7410} & \textbf{0.6977} & \textbf{0.6516} & \textbf{0.6085} & \textbf{0.5654} \\
      \midrule \midrule
      Relative Gain over RaLa (\%)
      & 57.24 & 62.00 & 65.84 & 72.34 & 81.59 & 74.78 \\
      Relative Gain over GritLM (\%)
      & 27.84 & 29.82 & 31.47 & 35.07 & 40.18 & 44.94 \\
      Relative Gain over NV-Embed-v2 (\%)
      & 19.18 & 20.19 & 20.56 & 22.05 & 24.08 & 25.81 \\
      \bottomrule
    \end{tabular*}
  \end{table*}
\endgroup

\begin{equation}\label{eq:anchot_prompt}
\mathrm{Relative}\ \mathrm{Gain}\ (\%)\ =\ \Big(\mathrm{R@10}_{\mathrm{candidate\, size}}^{\mathrm{ShiftRAG}} - \mathrm{R@10}_{\mathrm{candidate\, size}}^{\mathrm{baseline}} \Big) \times 100
\end{equation}

We evaluate the scalability of ShiftRAG by increasing the candidate pool size from 10,000 to 60,000. Table \ref{tab:appendix_analysis_scalability} shows R@10 under each candidate pool size. Retrieval performance decreases for all methods as the search space becomes larger, but ShiftRAG consistently achieves the best performance at every size. The performance gap remains large under larger candidate pools. Compared with RaLa, ShiftRAG improves Recall@10 by 57.24\% at 10,000 candidates (0.4789 → 0.7530) and 74.78\% at 60,000 candidates (0.3235 → 0.5654). The gain over GritLM increases from 27.84\% to 44.94\%, and the gain over NV-Embed-v2 increases from 19.18\% to 25.81\%. These results indicate that our encoder mode learns robust discriminative embeddings and preserves stronger retrieval quality as the number of candidate documents increases.

\subsection{Impact of Anchor Token Initialization}\label{sec:appendix_anchor_initalization}

\begingroup
  \renewcommand{\arraystretch}{0.80}
  \setlength{\tabcolsep}{3pt}

  \begin{table*}[h]
    \centering
    \footnotesize
    \caption{Impact of anchor token initialization on QA performance. The best performances are presented in bold.}
    \label{tab:appendix_anchor_initialization}
    \begin{tabular*}{\textwidth}{@{\extracolsep{\fill}}lcccccc}
      \toprule
      \multirow{2}{*}{\textbf{Category}} 
      & \multicolumn{2}{c}{\textbf{HotpotQA}} 
      & \multicolumn{2}{c}{\textbf{2WikiMultihopQA}} 
      & \multicolumn{2}{c}{\textbf{MuSiQue}} \\
      \cmidrule(lr){2-3} \cmidrule(lr){4-5} \cmidrule(lr){6-7}
      & EM & F1 & EM & F1 & EM & F1 \\
      \midrule
      Random initialization & 0.5530 & 0.6635 & 0.4960 & 0.5412 & 0.2990 & 0.4020 \\
      Anchor token initialization & \textbf{0.5610} & \textbf{0.6790} & \textbf{0.5080} & \textbf{0.5528} & \textbf{0.3050} & \textbf{0.4129} \\
      \bottomrule
    \end{tabular*}
  \end{table*}
\endgroup

We analyze the effect of anchor token initialization in SST. Table \ref{tab:appendix_anchor_initialization} compares random initialization with the anchor token initialization procedure described in Appendix \ref{sec:appendix_intialization}. Anchor token initialization consistently improves QA performance, increasing EM by 1.45\% on HotpotQA (0.5530→0.5610), 2.42\% on 2WikiMultihopQA (0.4960→0.5080), and 2.01\% on MuSiQue (0.2990→0.3050). The model remains competitive even with random initialization, suggesting that anchor token initialization mainly serves as an auxiliary mechanism for improving optimization stability. These results indicate that pre-aligning anchor tokens with the generator’s token space helps SST produce more stable soft knowledge tokens.

\subsection{Impact of SST on Confidence Calibration and LoRA Adaptation}\label{sec:appendix_SST_confidence}

\begingroup
  \renewcommand{\arraystretch}{0.80}
  \setlength{\tabcolsep}{3pt}

  \begin{table*}[h]
    \centering
    \footnotesize
    \caption{QA performance and confidence statistics under different uses of SST. “Generate \textit{w/o} SST” removes SST only at inference after SST-based training, whereas “Train \textit{w/o} SST” trains the generator without SST. The best and second-best performances are presented in bold and underlined, respectively.}
    \label{tab:appendix_SST_confidence}
    \begin{tabular*}{\textwidth}{@{\extracolsep{\fill}}lccccccc}
      \toprule
      \multirow{2}{*}{\textbf{Methods}} 
      & \multirow{2}{*}{\textbf{HotpotQA}} 
      & \multirow{2}{*}{\textbf{2WikiMultihopQA}} 
      & \multirow{2}{*}{\textbf{MuSiQue}} 
      & \multicolumn{2}{c}{\textbf{Correct}} 
      & \multicolumn{2}{c}{\textbf{Incorrect}} \\
      \cmidrule(lr){5-6} \cmidrule(lr){7-8}
      & & & & Entropy & Top-1 Prob. & Entropy & Top-1 Prob. \\
      \midrule
      {ShiftRAG} & \textbf{0.5610} & \textbf{0.5080} & \textbf{0.3050} & 0.4187 & 0.8343 & 1.3430 & 0.5091 \\
      Generate \textit{w/o} SST & \underline{0.5490} & \underline{0.4980} & \underline{0.2890} & 0.3869 & 0.8804 & 1.0063 & 0.6762 \\
      Train \textit{w/o} SST & 0.5230 & 0.4630 & 0.2450 & 0.2204 & 0.8831 & 0.4102 & 0.8455 \\
      Vanilla Backbone & 0.5020 & 0.4400 & 0.1970 & 0.1492 & 0.9186 & 0.1899 & 0.8957 \\
      \bottomrule
    \end{tabular*}
  \end{table*}
\endgroup

We analyze the impact of SST on QA performance and generator confidence. Motivated by the overconfidence observation in Figure 1, Table \ref{tab:appendix_SST_confidence} compares four settings: Vanilla Backbone, Train \textit{w/o} SST, Generate \textit{w/o} SST, and ShiftRAG. QA performance consistently improves as SST is more fully incorporated, with ShiftRAG achieving the highest EM on all datasets. At the same time, SST makes the generator less overconfident on incorrect answers rather than uniformly uncertain. In particular, the top-1 probability on incorrect answers decreases from 0.8957 in Vanilla Backbone to 0.5091 in ShiftRAG, while the entropy increases from 0.1899 to 1.3430. This produces a clearer confidence separation between correct and incorrect generations: the entropy gap increases from 0.0407 in Vanilla Backbone and 0.1898 in Train \textit{w/o} SST to 0.6194 in Generate \textit{w/o} SST and 0.9243 in ShiftRAG. These results indicate that SST mitigates premature commitment to incorrect answers and encourages correctness-aware decoding. Moreover, Generate \textit{w/o} SST still outperforms Train \textit{w/o} SST on all datasets, even though neither setting uses SST at inference, suggesting that SST not only provides auxiliary soft knowledge tokens at test time but also reshapes the generator-side LoRA adaptation during training.

\subsection{Impact of SST with Fixed External Retrievers}\label{sec:appendix_review2}
\begin{table}[h]
  \centering
  \setlength{\tabcolsep}{6pt}
  \renewcommand{\arraystretch}{0.80}
  \footnotesize
  \caption{QA performance with fixed external retrievers. Both inference settings use decoder-mode LoRA trained with SST and KD, while w/ SST additionally provides soft knowledge tokens at inference. The best performance within each retriever is presented in bold.}
  \label{tab:appendix_external_retriever_sst}
  \begin{tabular}{llcccccc}
    \toprule
    \textbf{Retriever} & \textbf{Inference}
    & \multicolumn{2}{c}{\textbf{HotpotQA}}
    & \multicolumn{2}{c}{\textbf{2WikiMultihopQA}}
    & \multicolumn{2}{c}{\textbf{MuSiQue}} \\
    \cmidrule(lr){3-4}
    \cmidrule(lr){5-6}
    \cmidrule(lr){7-8}
    & & EM & F1 & EM & F1 & EM & F1 \\
    \midrule
    LLM2Vec & w/o SST
    & 0.5030 & 0.6106 & 0.4400 & 0.4914 & 0.2430 & 0.3670 \\
    LLM2Vec & w/ SST
    & \textbf{0.5220} & \textbf{0.6281}
    & \textbf{0.4630} & \textbf{0.5202}
    & \textbf{0.2610} & \textbf{0.3902} \\
    \midrule
    NV-Embed-v2 & w/o SST
    & 0.5140 & 0.6220 & 0.4580 & 0.5064 & 0.2480 & 0.3783 \\
    NV-Embed-v2 & w/ SST
    & \textbf{0.5300} & \textbf{0.6392}
    & \textbf{0.4690} & \textbf{0.5251}
    & \textbf{0.2630} & \textbf{0.3913} \\
    \bottomrule
  \end{tabular}
\end{table}

We investigate whether the effectiveness of SST is restricted to the representations learned by our encoder. Table \ref{tab:appendix_external_retriever_sst} evaluates SST with two external retrievers, LLM2Vec and NV-Embed-v2. For each retriever, we compare generation with and without SST-generated soft knowledge tokens while keeping retrieval fixed. Both settings use decoder-mode LoRA trained with SST and KD; therefore, w/o SST removes the soft knowledge tokens only at inference, rather than excluding SST from training. This comparison examines whether retrieval-side signals remain useful when they originate from encoders other than the proposed encoder.

SST consistently improves EM and F1 across all three datasets with both external retrievers. On MuSiQue, SST improves EM by 7.41\% with LLM2Vec (0.2430 → 0.2610) and 6.05\% with NV-Embed-v2 (0.2480 → 0.2630). Since the retriever remains fixed within each comparison, these gains reflect the benefit of supplying retrieval-side latent signals during generation rather than improvements in document retrieval. These results indicate that SST can complement raw textual evidence using representations from different retrievers, extending its applicability beyond the embedding space learned by ShiftRAG.

\subsection{Theoretical Interpretation of Learning Strategy}\label{appendix_analysis_learning_strategy}
\begingroup
  \renewcommand{\arraystretch}{0.80}
  \setlength{\tabcolsep}{3pt}

  \begin{table*}[h]
    \centering
    \footnotesize
    \caption{Impact of learning strategies and parameter sharing on retrieval, QA, and training time. The best and second-best performances are presented in bold and underlined, respectively. The Parameter column indicates whether retrieval and generation adaptation parameters are shared or separate.}
    \label{tab:appendix_training_methods_analysis}
    \begin{tabular*}{\textwidth}{@{\extracolsep{\fill}}llccccccccc}
      \toprule
      &
      &
      \multicolumn{4}{c}{\textbf{HotpotQA}}
      & \multicolumn{4}{c}{\textbf{2WikiMultihopQA}}
      & \\
      \cmidrule(lr){3-6} \cmidrule(lr){7-10}
      \textbf{Training Methods}
      & \textbf{Parameter}
      & \multicolumn{2}{c}{\textbf{Retrieval}}
      & \multicolumn{2}{c}{\textbf{Generation}}
      & \multicolumn{2}{c}{\textbf{Retrieval}}
      & \multicolumn{2}{c}{\textbf{Generation}}
      & \textbf{Time (h)} \\
      \cmidrule(lr){3-4} \cmidrule(lr){5-6}
      \cmidrule(lr){7-8} \cmidrule(lr){9-10}
      &
      &
      R@2 & R@5
      & EM & F1
      & R@2 & R@5
      & EM & F1
      & \\
      \midrule
      \multirow{2}{*}{Joint Learning}
      & shared
      & 0.4896 & 0.6298
      & 0.1300 & 0.3430
      & 0.3230 & 0.4666
      & 0.1270 & 0.3432
      & 21 h \\
      & separate
      & 0.5264 & 0.6865
      & 0.4730 & 0.5885
      & 0.5159 & 0.6184
      & 0.4360 & 0.5115
      & 23 h \\
      \midrule
      \multirow{2}{*}{Alternating Learning}
      & shared
      & 0.5166 & 0.6634
      & 0.2200 & 0.4036
      & 0.4923 & 0.5012
      & 0.1860 & 0.4172
      & 20 h \\
      & separate
      & \underline{0.5938} & \underline{0.7665}
      & \underline{0.5220} & \underline{0.6451}
      & \underline{0.5654} & \underline{0.6682}
      & \underline{0.4780} & \underline{0.5409}
      & \textbf{10 h} \\
      \midrule
      Separate Learning
      & separate
      & \textbf{0.6598} & \textbf{0.8214}
      & \textbf{0.5610} & \textbf{0.6790}
      & \textbf{0.6080} & \textbf{0.7109}
      & \textbf{0.5080} & \textbf{0.5528}
      & \textbf{10 h} \\
      \bottomrule
    \end{tabular*}
  \end{table*}
\endgroup

We analyze why Separate Learning provides a more stable optimization strategy than Joint Learning or Alternating Learning. Table~\ref{tab:appendix_training_methods_analysis} compares these strategies under shared and separate LoRA parameter settings. Joint Learning with shared parameters is the most unstable configuration, causing severe QA collapse with EM dropping to 0.1300 on HotpotQA and 0.1270 on 2WikiMultihopQA. Separating LoRA parameters mitigates this collapse, and Alternating Learning~\cite{AlternatingLearning1,AlternatingLearning2,AlternatingLearning3} further reduces simultaneous objective interference by updating only one mode at each step. However, both remain below Separate Learning. In particular, Separate Learning achieves the best retrieval and QA performance with the same 10~h average training time as Alternating Learning with separate parameters. These results suggest that stable dual-mode training requires two conditions: encoder and decoder modes should not compete over the same adaptation parameters, and decoder-mode training should not depend on a moving retrieval-conditioned latent interface.

We first formalize the effect of parameter sharing. Let \(\mathbf{w}\) denote the vectorized shared LoRA parameters, and let \(\mathcal{L}_{\mathrm{enc}}\) and \(\mathcal{L}_{\mathrm{dec}}\) denote the encoder-mode and decoder-mode objectives, respectively. At step \(t\), Joint Learning updates the shared parameters as
\begin{equation}
\label{eq:joint_update_shared_lora}
\mathbf{w}_{t+1}
=
\mathbf{w}_t
-
\eta
\bigl(
\mathbf{g}_t^{\mathrm{enc}}
+
\lambda_J \mathbf{g}_t^{\mathrm{dec}}
\bigr),
\end{equation}
where
\begin{equation}
\label{eq:joint_gradient_definitions}
\mathbf{g}_t^{\mathrm{enc}}
=
\nabla_{\mathbf{w}}\mathcal{L}_{\mathrm{enc}}(\mathbf{w}_t),
\qquad
\mathbf{g}_t^{\mathrm{dec}}
=
\nabla_{\mathbf{w}}\mathcal{L}_{\mathrm{dec}}(\mathbf{w}_t).
\end{equation}
Here, \(\lambda_J>0\) controls the relative contribution of the decoder-mode objective.

\begin{lemma}[Gradient conflict under smoothness]
\label{lem:gradient_conflict_smoothness}
Assume that \(\mathcal{L}_{\mathrm{enc}}\) is \(\beta_{\mathrm{enc}}\)-smooth with respect to \(\mathbf{w}\). For the joint update in Equation~\ref{eq:joint_update_shared_lora}, the one-step change of the encoder-mode objective is bounded by

\begin{equation}
\label{eq:encoder_smoothness_bound}
\begin{aligned}
\mathcal{L}_{\mathrm{enc}}(\mathbf{w}_{t+1})
\le
\mathcal{L}_{\mathrm{enc}}(\mathbf{w}_t)
&-
\eta
\lVert
\mathbf{g}_t^{\mathrm{enc}}
\rVert^2
-
\eta\lambda_J
\langle
\mathbf{g}_t^{\mathrm{enc}},
\mathbf{g}_t^{\mathrm{dec}}
\rangle +
\frac{
\beta_{\mathrm{enc}}\eta^2
}{2}
\lVert
\mathbf{g}_t^{\mathrm{enc}}
+
\lambda_J \mathbf{g}_t^{\mathrm{dec}}
\rVert^2 .
\end{aligned}
\end{equation}
\end{lemma}

\noindent\textbf{Proof.}
By \(\beta_{\mathrm{enc}}\)-smoothness~\cite{beta-smoothness1, beta-smoothness2},
\begin{equation}
\label{eq:encoder_smoothness_inequality}
\mathcal{L}_{\mathrm{enc}}(\mathbf{w}_{t+1})
\le
\mathcal{L}_{\mathrm{enc}}(\mathbf{w}_t)
+
\langle
\nabla_{\mathbf{w}}\mathcal{L}_{\mathrm{enc}}(\mathbf{w}_t),
\mathbf{w}_{t+1}-\mathbf{w}_t
\rangle
+
\frac{\beta_{\mathrm{enc}}}{2}
\lVert
\mathbf{w}_{t+1}-\mathbf{w}_t
\rVert^2 .
\end{equation}
Substituting \(\nabla_{\mathbf{w}}\mathcal{L}_{\mathrm{enc}}(\mathbf{w}_t)=\mathbf{g}_t^{\mathrm{enc}}\) and \(\mathbf{w}_{t+1}-\mathbf{w}_t=-\eta(\mathbf{g}_t^{\mathrm{enc}}+\lambda_J\mathbf{g}_t^{\mathrm{dec}})\) gives Equation~\ref{eq:encoder_smoothness_bound}.

Lemma~\ref{lem:gradient_conflict_smoothness} does not assume that encoder-mode and decoder-mode gradients always conflict. Rather, it shows that conflict becomes harmful when the two gradients are negatively aligned. Let
\begin{equation}
\label{eq:gradient_cosine_alignment}
\cos\theta_t
=
\frac{
\langle
\mathbf{g}_t^{\mathrm{enc}},
\mathbf{g}_t^{\mathrm{dec}}
\rangle
}{
\lVert
\mathbf{g}_t^{\mathrm{enc}}
\rVert
\lVert
\mathbf{g}_t^{\mathrm{dec}}
\rVert
}.
\end{equation}
Then the first-order term in Equation~\ref{eq:encoder_smoothness_bound} can be written as
\begin{equation}
\label{eq:first_order_joint_term}
-
\eta
\lVert
\mathbf{g}_t^{\mathrm{enc}}
\rVert
\bigl(
\lVert
\mathbf{g}_t^{\mathrm{enc}}
\rVert
+
\lambda_J
\lVert
\mathbf{g}_t^{\mathrm{dec}}
\rVert
\cos\theta_t
\bigr).
\end{equation}
When \(\cos\theta_t<0\), the decoder-mode gradient weakens the descent direction for the encoder-mode objective. If \(\lambda_J\lVert\mathbf{g}_t^{\mathrm{dec}}\rVert\lvert\cos\theta_t\rvert\) becomes comparable to \(\lVert\mathbf{g}_t^{\mathrm{enc}}\rVert\), the first-order descent term can be largely canceled, while the smoothness term remains non-negative. The same argument applies symmetrically to \(\mathcal{L}_{\mathrm{dec}}\). Thus, shared-parameter Joint Learning can fail when discriminative retrieval adaptation and autoregressive generation adaptation push the same LoRA parameters in different directions.

We empirically verify this conflict by measuring the cosine similarity between \(\mathbf{g}_t^{\mathrm{enc}}\) and \(\mathbf{g}_t^{\mathrm{dec}}\) on the shared LoRA parameters. We summarize the conflict frequency using the negative-gradient ratio
\begin{equation}
\label{eq:negative_gradient_ratio}
\rho_{\mathrm{neg}}
=
\frac{1}{T}
\sum_{t=1}^{T}
\mathbb{I}[\cos\theta_t < 0].
\end{equation}
Over the first 500 updates, the cosine similarity is negative in 84.2\% of steps on average across three runs. This explains why Joint Learning with shared parameters shows the largest degradation in Table~\ref{tab:appendix_training_methods_analysis}.

Parameter separation removes this direct gradient conflict, but it does not fully stabilize decoder-mode training. Even with separate \(\mathrm{LoRA}_{\mathrm{enc}}\) and \(\mathrm{LoRA}_{\mathrm{dec}}\), the decoder-mode objective depends on SST-generated soft knowledge tokens, which are produced from encoder-side representations. Let
\begin{equation}
\label{eq:soft_token_interface}
\mathbf{T}_t
=
\mathbf{T}_{1:K}(\mathbf{w}_t^{\mathrm{enc}})
\end{equation}
denote the stacked soft knowledge tokens generated from the current encoder-mode parameters. For notational clarity, we fix the input examples and SST parameters, and isolate only the drift induced by encoder-mode updates. The decoder-mode objective can then be written as \(\mathcal{L}_{\mathrm{dec}}(\mathbf{w}_t^{\mathrm{dec}};\mathbf{T}_t)\). Hence, if encoder mode keeps changing during joint or alternating training, the generator is optimized against a latent interface whose distribution also changes.

\begin{lemma}[Encoder-side updates induce decoder-objective drift]
\label{lem:encoder_side_drift}
Assume that \(\mathcal{L}_{\mathrm{dec}}(\mathbf{w}^{\mathrm{dec}};\mathbf{T})\) is \(L_T\)-Lipschitz in \(\mathbf{T}\), and that \(\mathbf{T}_{1:K}(\mathbf{w}^{\mathrm{enc}})\) is \(L_w\)-Lipschitz in \(\mathbf{w}^{\mathrm{enc}}\). Then the decoder-objective drift caused by an encoder-side update satisfies
\begin{equation}
\label{eq:decoder_objective_drift}
\lvert
\mathcal{L}_{\mathrm{dec}}
(
\mathbf{w}_t^{\mathrm{dec}};\mathbf{T}_{t+1}
)
-
\mathcal{L}_{\mathrm{dec}}
(
\mathbf{w}_t^{\mathrm{dec}};\mathbf{T}_t
)
\rvert
\le
L_TL_w
\lVert
\mathbf{w}_{t+1}^{\mathrm{enc}}
-
\mathbf{w}_t^{\mathrm{enc}}
\rVert .
\end{equation}
\end{lemma}

\noindent\textbf{Proof.} Since \(\mathcal{L}_{\mathrm{dec}}\) is \(L_T\)-Lipschitz in \(\mathbf{T}\)~\cite{Lipschitz1,Lipschitz2},
\begin{equation}
\label{eq:decoder_lipschitz_in_tokens}
\lvert
\mathcal{L}_{\mathrm{dec}}
(
\mathbf{w}_t^{\mathrm{dec}};\mathbf{T}_{t+1}
)
-
\mathcal{L}_{\mathrm{dec}}
(
\mathbf{w}_t^{\mathrm{dec}};\mathbf{T}_t
)
\rvert
\le
L_T
\lVert
\mathbf{T}_{t+1}
-
\mathbf{T}_t
\rVert .
\end{equation}
Since \(\mathbf{T}_{1:K}(\mathbf{w}^{\mathrm{enc}})\) is \(L_w\)-Lipschitz in \(\mathbf{w}^{\mathrm{enc}}\), we have
\begin{equation}
\label{eq:token_lipschitz_in_encoder}
\lVert
\mathbf{T}_{t+1}
-
\mathbf{T}_t
\rVert
\le
L_w
\lVert
\mathbf{w}_{t+1}^{\mathrm{enc}}
-
\mathbf{w}_t^{\mathrm{enc}}
\rVert .
\end{equation}
Combining Equations~\ref{eq:decoder_lipschitz_in_tokens} and~\ref{eq:token_lipschitz_in_encoder} gives Equation~\ref{eq:decoder_objective_drift}. 

Lemma~\ref{lem:encoder_side_drift} shows that parameter separation alone does not guarantee a stationary decoder-mode objective. If the encoder-mode parameters are updated as
\begin{equation}
\label{eq:encoder_update_step}
\mathbf{w}_{t+1}^{\mathrm{enc}}
=
\mathbf{w}_t^{\mathrm{enc}}
-
\eta_{\mathrm{enc}}
\mathbf{g}_t^{\mathrm{enc}},
\end{equation}
then Equation~\ref{eq:decoder_objective_drift} becomes
\begin{equation}
\label{eq:single_step_interface_drift}
\lvert
\mathcal{L}_{\mathrm{dec}}
(
\mathbf{w}_t^{\mathrm{dec}};\mathbf{T}_{t+1}
)
-
\mathcal{L}_{\mathrm{dec}}
(
\mathbf{w}_t^{\mathrm{dec}};\mathbf{T}_t
)
\rvert
\le
\eta_{\mathrm{enc}}L_TL_w
\lVert
\mathbf{g}_t^{\mathrm{enc}}
\rVert .
\end{equation}
Thus, larger encoder-mode updates can induce larger changes in the decoder-mode objective, even when the two modes use separate LoRA parameters. Over multiple encoder-mode updates, this drift accumulates as
\begin{equation}
\label{eq:multi_step_interface_drift}
\lvert
\mathcal{L}_{\mathrm{dec}}
(
\mathbf{w}_t^{\mathrm{dec}};\mathbf{T}_{t+m}
)
-
\mathcal{L}_{\mathrm{dec}}
(
\mathbf{w}_t^{\mathrm{dec}};\mathbf{T}_t
)
\rvert
\le
L_TL_w
\sum_{i=t}^{t+m-1}
\lVert
\mathbf{w}_{i+1}^{\mathrm{enc}}
-
\mathbf{w}_i^{\mathrm{enc}}
\rVert .
\end{equation}

This explains why Joint Learning with separate parameters remains suboptimal. Although direct gradient conflict is reduced, the generator is still trained under a non-stationary SST interface. Alternating Learning mitigates simultaneous objective interference by updating only one mode at a time, but it does not remove the drift because \(\mathbf{T}_{1:K}\) can still change after encoder-mode updates. Separate Learning removes this source of instability by fixing \(\mathbf{w}^{\mathrm{enc}}\) before decoder-mode training. In this case, encoder-induced changes satisfy \(\mathbf{w}_{t+1}^{\mathrm{enc}}=\mathbf{w}_t^{\mathrm{enc}}\), so \(\mathbf{T}_{t+1}=\mathbf{T}_t\), and the drift term in Lemma~\ref{lem:encoder_side_drift} becomes zero.

This interpretation matches the empirical trend in Table~\ref{tab:appendix_training_methods_analysis}. Joint Learning with shared parameters performs worst, consistent with the gradient conflict in Lemma~\ref{lem:gradient_conflict_smoothness} and the measured negative-gradient ratio. Joint Learning with separate parameters improves over the shared-parameter variant, but remains below Alternating Learning and Separate Learning because the generator still follows a moving soft-token interface. Alternating Learning with separate parameters reduces simultaneous updates, but still allows encoder-side drift between decoder-mode updates. In contrast, Separate Learning first learns stable encoder-mode representations and then trains decoder mode with a fixed retrieval-conditioned interface. Compared with Alternating Learning with separate parameters, Separate Learning improves HotpotQA R@2 by 11.11\% (\(0.5938 \rightarrow 0.6598\)) and EM by 7.53\% (\(0.5220 \rightarrow 0.5610\)), with the same 10~h average training time. Overall, these results support our design choice of using separate adaptation paths and stage-wise training for encoder and decoder modes.

\subsection{Overconfidence Analysis on Correct and Incorrect Cases}\label{appendix_analysis_overconfidence}
\begin{figure*}[h]
    \centering
    \includegraphics[width=1\textwidth]{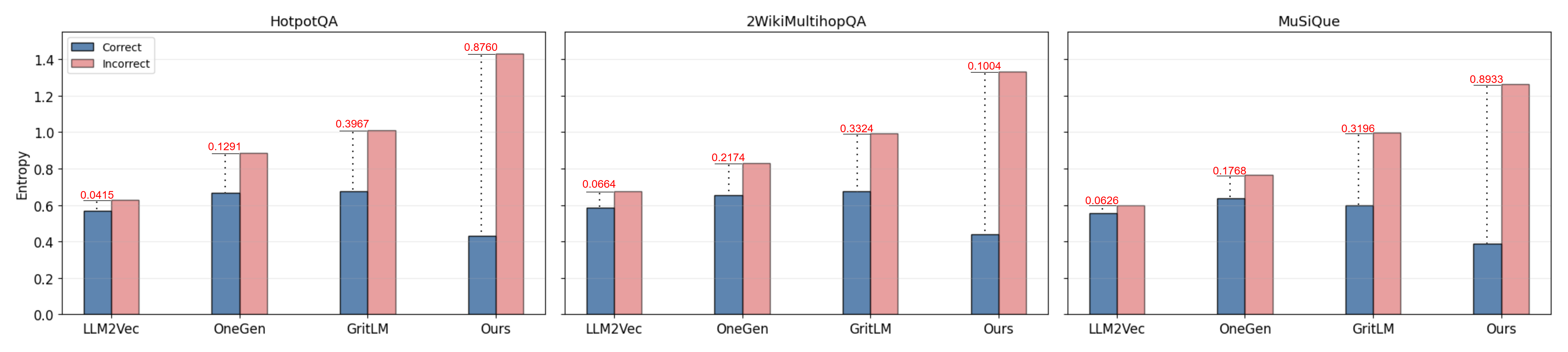}
    \caption{Average token entropy over the generated answer on MHQA benchmarks, grouped by answer correctness. Red values indicate the entropy gap between incorrect and correct cases.}
    \label{fig:appendix_entropy}
\vspace{-10pt}
\end{figure*}

As shown in Figure \ref{fig:appendix_entropy}, for LLM2Vec\cite{LLM2Vec}, retrieval is performed using the LLM2Vec encoder, while answer generation is conducted by the backbone LLM. Across all three datasets, ShiftRAG exhibits the clearest entropy separation between correct and incorrect answers. On HotpotQA, the entropy gap is 0.8760 for ShiftRAG, compared with 0.3967 for GritLM\cite{GritLM}, 0.1291 for OneGen\cite{OneGen}, and 0.0415 for LLM2Vec. A similar trend is observed on 2WikiMultihopQA, where ShiftRAG again shows the largest entropy gap of 1.0004, substantially exceeding those of GritLM (0.3324), OneGen (0.2174), and LLM2Vec (0.0664). On MuSiQue, ShiftRAG also achieves the largest gap at 0.8933, compared with 0.3196 for GritLM, 0.1768 for OneGen, and 0.0626 for LLM2Vec. These results indicate that existing baselines remain relatively overconfident even when their answers are incorrect, as reflected by their limited entropy separation between correct and incorrect cases. In contrast, ShiftRAG assigns substantially higher entropy to incorrect decodings while maintaining lower entropy on correct ones, yielding a much clearer separation by answer correctness. This supports our claim that transferring retrieval-side knowledge through SST and soft knowledge tokens leads to more correctness-aware decoding behavior and better mitigates hallucination-prone overconfidence.

\subsection{Effect of Soft Knowledge Token Placement}\label{appendix_analysis_placement} 
\begingroup
  \renewcommand{\arraystretch}{0.80}
  \setlength{\tabcolsep}{3pt}

  \begin{table*}[h]
    \centering
    \footnotesize
    \caption{Effect of different placements of soft knowledge tokens on QA performance and soft-token injection latency in ShiftRAG.}
    \label{tab:appendix_ct_position}
    \begin{tabular*}{\textwidth}{@{\extracolsep{\fill}}lccccccccc}
      \toprule
      \multirow{2}{*}{\textbf{Category}} 
      & \multicolumn{3}{c}{\textbf{HotpotQA}} 
      & \multicolumn{3}{c}{\textbf{2WikiMultihopQA}} 
      & \multicolumn{3}{c}{\textbf{MuSiQue}} \\
      \cmidrule(lr){2-4} \cmidrule(lr){5-7} \cmidrule(lr){8-10}
      & EM & F1 & Time (ms) & EM & F1 & Time (ms) & EM & F1 & Time (ms) \\
      \midrule
      Pairwise & \textbf{0.5630} & \textbf{0.6827} & 1.65 & \underline{0.5060} & \textbf{0.5736} & 1.75 & \underline{0.2910} & \underline{0.4032} & 1.35 \\
      Between & 0.5580 & 0.6779 & 0.87 & 0.4980 & 0.5470 & 1.01 & 0.2880 & 0.4008 & 0.92 \\
      Behind & \underline{0.5610} & \underline{0.6790} & 0.86 & \textbf{0.5080} & \underline{0.5528} & 1.02 & \textbf{0.3050} & \textbf{0.4129} & 0.91 \\
      \bottomrule
    \end{tabular*}
  \end{table*}
\endgroup


We compare three soft knowledge token placement strategies in generation. Pairwise inserts each set of question–document-specific soft knowledge tokens immediately after its corresponding document, Between places all soft knowledge tokens between the document texts and the question, and Behind places all soft knowledge tokens after the textual input, i.e., behind the documents and question. Time (ms) measures the additional latency required to arrange and inject the generated soft knowledge tokens into the input embeddings. As shown in Table \ref{tab:appendix_ct_position}, Pairwise is consistently the slowest because it requires document-wise iterative insertion, taking 1.65/1.75/1.35 ms on HotpotQA, 2WikiMultihopQA, and MuSiQue, respectively, which is about 1.70× slower on average than Behind. Although Pairwise achieves slightly better performance on HotpotQA and 2WikiMultihopQA in some metrics, Behind yields the best average EM and the strongest overall trade-off between accuracy and efficiency. In particular, Behind consistently outperforms Between on all datasets and metrics, and achieves the best MuSiQue performance with 0.3050 EM and 0.4129 F1-score. Therefore, we adopt Behind as the default soft knowledge token placement in all experiments.

\subsection{$t$-SNE Visualization Before and After Training with Bidirectional and Unidirectional Attention}\label{appendix_analysis_tSNE}
\begin{figure*}[h]
    \centering
    \includegraphics[width=0.8\textwidth]{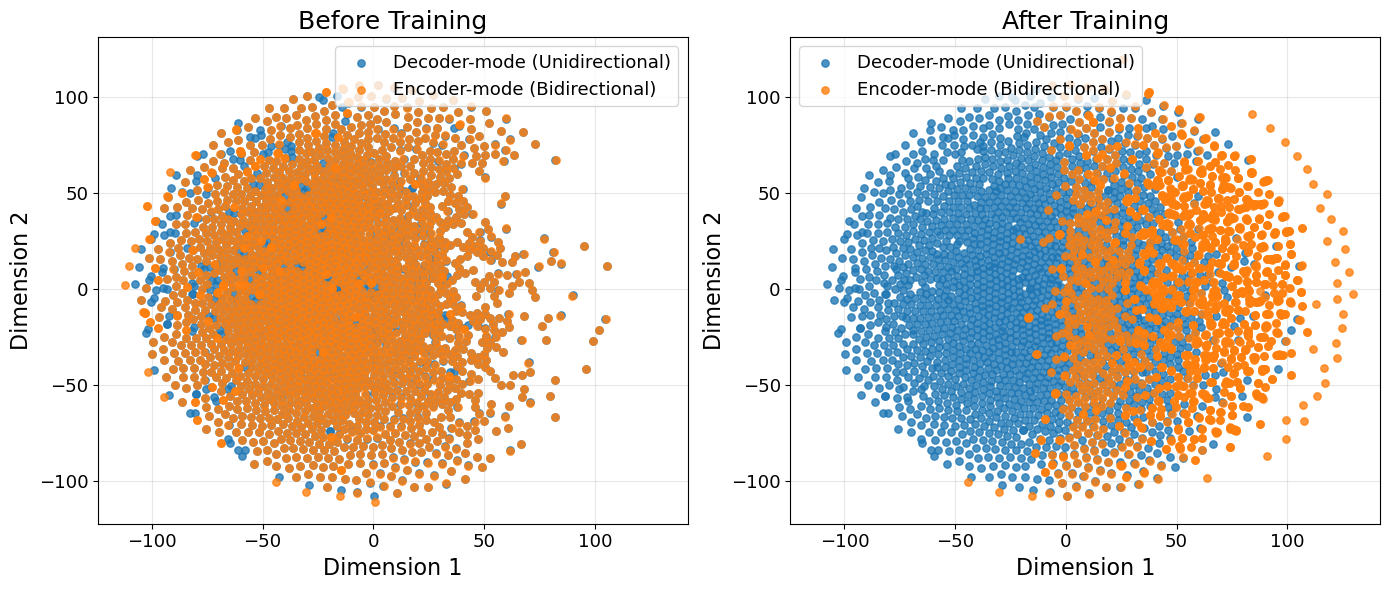}
    \caption{$t$-SNE visualization of document embeddings in the 2WikiMultihopQA dataset, comparing distributions before and after training under bi-directional and uni-directional attention mechanisms.}
    \label{fig:appendix_tSNE}
\vspace{-10pt}
\end{figure*}

To leverage pre-trained Large Language Models (LLMs) as encoders for the retrieval tasks, the standard uni-directional attention mechanism of decoder-only models must be transitioned to bi-directional attention, followed by task-specific fine-tuning for document embeddings. As illustrated in Figure \ref{fig:appendix_tSNE}, the distributions of document embeddings derived from each attention mechanism after fine-tuning exhibit a distinct separation. This clear visualization indicates that the embedding space optimized for retrieval (bidirectional attention) is fundamentally different from the embedding space utilized for generation (unidirectional attention). These findings provide strong empirical justification for the necessity of separate training for encoder and decoder modes. Furthermore, experimental results on the HotpotQA and MuSiQue datasets show consistent trends, reinforcing the generalizability of this observation.

\subsection{Effective Rank Before and After Encoder-mode Training across LLM Backbones}\label{appendix_analysis_eRank_backbones}
\begin{figure*}[h]
    \centering
    \includegraphics[width=1\textwidth]{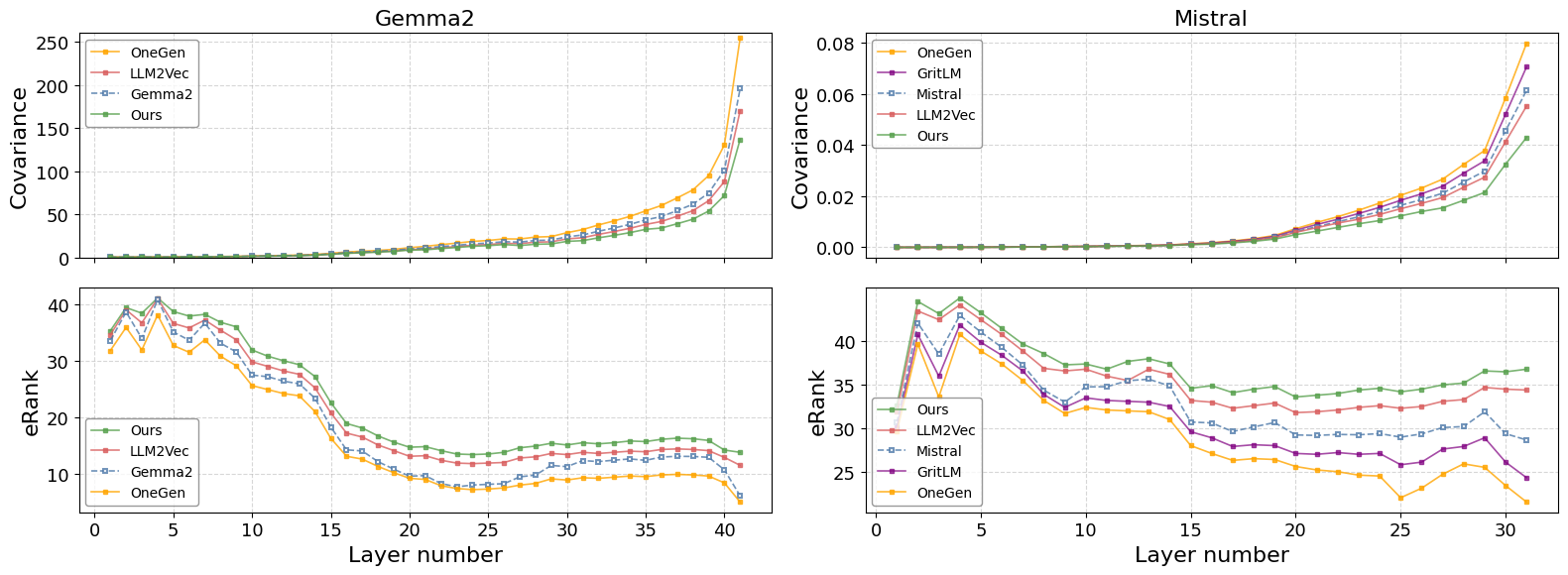}
    \caption{Layer-wise covariance between hidden representations of different tokens and effective rank (eRank) of token representations on Gemma2-9B (left) and Mistral-7B (right), respectively.}
    \label{fig:appendix_eRank1}
\vspace{-10pt}
\end{figure*}
\begin{figure*}[h]
    \centering
    \includegraphics[width=1\textwidth]{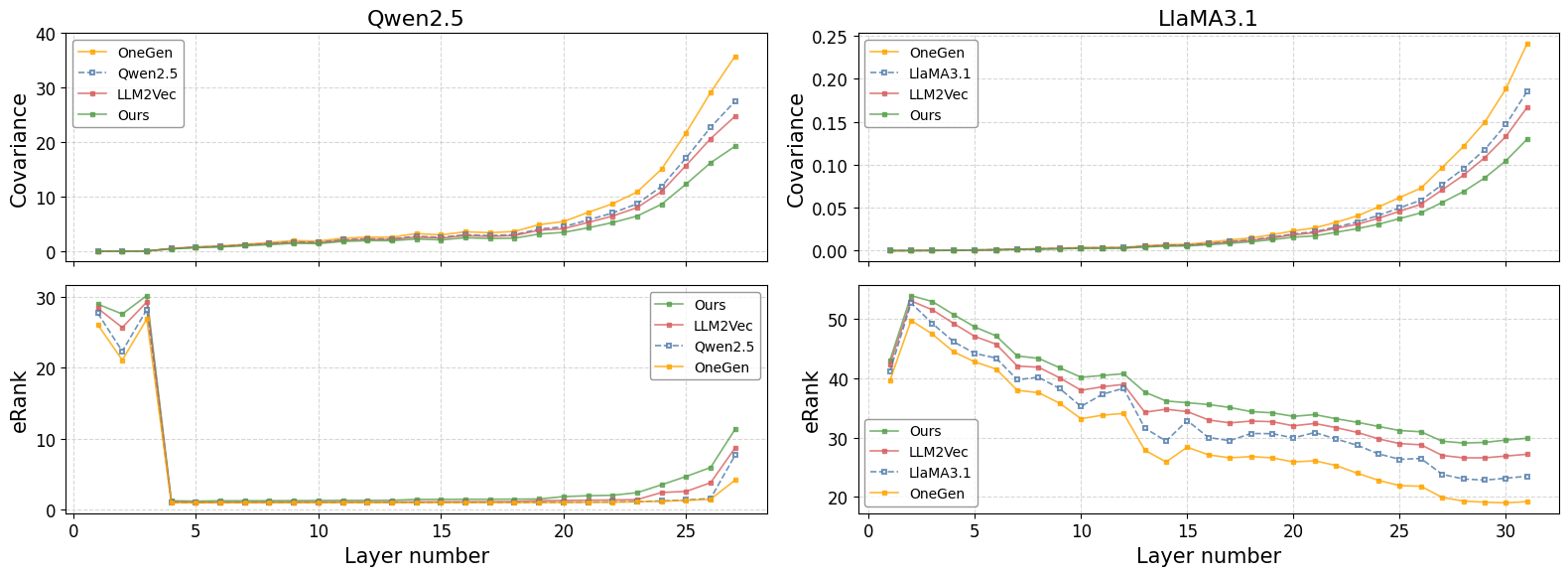}
    \caption{Layer-wise covariance between hidden representations of different tokens and effective rank (eRank) of token representations on Qwen2.5-7B (left) and LlaMA3.1-8B (right), respectively.}
    \label{fig:appendix_eRank2}
\vspace{-10pt}
\end{figure*}

We further analyze the layer-wise rank collapse tendency using covariance between hidden representations of different tokens and eRank across four backbone families. Figures \ref{fig:appendix_eRank1} and \ref{fig:appendix_eRank2} show consistent patterns on 2WikiMultihopQA using Gemma2-9B~\cite{Gemma2}, Mistral-7B~\cite{Mistral}, Qwen2.5-7B~\cite{Mistral}, and LlaMA3.1-8B~\cite{LlaMA3}. As the layer depth increases, covariance generally increases, while eRank decreases or remains low in deeper layers. This indicates that token hidden representations become increasingly aligned with each other and are concentrated into fewer effective directions. The trend is particularly clear in late layers, supporting our claim that decoder-only LLM representations can suffer from rank collapse when used as retrieval encoders.

The same tendency is consistently observed in unified LLMs. Compared with backbone LLMs, unified LLMs such as OneGen~\cite{OneGen} generally show higher covariance and lower eRank in deeper layers, indicating stronger directional alignment and reduced effective dimensionality. In Figure \ref{fig:appendix_eRank1}, GritLM~\cite{GritLM} on Mistral-7B also follows the same tendency as OneGen, showing stronger late-layer collapse than the backbone model. Figure \ref{fig:appendix_eRank2} further confirms that this behavior is not specific to Gemma2 or Mistral, but also appears on Qwen2.5 and LlaMA3.1. These results support our claim that unified LLMs tend to intensify rank collapse in deeper representations, which can make it difficult to preserve the token-level diversity needed for discriminative retrieval representations. We observe the same tendency on other MHQA benchmarks, as well as SimpleQA and NaturalQuestions.

\subsection{SST Effectiveness on Modern Large LLMs}\label{sec:appendix_review5}
\begin{table}[h]
  \centering
  \setlength{\tabcolsep}{6pt}
  \renewcommand{\arraystretch}{0.80}
  \footnotesize
  \caption{QA performance of large LLM generators with a fixed NV-Embed-v2 retriever. For each generator, both settings use the same SST-trained checkpoint and differ only in whether SST-generated soft knowledge tokens are provided at inference. The best performance within each generator is presented in bold.}
  \label{tab:appendix_large_llm_sst}
  \begin{tabular}{llcccccc}
    \toprule
    \textbf{Generator} & \textbf{Inference}
    & \multicolumn{2}{c}{\textbf{HotpotQA}}
    & \multicolumn{2}{c}{\textbf{2WikiMultihopQA}}
    & \multicolumn{2}{c}{\textbf{MuSiQue}} \\
    \cmidrule(lr){3-4}
    \cmidrule(lr){5-6}
    \cmidrule(lr){7-8}
    & & EM & F1 & EM & F1 & EM & F1 \\
    \midrule
    Gemma4-31B & w/o SST
    & 0.5920 & 0.7012
    & 0.4960 & 0.5976
    & 0.3120 & 0.4358 \\
    Gemma4-31B & w/ SST
    & \textbf{0.6140} & \textbf{0.7116}
    & \textbf{0.5280} & \textbf{0.6142}
    & \textbf{0.3410} & \textbf{0.4873} \\
    \midrule
    Qwen3.6-35B & w/o SST
    & 0.6050 & 0.7293
    & 0.5070 & 0.5624
    & 0.3290 & 0.4437 \\
    Qwen3.6-35B & w/ SST
    & \textbf{0.6310} & \textbf{0.7548}
    & \textbf{0.5330} & \textbf{0.5788}
    & \textbf{0.3660} & \textbf{0.5026} \\
    \bottomrule
  \end{tabular}
\end{table}

We investigate whether SST remains effective with larger, more recent LLM generators. Table \ref{tab:appendix_large_llm_sst} compares Gemma4-31B and Qwen3.6-35B with and without SST-generated soft knowledge tokens at inference. We use NV-Embed-v2 as the fixed retriever and keep the retrieved documents, document order, relevance scores, generator checkpoint, and decoding configuration unchanged within each comparison. Both settings retain the decoder-mode LoRA learned with SST, while w/o SST removes the soft knowledge tokens only at inference. This controlled comparison isolates the inference-time contribution of retrieval-side latent signals beyond raw textual evidence.

SST consistently improves EM and F1 across all datasets for both generators. The gains are particularly pronounced on MuSiQue, where SST improves EM by 9.29\% with Gemma4-31B (0.3120 → 0.3410) and 11.25\% with Qwen3.6-35B (0.3290 → 0.3660). The corresponding F1 improvements are 11.82\% (0.4358 → 0.4873) and 13.27\% (0.4437 → 0.5026), respectively. These results indicate that the benefits of SST are not restricted to the smaller backbones used in our main experiments. Larger generators can reason over raw retrieved text while still benefiting from explicitly transferred question–document alignment signals, suggesting that the latent interface provides complementary guidance for evidence-grounded generation.

\section{Hyperparameter Sensitivity}\label{sec:appendix_hyperparameter}
\begin{figure*}[h]
    \centering
    \includegraphics[width=1\textwidth]{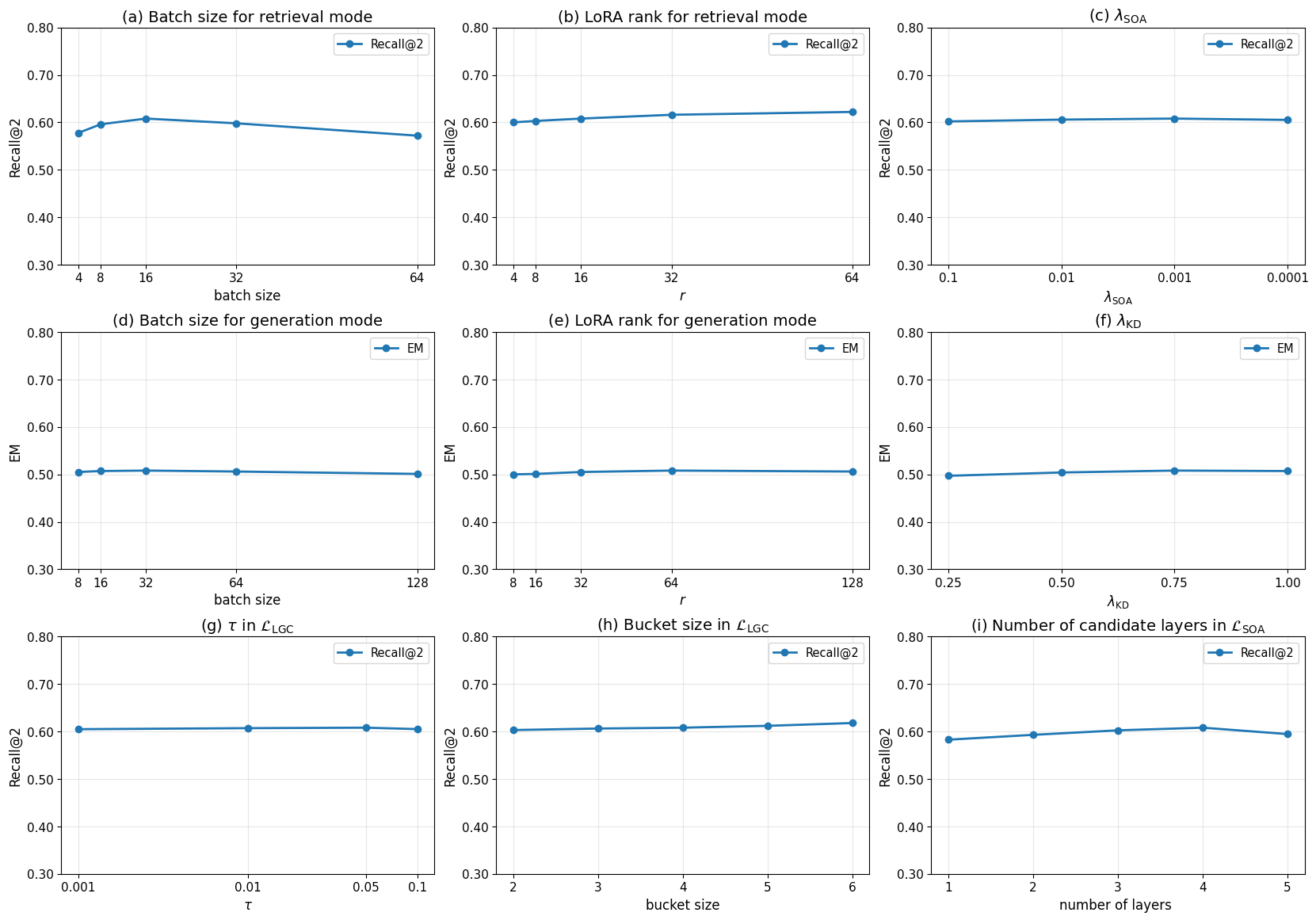}
    \caption{Hyperparameter sensitivity analysis on (a) batch size in encoder mode, (b) LoRA rank in encoder mode, (c) $\lambda_{\text{SOA}}$ in encoder mode training, (d) batch size in decoder mode, (e) LoRA rank in decoder mode, (f) $\lambda_{\text{KD}}$ in decoder mode training, (g) temperature $\tau$ in $\mathcal L_{\text{LGC}}$, (h) bucket size in $\mathcal L_{\text{LGC}}$. (i) number of candidate layers in $\mathcal L_{\text{SOA}}$.}
    \label{fig:appendix_hyperparameter}
\vspace{-10pt}
\end{figure*}

\subsection{Impact of Batch Size}\label{sec:appendix_hyperparameter_batchsize}
We conduct experiments on encoder mode and decoder mode to evaluate the impact of batch size. Figure \ref{fig:appendix_hyperparameter} (a) illustrates the R@2 metrics for batch sizes of 4, 8, 16, 32, and 64. The results show that a batch size of 16 yields the highest R@2, while performance drops sharply when the batch size exceeds 32. This trend can be attributed to the dynamics of training with in-batch negatives; smaller batch sizes increase the relative proportion of hard negatives within the batch, thereby facilitating the optimization of the layer-wise gap-aware contrastive (LGC) loss. However, when the batch size is reduced below 16, the absolute number of negative samples becomes insufficient, hindering the overall training process. In contrast to the retrieval performance, the decoder mode shows no significant variation in EM as the batch size increases, as illustrated in Figure \ref{fig:appendix_hyperparameter} (d). This is because the objective functions for decoder mode do not depend on the number of negative samples, unlike the encoder mode objective. While the remaining figures focus on the 2WikiMultihopQA dataset, other datasets, including HotpotQA and MuSiQue, follow a similar trend.

\subsection{Impact of LoRA Rank}\label{sec:appendix_hyperparameter_LoRA_rank}
We conduct experiments in both encoder and decoder modes to analyze the impact of the LoRA rank $r$. As illustrated in Figures \ref{fig:appendix_hyperparameter} (b) and \ref{fig:appendix_hyperparameter} (e), the model exhibits highly stable performance across various rank settings. Specifically, while R@2 and EM both peak at $r$ = 64, the overall variance remains marginal. In encoder mode, R@2 shows only minor fluctuations as $r$ increases from 4 to 64. Similarly, in decoder mode, EM remains relatively consistent across rank values, with only a negligible decline observed at $r$ = 128. While remaining figures focus on the 2WikiMultihopQA dataset, other datasets, including HotpotQA and MuSiQue, follow a similar trend.

\subsection{Impact of $\lambda_{\text{SOA}}$}\label{sec:appendix_hyperparameter_lambda_SOA}
We conduct experiments in encoder mode to analyze the effect of the hyper-parameter $\lambda_{\text{SOA}}$. Figure \ref{fig:appendix_hyperparameter} (c) shows R@2 for $\lambda_{\text{SOA}}$ values of 0.1, 0.01, 0.001, and 0.0001. The highest R@2 is achieved at $\lambda_{\text{SOA}}$= 0.001, although the performance remains remarkably stable across the tested range. While the remaining figures focus on the 2WikiMultihopQA dataset, other datasets, including HotpotQA and MuSiQue, follow a similar trend.

\subsection{Impact of $\lambda_{\text{KD}}$}\label{sec:appendix_hyperparameter_labda_KD}
We conduct experiments in decoder mode to analyze the effect of the hyper-parameter $\lambda_{\text{KD}}$. Figure \ref{fig:appendix_hyperparameter} (f) shows the EM for $\lambda_{\text{KD}}$ values of 0.25, 0.5, 0.75, and 1. The highest EM is achieved at $\lambda_{\text{KD}}$= 0.75, although the performance exhibits a gradual increase until $\lambda_{\text{KD}}$ reaches 0.75. While the remaining figures focus on the 2WikiMultihopQA dataset, other datasets, including HotpotQA and MuSiQue, follow a similar trend.

\subsection{Impact of Temperature $\tau$ in $\mathcal{L}_{\text{LGC}}$}\label{sec:appendix_hyperparameter_tau}
We conduct experiments in encoder mode to analyze the effect of the temperature hyper-parameter $\tau$ in the Layer-wise Gap-aware Contrastive (LGC) loss. Figure \ref{fig:appendix_hyperparameter} (g) shows the R@2 scores for $\tau$ values of 0.001, 0.01, 0.05, and 0.1. The highest R@2 is attained at $\tau$ = 0.05, with the model demonstrating minimal sensitivity to variations in $\tau$ within the tested range. While the remaining figures focus on the 2WikiMultihopQA dataset, other datasets, including HotpotQA and MuSiQue, follow a similar trend.

\subsection{Impact of Bucket Size in $\mathcal{L}_{\text{LGC}}$}\label{sec:appendix_hyperparameter_bucket_size}
We conduct experiments in encoder mode to analyze the effect of the bucket size in $\mathcal L_{\text{LGC}}$. Figure \ref{fig:appendix_hyperparameter} (h) shows the R@2 scores for the bucket size ranging from 2 to 6. The highest R@2 is attained using 6 layers as a bucket, although the performance increases only marginally as the bucket size increases. While the remaining figures focus on the 2WikiMultihopQA dataset, other datasets, including HotpotQA and MuSiQue, follow a similar trend.

\subsection{Impact of Number of Candidate Layers in $\mathcal{L}_{\text{SOA}}$}\label{sec:appendix_hyperparameter_candidate_layers}
We conduct experiments in encoder mode to analyze the effect of the number of the final layers used to calculate the $\mathcal L_{\text{SOA}}$. Figure \ref{fig:appendix_hyperparameter} (i) shows the R@2 scores for using the last 1 to 5 layers. The highest R@2 is attained using the last 4 layers. While the remaining figures focus on the 2WikiMultihopQA dataset, other datasets, including HotpotQA and MuSiQue, follow a similar trend.

\subsection{Impact of LLM Backbones}\label{sec:appendix_hyperparameter_llm_backbone}
\begin{figure*}[h]
    \centering
    \includegraphics[width=1\textwidth]{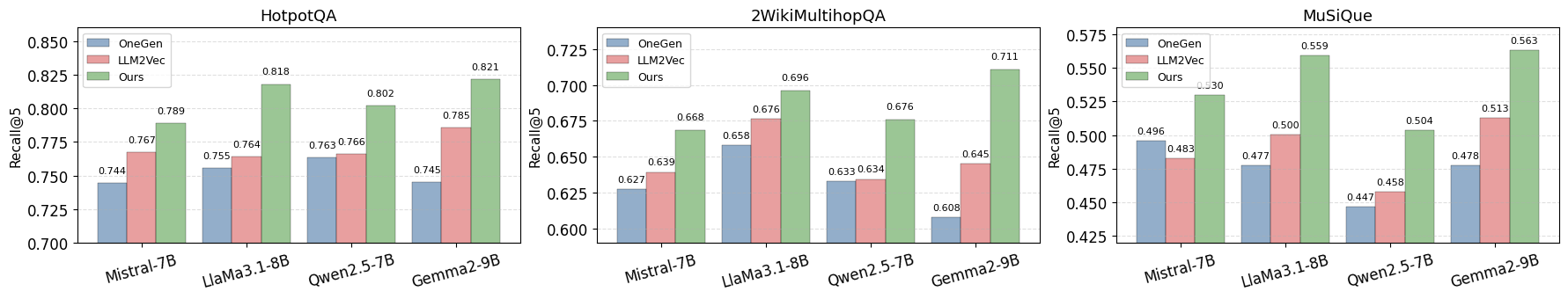}
    \caption{Retrieval performance of ShiftRAG and baselines across different LLM backbones.}
    \label{fig:appendix_backbone1}
\vspace{-10pt}
\end{figure*}
\begin{figure*}[h]
    \centering
    \includegraphics[width=1\textwidth]{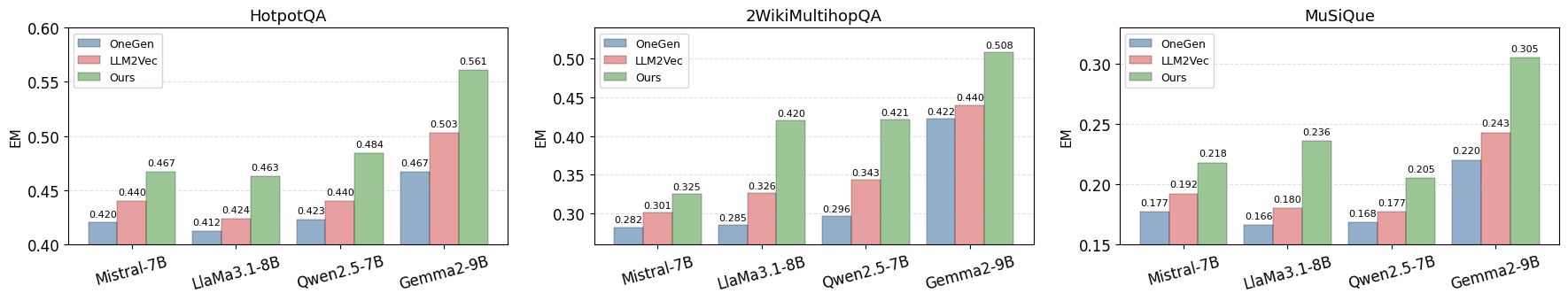}
    \caption{QA performance of ShiftRAG and baselines across different LLM backbones.}
    \label{fig:appendix_backbone2}
\vspace{-10pt}
\end{figure*}
To demonstrate the generalizability of our approach beyond the 2WikiMultihopQA dataset discussed in the main text, we include retrieval and end-to-end Question Answering performance for the HotpotQA and MuSiQue datasets across various Large Language Model (LLM) backbones, including Mistral-7B\cite{Mistral}, LlaMA3.1-8B\cite{LlaMA3}, Qwen2.5-7B\cite{Qwen2}, and Gemma2-9B\cite{Gemma2}.

As illustrated in Figures \ref{fig:appendix_backbone1} and \ref{fig:appendix_backbone2}, our experimental results indicate that ShiftRAG consistently outperforms LLM2Vec\cite{LLM2Vec} and OneGen\cite{OneGen} across all evaluated datasets and backbones. Regarding retrieval performance, the most significant gain was observed on the MuSiQue dataset using the LlaMA3.1-8B backbone, where ShiftRAG achieved an 11.75\% increase in R@5 compared to LLM2Vec. Furthermore, performance improvements were noted on HotpotQA with the LlaMA backbone at 7.04\% and on 2WikiMultihopQA with Gemma at 4.02\%. Regarding end-to-end QA performance, ShiftRAG achieved its most significant improvement on the MuSiQue dataset using LlaMA3.1-8B, recording a 31.11\% higher EM than LLM2Vec. Notable gains were observed on 2WikiMultihopQA with the LlaMA backbone at 28.83\% and on HotpotQA with the Gemma backbone at 11.53\%.

The performance enhancements remained consistent across diverse model architectures, including Mistral, LlaMA, Gemma, and Qwen. This stability suggests that the methodology of ShiftRAG is not dependent on a specific model structure but serves as a universally applicable optimization framework for RAG-based systems. Notably, the Gemma2-9B model exhibited the most robust and consistent performance across both retrieval and reasoning tasks compared to other backbones. Based on these results, Gemma was selected as the primary backbone for ShiftRAG.

\section{Case Study}\label{sec:appendix_case_study}
\subsection{Knowledge Comparison across Different Attention Policies}\label{sec:case1}
Figure \ref{fig:case_knowledge} illustrates how the attention policy used in encoder-mode training affects the retrieved evidence. For the question “Where was the place of death of Abdul-Aziz Bin Muhammad’s father?”, the bidirectionally trained retriever returns both Abdul-Aziz bin Muhammad and Muhammad bin Saud, where the latter provides the key evidence that the father’s place of death was “Diriyah”. In contrast, the unidirectionally trained retriever still retrieves “Abdul-Aziz bin Muhammad”, but fails to retrieve Muhammad bin Saud and instead returns lexically similar yet non-supporting documents such as Abdul Aziz bin Fahd and Abdullah ibn Muhammad. This comparison shows that the two attention mechanisms rely on different signals. Bidirectional retrieval is better at capturing global relational semantics needed to connect father and place of death, whereas unidirectional retrieval is more prone to lexical-overlap bias over surface entity names. As a result, the bidirectionally trained retriever identifies more answer-supporting evidence, while the unidirectionally trained retriever tends to retrieve documents that are locally similar in wording but less useful for reasoning.

\section{Prompt}\label{sec:appendix_prompt}
\subsection{Prompt 1. Prompt for Rationale Generation}\label{sec:appendix_prompt1}
We generate pair-specific textual rationales using GPT-5-nano~\cite{GPT-5-nano} with the prompt in Figure \ref{fig:appendix_prompt1}. For each question–document pair, the model produces a single concise sentence that explains whether the passage helps answer the question and how the passage contributes to the answer. If the passage is not useful, the prompt forces the model to first output a fixed rejection sentence and then briefly explain why the passage is irrelevant. The generated rationales are constructed offline and used only as teacher-side supervision for the knowledge distillation loss in decoder-mode training.

\subsection{Prompt 2. Prompt for Anchor Token Initialization}\label{sec:appendix_prompt2}
We use the prompt in Figure \ref{fig:appendix_prompt2} to initialize anchor token embeddings. Given a question, a passage, and a pre-generated textual rationale, the model is trained to reproduce the rationale in a concise sentence form. This procedure aligns anchor token embeddings with the generator’s token space before decoder-mode training.

\subsection{Prompt 3. Prompt for Decoder-Mode Training}\label{sec:appendix_prompt3}
We use the prompt in Figure \ref{fig:appendix_prompt3} for decoder-mode training. The model receives the retrieved documents and the question as textual input, while $<\texttt{LRN\_VIRTUAL\_TOKENS}>$ denotes the position where SST-generated soft knowledge tokens are inserted. The model is trained to generate only the final answer, without additional explanations or formatting.

\newpage
\begin{figure}[t]
\centering
\tcbset{colback=white, colframe=black!70, fonttitle=\bfseries}

\begin{tcolorbox}[title=Case Study: Knowledge Discrimination, width=\textwidth, colback=white, boxrule=1pt, arc=2pt, outer arc=2pt, top=2pt, bottom=2pt, left=2pt, right=2pt]

\footnotesize
\textbf{Question:} Where was the place of \textbf{\textcolor{red}{death}} of \textbf{\textcolor{orange}{Abdul-Aziz Bin Muhammad}}'s \textbf{\textcolor{blue}{father}}? \\
\textbf{Answer:} Diriyah

\vspace{4pt}

\par\noindent\rule{\linewidth}{0.4pt}\par

\vspace{4pt}
\textbf{Train Retrieval Mode using Bidirectional Attention} \\
\vspace{4pt}

\textbf{Document [1] (Title: \textcolor{orange}{Abdul-Aziz Bin Muhammad})} Imam Abdul Aziz bin Muhammad bin Saud d. 1803 was the second ruler of the First Saudi State and \textbf{\textcolor{blue}{son of Muhammad}} bin Saud. He was also the son-in-law of Muhammad bin Abdul Wahhab. He ruled from 1765 until 1803 \\
\textbf{is supporting}: True \\

\vspace{4pt}

\textbf{Document [2] (Title: Saud bin \textbf{\textcolor{orange}{Abdul-Aziz Bin Muhammad} bin Saud)}} Imam Saud al Kabeer bin \textbf{\textcolor{orange}{Abdul-Aziz Bin Muhammad}} bin Saud d. 1814 ruled the First Saudi State from 1803 to 1814. During his rule, Saud annexed Mecca and Medina from the Ottoman Empire. \\
\textbf{is supporting}: False \\

\vspace{4pt}

\textbf{Document [3] (Title: Muhammad bin Saud)} Muhammad ibn Saud ibn Muhammad ibn Muqrin Al-Maridi \textbf{\textcolor{red}{died}} 1765, also known as Ibn Saud, was the emir of Ad-Diriyyah and is considered the founder of the First Saudi State and the Saud dynasty, which are named for his father, Saud ibn Muhammad ibn Muqrin \textbf{\textcolor{red}{died}} 1725. Ibn Saud's family then known as the Al Muqrin traced its descent to the tribe of Banu Audi and Hanifa tribes but, despite popular misconceptions, Muhammad ibn Saud was neither a nomadic bedouin nor was he a tribal leader. Rather, he was the chief \"emir\" of an agricultural settlement near modern-day Riyadh, called Diriyah. Furthermore, he was a competent and ambitious desert warrior. \\
\textbf{is supporting}: True \\

\par\noindent\rule{\linewidth}{0.4pt}\par

\vspace{4pt}
\textbf{Train Retrieval Mode using Unidirectional Attention} \\
\vspace{4pt}

\textbf{Document [1] (Title: \textcolor{orange}{Abdul Aziz bin} Fahd)} Abdul Aziz bin Fahd is a Saudi Arabian Prince and member of the Royal House of Saud. \\
\textbf{is supporting}: False \\

\vspace{4pt}

\textbf{Document [2] (Title: \textcolor{orange}{Abdul-Aziz Bin Muhammad})} Imam Abdul Aziz bin Muhammad bin Saud d. 1803 was the second ruler of the First Saudi State and son of Muhammad bin Saud. He was also the son-in-law of Muhammad bin Abdul Wahhab. He ruled from 1765 until 1803.
\textbf{is supporting}: True \\

\vspace{4pt}

\textbf{Document [3] (Title: \textcolor{orange}{Abdullah ibn Muhammad})} \textbf{\textcolor{orange}{Abdullah ibn Muhammad}} also known as Tahir \textbf{\textcolor{orange}{ibn Muhammad}} was one of the sons of \textbf{\textcolor{orange}{Muhammad}} and Khadija. Qasim ibn \textbf{\textcolor{orange}{Muhammad}} was his older brother. Muhammad gave him the name of his father. 

\textbf{is supporting}: False \\

\vspace{4pt}

\end{tcolorbox}
\caption{A qualitative case study showing that bidirectionally trained retrieval preserves the full evidence chain needed for the answer, including the surface entity cue (\textcolor{orange}{orange}), the father relation (\textcolor{blue}{blue}), and the death-related clue (\textcolor{red}{red}), whereas unidirectionally trained retrieval overemphasizes surface lexical overlap (\textcolor{orange}{orange}) and retrieves non-supporting documents that miss the relation and death evidence (\textcolor{blue}{blue}, \textcolor{red}{red}) required to infer the correct answer.}
\label{fig:case_knowledge}
\end{figure}
\clearpage

\newpage
\begin{figure}[t]
\centering
\tcbset{colback=white, colframe=black!70, fonttitle=\bfseries}

\begin{tcolorbox}[title=Prompt: Rationale Generation, width=\textwidth, colback=white, boxrule=1pt, arc=2pt, outer arc=2pt, top=2pt, bottom=2pt, left=2pt, right=2pt]

\footnotesize
Given a question and a passage, write a rationale explaining whether the passage helps answer the question. If the passage helps answer the question, write exactly one concise English sentence explaining which fact in the passage is useful and how it supports answering the question. The passage may provide a direct answer, a required attribute of an entity, or one necessary piece of evidence for a multi-hop question.

\vspace{4pt}

If the passage does not help answer the question, first output exactly: \\
- Start "This passage does not help answer the given question.". Then write exactly one concise English sentence explaining why the passage is not useful, such as an entity mismatch, missing required information, no relation to the question, or irrelevant topic. \\
- Do not answer the question directly unless it is necessary to explain the passage's usefulness. Do not add labels, bullet points, JSON, citations, or any extra text. 
\\
\vspace{4pt}

\textbf{Passage:}\\
(Title: \{title\}) \{paragraph\} \\

\textbf{Question}:\\
\{question\}\\
\\
\textbf{Rationale:}
\end{tcolorbox}
\caption{Prompt for generating pair-specific textual rationales.}
\label{fig:appendix_prompt1}
\end{figure}
\begin{figure}[t]
\centering
\tcbset{colback=white, colframe=black!70, fonttitle=\bfseries}

\begin{tcolorbox}[
title=Prompt: Anchor Token Initialization,
width=\textwidth,
colback=white,
boxrule=1pt,
arc=2pt,
outer arc=2pt,
top=2pt,
bottom=2pt,
left=2pt,
right=2pt
]

\footnotesize
Given a question and a passage: if helpful, write one English sentence why; otherwise output exactly ``This passage does not help answer the given question.'' Then write one sentence why.

\vspace{4pt}

\textbf{Passage:}\par
(Title: \{title\}) \{paragraph\} \\

\textbf{Question:}\par 
\{question\} \\

\textbf{Rationale:}\par
\{rationale\}

\end{tcolorbox}
\caption{Prompt used for anchor token initialization with textual rationale supervision.}
\label{fig:appendix_prompt2}
\end{figure}
\begin{figure}[t]
\centering
\tcbset{colback=white, colframe=black!70, fonttitle=\bfseries}

\begin{tcolorbox}[title=Prompt: Question Answering, width=\textwidth, colback=white, boxrule=1pt, arc=2pt, outer arc=2pt, top=2pt, bottom=2pt, left=2pt, right=2pt]

\footnotesize
According to the previous relevant passages, please answer the following question. Only return the answer without any other words.

\vspace{4pt}

\textbf{Passage:}\\
Document [1] (Title: \{$\text{title}_1$\}) \{$\text{paragraph}_1$\} \\
Document [2] (Title: \{$\text{title}_2$\}) \{$\text{paragraph}_2$\} \\
... \\
Document [$K$] (Title: \{$\text{title}_K$\}) \{$\text{paragraph}_K$\} \\

\textbf{Question}:\\
\{question\}\\
\\
\textbf{Answer:} \\
\{answer\} \\

$<\texttt{Soft Knowledge Tokens}>$

\end{tcolorbox}
\caption{Prompt used for generation-mode training with retrieved documents and contextualized tokens.}
\label{fig:appendix_prompt3}
\end{figure}
\clearpage


\end{document}